\documentclass[12pt,letterpaper,openright,oneside]{report}

\usepackage{amsmath}
\usepackage{amsfonts}
\usepackage{graphicx} 
\usepackage{setspace} 
\usepackage{tocloft}
\usepackage{bbm}
\usepackage{booktabs}
\usepackage{siunitx}
\usepackage{hyperref}
\usepackage{array,multirow,times,epsfig,float}
\usepackage{enumitem}
\usepackage[export]{adjustbox}
\usepackage{amssymb}
\usepackage{algorithm,algorithmic}
\usepackage{algorithm}
\usepackage{listings}
\usepackage{xcolor,pifont}
\newcommand*\colourcheck[1]{%
  \expandafter\newcommand\csname #1check\endcsname{\textcolor{#1}{\ding{52}}}%
}
\colourcheck{green}
\usepackage{colortbl}
\usepackage{wrapfig, blindtext}
\DeclareMathOperator*{\argmax}{arg\,max}

\usepackage{chngcntr}
\usepackage[top=2.54cm, bottom=3.81cm, left=2.54cm, right=2.54cm]{geometry}

\usepackage{multirow} 
\usepackage{listings} 
\usepackage{pdfpages} 
\usepackage{etoolbox}
\usepackage{chngcntr}
\counterwithout{figure}{chapter}

\usepackage[numbers]{natbib}
\usepackage{fancyhdr}

\newcommand{\tdesc}{\textbf{Embracing Annotation Efficient Learning (AEL) for Digital Pathology and Natural Images}}

\begin{document}

\pagenumbering{roman}
\pagestyle{fancy}
\fancyhf{}
\fancyhead[R]{\roman{page}}
\fancypagestyle{plain}{%
    \renewcommand{\headrulewidth}{0pt}%
    \fancyhf{}%
    \fancyhead[R]{\roman{page}}%
}
\renewcommand{\headrulewidth}{0pt}%

\setcounter{tocdepth}{3}
\setcounter{secnumdepth}{3}

\clearpage
\thispagestyle{empty}

\begin{center}
\vspace*{\fill}
\Large{\tdesc}

\normalsize{
\vspace{15mm}

by

Eu Wern Teh

\vspace{15mm}
\vspace{15mm}

A Thesis

presented to

The University of Guelph

\vspace{15mm}
\vspace{15mm}

In partial fulfilment of requirements

for the degree of

Doctor of Philosophy

in

Engineering

\vspace{15mm}
\vspace{15mm}

Guelph, Ontario, Canada

\copyright \:Eu Wern Teh, August, 2022

}
\end{center}
\vspace*{\fill}

\newpage

\clearpage
\thispagestyle{empty}

\addcontentsline{toc}{chapter}{Abstract}

\doublespacing

\begin{center} 
\textbf{\Large{ ABSTRACT}}

\vspace{16mm}

\MakeUppercase{Embracing Annotation Efficient Learning (AEL) for Digital Pathology and Natural Images}

\end{center}

\vspace{14mm}

\noindent{Eu Wern Teh	\hfill Advisor:\hspace{42mm}}

\noindent{University of Guelph, 2022	\hfill	Dr. Graham W. Taylor\hspace{19mm}}

\vspace{20mm}

Jitendra Malik~\cite{malik_twitter} once said, ``Supervision is the opium of the AI researcher". Most deep learning techniques heavily rely on extreme amounts of human labels to work effectively. In today's world, the rate of data creation greatly surpasses the rate of data annotation.
Full reliance on human annotations is just a temporary means to solve current closed problems in AI. In reality, only a tiny fraction of data is annotated. 
Annotation Efficient Learning (AEL) is a study of algorithms to train models effectively with fewer annotations.
To thrive in AEL environments, we need deep learning techniques that rely less on manual annotations (e.g., image, bounding-box, and per-pixel labels), but learn useful information from unlabeled data.  
In this thesis, we explore five different techniques for handling AEL: 
\thispagestyle{empty}
\begin{enumerate}
  \item 
  Learning with Less Data via Weakly Labeled Patch Classification in Digital Pathology, International Symposium on Biomedical Imaging (2020), Eu Wern Teh and Graham W. Taylor
  \item 
  Learning with Less Labels in Digital Pathology via Scribble Supervision from Natural Images, International Symposium on Biomedical Imaging (2022), Eu Wern Teh and Graham W. Taylor
  \item 
  Understanding the Impact of Image and Input Resolution on Deep Digital Pathology Patch Classifiers, Canadian Conference on Computer and Robot Vision (2022), Eu Wern Teh and Graham W. Taylor
 \thispagestyle{empty}
  \item 
  ProxyNCA++: Revisiting and Revitalizing Proxy Neighborhood Component Analysis, European Conference on Computer Vision (2020), Eu Wern Teh, Terrance DeVries, and Graham W. Taylor
  \item 
  The GIST and RIST of Iterative Self-Training for Semi-Supervised Segmentation, Canadian Conference on Computer and Robot Vision (2022), Eu Wern Teh, Terrance DeVries, Brendan Duke ,Ruowei Jiang, Parham Aarabi, and Graham Taylor
  \thispagestyle{empty}
\end{enumerate}


\chapter*{\centering \MakeUppercase{ \Large{Acknowledgements}}}
\doublespacing \addcontentsline{toc}{chapter}{Acknowledgements}
There are countless people that I want to give thanks to for their support in my adventurous Ph.D. journey. 
Firstly, I want to thank GOD for his continuous support during my Ph.D. This humbling ride has brought me closer to you. Thank you for placing the right persons and opportunities in my paths, allowing me to grow as a person and a researcher. May you continue to watch over me as I transition to the next phase of my life. 

Secondly, I want to thank my parents, Swee Lai Teh and Eng Nee Chung, and my siblings, Jo Way Teh and Eu Jin Teh, for their love, support, and encouragement during my Ph.D. They have listened to me attentively during my hardships. May GOD bless them with good health and overflowing wisdom. 

Thirdly, I want to thank my advisor, Graham W. Taylor, for giving me the opportunity and funding to study under his wing. Your advice and guidance have made me a better researcher. I also want to thank our lab managers, Brittany Reiche and Magdalena Sobol, for their efforts in improving my manuscripts. Your reviews have made me a better academic writer. 

Fourthly, I want to thank my fellow lab mates for their support, insights, and fun companies. 
Specifically, I want to thank Terrance Devries, Brendan Duke, Boris Knyazev, Angus Galloway, Michal Lisicki, Kristina Kupferschmidt, Adam Balint, Colin Brennan, Shashank Shekhar, Rylee Thompson, Mahmoud Gamal, Sara El-Shawa, Nolan Dey, Kenyon Tsai, Vikram Voleti, Devinder Kumar, Ahmed Elshamli, Eric Taylor, Ethan Jackson, Alaa El-Nouby, Vithursan Thangarasa, Nikhil Sapru, Dhanesh Ramachandram, Thor Jonsson, Maeve Kennedy, and many more. I want to thank our IT wizards: Joel Best and Matthew Kent, for their seamless support in handling and maintaining our servers. I am lucky to have the opportunity to be part of such an amazing group.    


\clearpage

\singlespacing

\renewcommand{\contentsname}{\hspace{150pt}\Large{TABLE OF CONTENTS}}
\addcontentsline{toc}{chapter}{Table of Contents}
\tableofcontents

\renewcommand{\listtablename}{\hspace{170pt}\Large{LIST OF TABLES}}
\addcontentsline{toc}{chapter}{List of Tables}
\listoftables
\clearpage

\renewcommand{\listfigurename}{\hspace{170pt}\Large{LIST OF FIGURES}}
\addcontentsline{toc}{chapter}{List of Figures}
\listoffigures

\chapter*{\centering \MakeUppercase{ \Large{Symbols, Datasets, and Abbreviations}}}
\singlespacing \addcontentsline{toc}{chapter}{Symbols, Datasets, and Abbreviations}

\begin{table}[H]
\centering
\setlength{\tabcolsep}{8pt}
\begin{tabular}{ll}
\hline
Symbol & Meaning \\ \hline \hline
$f(x)$ & A function that accepts input $x$. \\
$d(x,y)$ & A distance function that computes the distance between $x$ and $y$.\\
$x_i$ & A variable $x$ indexed with $i$.\\
$\log(x)$ & A natural logarithm function that accepts input $x$. \\
$\exp(x)$ & An exponent function that accepts input $x$. \\
$||x||_2$ & The $L^2$-Norm of vector $x$. \\
$\sum_{i=1}^{N}x_i$  & A summation of $x_i$ from 1 to $N$. \\
\hline
\end{tabular}
\caption{Table of commonly used mathematical symbols. The description of mathematical symbols are defined before or after a given equation or algorithm.}
\end{table}

\begin{table}[H]
\centering
\setlength{\tabcolsep}{8pt}
\begin{tabular}{ll}
\hline
Dataset & Description \\ \hline \hline
CRC & Colorectal cancer dataset~\cite{kather2016multi} \\
PCam & Patch Camelyon breast cancer dataset~\cite{veeling2018rotation}\\
KimiaPath24 & Kimia pathology dataset~\cite{babaie2017classification} \\
Pascal VOC 2012 & Pascal visual object classes challenge 2012 dataset~\cite{pascal-voc-2012} \\
CUB200 & Caltech-UCSD birds dataset~\cite{wah2011caltech} \\
CARS196 & Stanford cars dataset~\cite{KrauseStarkDengFei-Fei_3DRR2013} \\
SOP & Stanford online product dataset~\cite{song2016deep} \\
InShop & In shop clothing retrieval dataset~\cite{liu2016deepfashion} \\
Cityscapes & City and urban street scenes dataset~\cite{Cordts2016Cityscapes} \\
\hline
\end{tabular}
\caption{Table of datasets.}
\end{table}

\begin{table}[H]
\centering
\setlength{\tabcolsep}{8pt}
\begin{tabular}{ll}
\hline
Abbreviation & Meaning \\ \hline \hline
AEL & Annotation Efficient Learning \\
NI & Natural Image\\
DP & Digital Pathology \\
SOTA & State-Of-The-Art \\
CNN & Convolutional Neural Network \\
SVM & Support Vector Machines \\
GPU & Graphic Processing Units \\
DML & Distance Metric Learning \\
NCA & Neighborhood Component Analysis \\
GMP & Global Max Pooling \\
GAP & Global Average Pooling \\
NMI & Normalized Mutual Information \\
GIST & Greedy Iterative Self-Training \\
RIST & Random Iterative Self-Training \\
NLP & Natural Language Processing \\
CL & Consistency Loss \\
LE & Label Erase \\
TS & Temperature Scaling \\
ASPP & Atrous Spatial Pyramid Pooling \\
\hline
\end{tabular}
\caption{Table of abbreviations.}
\end{table}

\clearpage

\onehalfspace

\normalsize
\makeatletter

\pagenumbering{arabic}
\pagestyle{fancy}
\fancyhf{}
\fancyfoot[C]{\arabic{page}}
\fancypagestyle{plain}{%
    \renewcommand{\headrulewidth}{0pt}%
    \fancyhf{}%
    \fancyfoot[C]{\arabic{page}}%
}
\renewcommand{\headrulewidth}{0pt}%


\chapter{Introduction}
\pagenumbering{arabic} \setcounter{page}{1} 

Humanity quest to solve computer vision began in 1966 when a group of MIT researchers attempted to solve a significant part of the visual system in a single summer internship~\cite{papert1966summer}. The failure of the infamous Summer Vision project shows us the difficulty of human vision, something that we take for granted as it is such a natural skill for humans. 
We have come quite far in computer vision since 1966, primarily due to the advancement in Deep Learning.
Today most deep learning models are trained in a fully supervised environment, where large amounts of human-annotated data are required~\cite{goodfellow2016deep}. The major driving force for the advancement of supervised deep learning is 

\begin{figure}[H]
  \centering
  \includegraphics[width=6.40in, ]{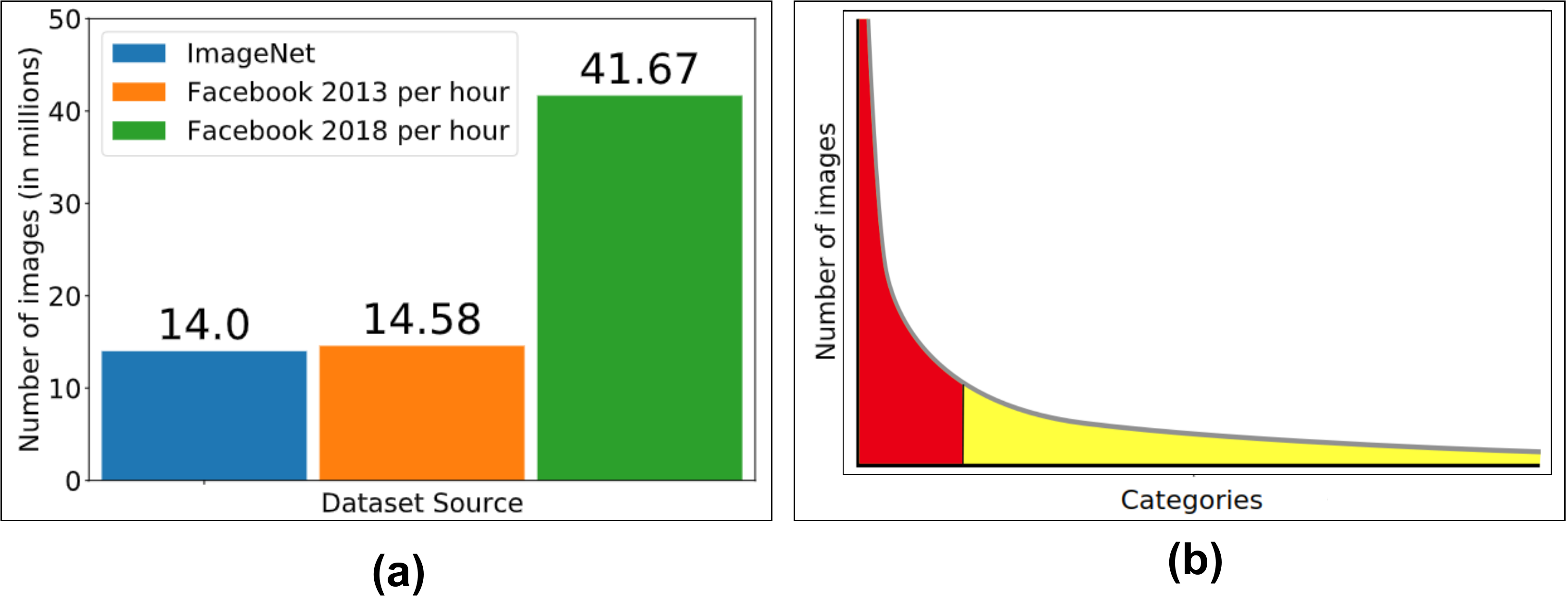}
  \caption{Panel (a) shows the average number of images generated by Facebook users in an hour in 2013 and 2018, compared to the total number of images in the ImageNet dataset. Panel (b) illustrates a synthetic example of a long-tailed dataset. The red area represents the head distribution, while the yellow area represents the long-tailed distribution. }
  \label{fig:mot0}
\end{figure}

\noindent mainly credited to ImageNet, a large-scale visual recognition dataset with about 14 million images and 22 thousand categories~\cite{russakovsky2015imagenet}. It took approximately 19 human years to annotate this massive dataset. 

However, the sheer amount of new images added per day to the web in today's world greatly surpasses the total number of images in ImageNet. For instance, there were, on average, approximately 350 million images added to Facebook in 2013, and this number grew to one billion images by 2018~\cite{facebook2013,facebook2018}. To put things into perspective, Facebook users uploaded approximately one ImageNet size dataset per hour in 2013 and created close to three times ImageNet size dataset per hour in 2018. If we were to attempt to annotate the number of images created in an hour for 2018, it would take approximately 55 human years. It is fair to say that the data growth dramatically surpasses the number of person-hours to annotate them.

Apart from the overwhelming size of newly created data, a significant problem of deploying a machine learning system is the appearance of head and tail distributions in the real-world data (Figure~\ref{fig:mot0}b). 
The head part of the distribution consists of categories with a large number of data, while the tail part of the distribution consists of categories with a small number of data. 
A study has shown that knowledge learned from the head distribution works poorly on unseen data from the tail~\cite{van2017devil}. As the volume of the dataset increases, the number of new categories also increases. In classification, which is the most common machine learning task, when we have new categories, we are forced to retrain machine learning models to adapt to the newly discovered data distribution. Considering the rate of data annotations and the rate of data creation, we are very likely to get stuck in the never-ending train-annotate-train cycle. 

\begin{figure}[H]
  \centering
  \includegraphics[width=6.40in, ]{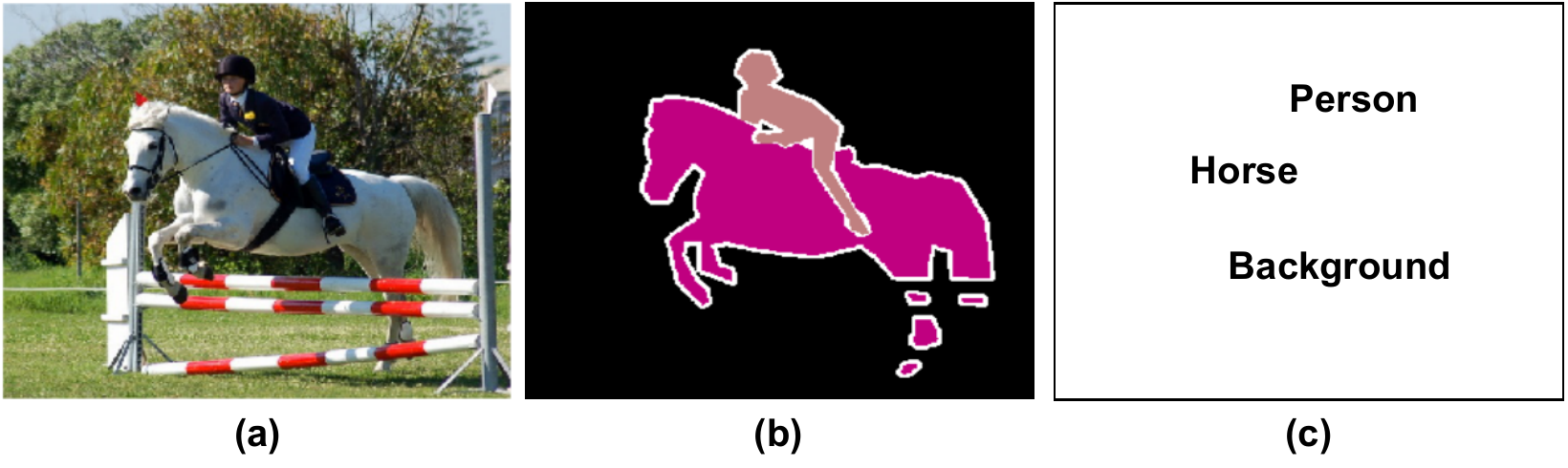}
  \caption{Panel (b) shows the corresponding full segmentation masks and panel (c) shows the corresponding class level annotations given an image in panel (a). }
  \label{fig:mot1}
\end{figure}

\begin{figure}[h]
  \centering
  \includegraphics[width=6.40in, ]{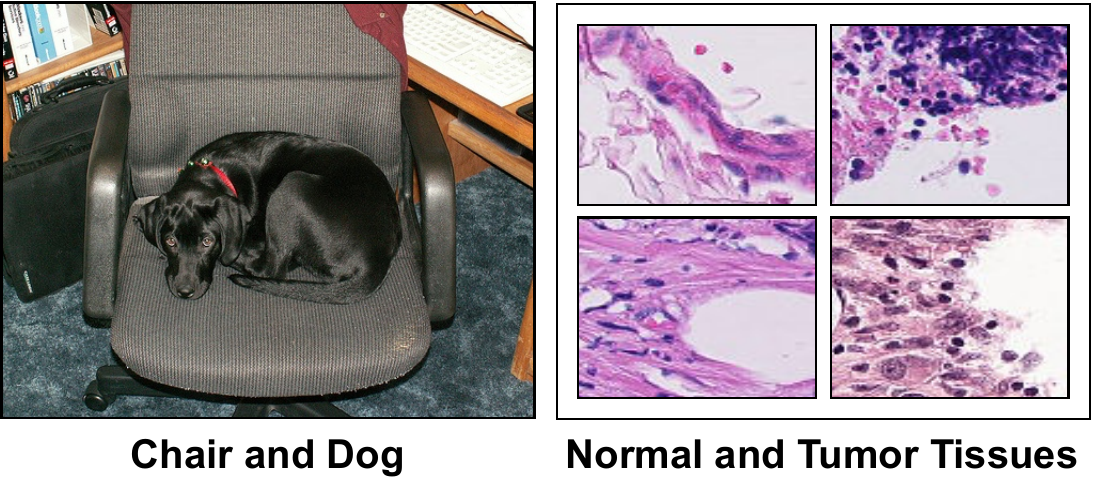}
  \caption{Examples of class level annotation for natural images and digital pathology images.}
  \label{fig:mot2}
\end{figure}

Data annotation costs can be astronomical. The annotation cost hinges on the quantity of unlabeled and the expertise of human annotators. For instance, obtaining full segmentation annotations is relatively much harder when compared to obtaining class-level annotations (Figure~\ref{fig:mot1}). On the other hand, it is much more expensive to obtain class-level annotation from a domain expert such as a medical doctor when compared to a layperson ~\cite{md_salary,dp_edu1,dp_edu2} (Figure~\ref{fig:mot2}).

\begin{figure}[h]
  \centering
  \includegraphics[width=6.30in, ]{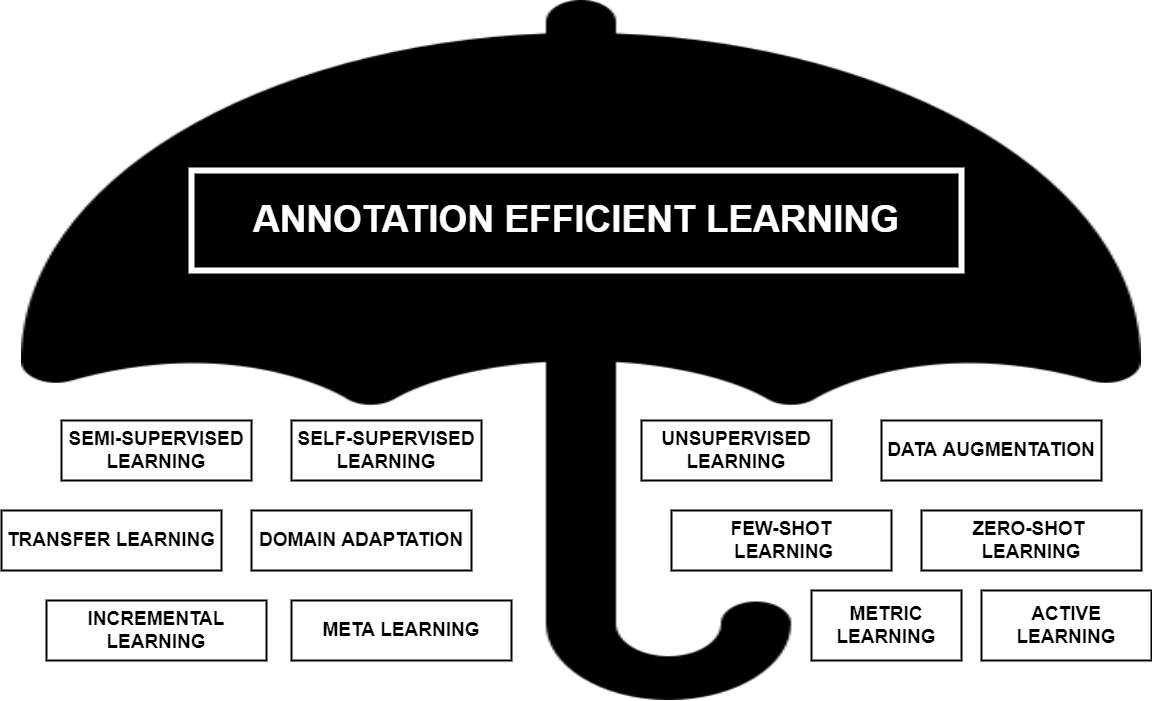}
  \caption{This figure illustrates a few Deep Learning topics that fall under the umbrella of Annotation Efficient Learning.}
  \label{fig:umb}
\end{figure}

Seeing that data annotation is expensive and deep learning models are annotation hungry, this thesis explores Annotation Efficient Learning (AEL), which we defined as a study of algorithms to train a model efficiently with fewer annotations. 
In the AEL environments, we assume that annotations are scarce and limited while unlabeled data is bountiful. There is a lot of work in Deep Learning that falls under AEL (Figure~\ref{fig:umb}). However, we focus on transfer learning, data augmentation, metric learning, and semi-supervised learning to reduce the scope of this thesis.

\section{Thesis Contributions}

We explore five different techniques in handling AEL. In Chapter~\ref{isbi2020} to \ref{midl2022}, we focus on handling AEL for the Digital Pathology (DP) domain, where microscopic images are analyzed and annotated by medical experts. In the DP domain, handling AEL effectively is crucial due to the scarcity of annotated data caused by high annotation costs.   
We utilize weakly labeled data to aid transfer learning for DP tasks (Chapter~\ref{isbi2020}). We learn transferable features by fine-tuning on a weakly labeled dataset provided by non-medical experts and transfer the learned knowledge to a target DP dataset, which is annotated by medical experts. We leverage the availability of inexpensive annotation from a layperson to aid DP models in achieving better performance on a DP dataset, where an expert's annotations are scarce. 
Next, we explore the use of scribble supervision in the natural image (NI) domain to boost the performance of DP models (Chapter~\ref{isbi2022}). We show that spatial annotation such as scribbles and segmentation labels are important for transferring learning from the NI to the DP domain. We also show that scribble annotations, which are significantly much easier to obtain than segmentation labels, can perform equally well compared to segmentation labels on a transfer learning task from the NI to DP domain. 
Furthermore, we study the effect of image and input resolution on DP models (Chapter~\ref{midl2022}). We discover that the performance of DP models can be improved significantly in the scarcely annotated environment by increasing the image and input resolution. Our study shows that an increase in image and input resolution allows fine-grained features to be captured while retaining global features. 


In Chapter~\ref{eccv2020}, we study algorithms that tackle the zero-shot learning environment, which is the lower bound of AEL. In the zero-shot learning environment, our goal is to explore algorithms that can be generalized to unseen classes. We utilize metric learning to learn the distances between samples in the dataset by grouping data points with similar labels and repelling data points with different labels. We revisit a proxy-based metric learning technique and propose six improvements to boost the performance of models in image retrieval. 
Lastly, we explore an iterative self-training technique to tackle semi-supervised semantic segmentation in the NI domain (Chapter~\ref{crv2022}). We discover that a naive iterative self-training strategy that combines both the human annotation and pseudo annotation leads to performance degradation due to overconfidence in the model's prediction. 
We further propose a greedy and random iterative self-training strategy to overcome this performance degradation and show that our strategy can further improve the models' performance trained by other semi-supervised learning methods. 
\clearpage

\patchcmd{\chapter}{\if@openright\cleardoublepage\else\clearpage\fi}{}{}{}
\makeatother

\chapter{Background}\label{chap:2}

In this thesis, we explore methods of handling Annotation Efficient Learning (AEL) for both natural images~\footnote{Natural images are normal images taken from day-to-day environments by using in-expensive RGB cameras} and digital pathology. 
Chapter\ref{rel:hist} and \ref{rel:dp} give a brief overview of deep learning and digital pathology.
The remaining sections in this chapter will introduce relevant work of handling AEL, such as data augmentation (Chapter~\ref{rel:da}), transfer learning (Chapter~\ref{rel:transfer}), metric learning (Chapter~\ref{rel:metric}), and semi-supervised learning (Chapter~\ref{rel:semi}).

\section{Deep Learning}~\label{rel:hist}

Deep learning can be traced back to the 1960s, when Frank Rosenblatt introduced a feed-forward perceptron, a biologically inspired machine learning model that can solve binary classification problems~\cite{rosenblatt1958perceptron}. Despite its popularity,  there are a few major drawbacks. Firstly, a single  Roseblatt's perceptron cannot solve the eXclusive-OR (XOR) problem, which is linearly non-separable~\cite{block1970review}~\footnote{XOR problem can be solved by having more than one layer~\cite{goodfellow2016deep}.}. Secondly, this simple perceptron requires heuristic weight adjustments that have to be performed for every new task because the optimal weight may differ depending on the tasks.  In 1986, Rumelhart, Hinton, and  Williams~\cite{rumelhart1986learning} popularized backward propagation in a multi-layered perceptron where the weights of a neuron can be tuned automatically with respect to its loss function.   
Later in 1989, Cybenko et al.~\cite{cybenko1989approximation} showed that a feed-forward neural network coupled with a non-linear activation function (Sigmoid) is a universal approximator if it has enough hidden units.  
In 1980, inspired by Hubel and Wiesel's work~\cite{hubel1959receptive} on the receptive field of a single neuron in the visual cortex, Fukushima~\cite{fukushima1980neocognitron} introduced the very first Convolutional Neural Network (CNN), consisting of convolution and downsampling layers. Later in 1989, Le Cun et al.~\cite{lecun1989backpropagation} used backpropagation to update the weights in the convolutional neural networks directly from images of handwritten numbers, showing that automatic weight updates are better than manually hand-crafted convolutional layers.
Despite some success in the 1980s, deep learning research was thwarted by the lack of computation power and data quantity. Deep Learning was superseded by other competitive learning algorithms, such as Support Vector Machines (SVM) and its extensions.

In the late 2000s, deep learning became the dominant algorithm in solving computer vision tasks. In 2009, Deng et al. created a large-scale supervised dataset known as ImageNet~\cite{imagenet_cvpr09}, where they aimed to use big data to promote data-driven pattern recognition. Around the same time, several research groups had started to use Graphics Processor Units (GPU) to accelerate the speed of deep network training~\cite{raina2009large}.
In 2012, deep learning outperformed other learning algorithms by a large margin in the image recognition task where it utilized both a large-scale dataset (ImageNet) and GPU's fast computation speed.
 From 2012 onwards, there has been an extraordinary rise in research on deep learning in the computer vision community.

\section{Digital Pathology}~\label{rel:dp}

Digital Pathology (DP) is a field of medical imaging where microscopic images are analyzed.  
Most machine learning and computer vision works are designed to handle Natural Image (NI) tasks. However, there are distinctions between NI and DP domains. 
One key difference between DP and NI domain is the image size. In DP, each WSIs are usually giga-pixels in size, whereas in NI, each image is usually mega-pixels in size. Another key difference is that a coarse-grain view (seeing the whole image) conveys little to no meaningful information in DP. In NI, a human can describe the content of an image just by viewing the whole image.
Additionally,  NI images as a whole are often use directly for machine learning training. In DP, researchers usually divide each giga-pixel image into small patches at various zoom-level~\cite{45922,aresta2019bach}. After converting WSI to image patches, a standard pipeline in deep learning can be used for training. 
In NI, human annotations are relatively inexpensive when compared to DP because the annotation can be completed by laypersons. 
In DP, the cost of human annotation increases by a large margin because annotations are obtained from medical experts that required specialized knowledge~\cite{md_salary,45922}. 
The high annotation cost of DP highlights the importance of handling AEL, where human annotations are limited. 

\section{Data Augmentation}~\label{rel:da}
 
A neural network is a universal approximator that can approximate any continuous function if it has enough hidden units~\cite{cybenko1989approximation}. Since a deep model consists of many neural networks, it can easily overfit a given dataset, which is why deep learning requires significant amount of annotated data to be effective.
In annotation efficient learning environments, where annotated data are limited, data augmentation is an excellent strategy to combat overfitting without collecting more annotations.   
Overfitting can be reduced via simple data augmentation strategies such as random cropping, random rescaling, and image jittering~\cite{krizhevsky2012imagenet}. 
Srivastava et al.~\cite{srivastava2014dropout} discovered Dropout where they randomly zero-outed the signals within a deep network. 
DeVries et al. proposed CutOut, which randomly blocks out a continuous chunk of an image,  providing image-level augmentation during training~\cite{devries2017improved}. Ghiasi et al. proposed Dropblock that randomly blocks out a continuous chunk of feature space, yielding feature-level augmentation during training~\cite{ghiasi2018dropblock}. Zhang et al. proposed MixUp to interpolate between two images and their corresponding annotations~\cite{zhang2018mixup}. Yun et al. proposed CutMix that creates a composite image by combining two images and their corresponding annotations~\cite{yun2019cutmix}. All these augmentation techniques regularize deep learning networks to reduce overfitting and improve generalization~\footnote{The goal of machine learning is to train a model that can perform well to unseen data, a.k.a generalization~\cite{goodfellow2016deep}. Overfitting happens when a model memorizes the training data, and underfitting happens when a model fails to learn from the training data. There is usually a range of sweet spots between overfitting and underfitting, where a decrease in overfitting will yield better generalization. This trade-off is also known as the bias-variance trade-off.}.

\section{Transfer Learning}~\label{rel:transfer}

Transfer Learning is a type of machine learning strategy where knowledge is transferred from a source dataset to a target dataset. 
With the help of a large scale supervised dataset like ImageNet~\cite{russakovsky2015imagenet}, deep learning can extend its reach through transfer learning to solve many computer vision problems such as object detection, object segmentation, object tracking, image retrieval, and many more.
 In deep learning, fine-tuning is the most common form of transfer learning due to ImageNet pre-trained models' effectiveness in capturing high-level features. 
 To be specific, we refer to fine-tuning (heterogeneous transfer learning), where the source and target dataset have different tasks, and the features are transferred via the weights of a model trained on the source dataset.
 Other forms of transfer learning are outside the scope of this work, such as domain adaption~\cite{wang2018deep}. 
 Without transfer learning, more labelled images are needed during training. Therefore, transfer learning via fine-tuning is a good strategy, where labeled data is sparse and limited~\cite{huh2016makes}. 
 
 \section{Metric Learning}\label{rel:metric}
 
 Distance metric learning (DML) is a type of learning that learns an embedding space where similar examples are attracted and dissimilar examples are repelled. DML can be executed in a supervised or unsupervised fashion. Supervised DML can improve the generalization of features and is often applied to image retrieval, facial recognition, and object re-identification tasks~\cite{wang2019multi,Schroff_2015_CVPR,song2016deep,liu2016deepfashion}.
 
 \noindent Zero-shot image retrieval is a task where images of unseen classes are retrieved. During training, a model is trained with a set of images with known classes, and the goal is to cluster images of unknown classes.
Zero-shot image retrieval is an important problem because it examines the generalizability of a model on unknown classes. Distance metric learning (DML) is commonly used to solve zero-shot image retrieval task. 
Where labeled data is scarce, DML improves the generalization of a model without additional labeled data~\cite{liu2019large}. 

\section{Semi-supervised Learning}\label{rel:semi}

Semi-supervised learning is a fundamental topic in machine learning. It allows models to learn useful representations with a few human labels, abling them to reduce the reliance on manual data annotation. 
This property is advantageous when labeled data is limited but unlabeled data is abundant.  
One common way of handling semi-supervised learning is to minimize consistency loss between two outputs of the same image via various augmentations. For instance, training a model with consistency loss via input and feature perturbations was shown to be an effective strategy for semi-supervised classification~\cite{sajjadi2016regularization}. Additionally, temporal ensembling provided a new dimension for feature comparison yielding a performance gain when coupled with input and feature perturbations~\cite{laine2016temporal}. With temporal ensembling, a model compares its features computed in the current iteration with their corresponding temporal ensembled features, which are the exponential moving averages of features computed from previous iterations. 
A mean teacher model was further proposed by computing the model's weights via an exponential moving average of the model's weight from previous iterations \cite{li2020unsupervised}. Consistency loss via a mean teacher model yielded a higher performance compared to the temporal ensembling method. 
In short, there are various ways to compute consistency loss for semi-supervised classification. 

Another way of tackling semi-supervised learning is via data interpolation. 
MixUp was proposed to interpolate between two labeled images via linear combinations of both the input and label space~\cite{zhang2018mixup}. The distribution of the composite images and labels follows a beta distribution which is heavily weighted towards one or the other images and their corresponding labels. MixMatch further extends MixUp to unlabeled images by interpolating between labeled images and unlabeled images~\cite{berthelot2019mixmatch}. In MixMatch, label sharpening was proposed to improve the predictions of unlabeled images by taking the average predictions of perturbed unlabeled images. 
Additionally, ReMixMatch was proposed to further improve the predictions of unlabeled images by aligning the marginal distribution of unlabeled images and labeled images \cite{li2020unsupervised}. 
In summary, a better prediction of unlabeled data will yield a more accurate data interpolation between labeled and unlabeled data resulting a boost in performance for semi-supervised classification. 

Recently, there has been a resurgence of using self-training for semi-supervised learning. 
Self-training or pseudo-labeling is a classic semi-supervised learning method that uses pseudo-labels to guide the process of learning. Pseudo-labels are defined as labels computed by a model. A major benefit of self-training is that it allows easy extension from an existing supervised model without discarding any information. In 1996, self-training was used to solve semi-supervised sentence classification by iteratively adding unlabeled data to labeled data via pseudo-labeling~\cite{yarowsky1995unsupervised}. 
The introduction of joint loss between labeled and unlabeled data via pseudo-labels popularized self-training in the deep learning literature~\cite{lee2013pseudo}. During training, pseudo-labels were gradually integrated into the joint loss by increasing the weight of unlabeled data via an iteration-dependent co-efficient.  After the re-appearance of self-training, it has gained traction in recent years. Mixture of all models (MOAM) was shown to be an effective way to tackle semi-supervised classification via the integration of unsupervised learning and self-training training~\cite{zhai2019s4l}. 
Additionally, self-training was shown to improve a randomly initialized model outperforming a pre-trained model on object detection~\cite{zoph2020rethinking}. 
Self-training with noisy pseudo-labels via random augmentations improves a classification model's accuracy and robustness~\cite{xie2020self}. Self-training via pseudo-labels ensembling also surpasses a supervised model in a omni-supervised learning environment that includes out-of-distribution samples during training~\cite{radosavovic2018data}.  
In brief, self-training is a competitive alternative to tackle semi-supervised learning. 

\clearpage

\chapter{Learning with Less data via weakly labeled patch classification in Digital Pathology}\label{isbi2020}


\section{Prologue}
This work has been accepted to The International Symposium on Biomedical Imaging (2020). Eu Wern Teh is the first author of this publication, and Graham W. Taylor is the co-author. 

\section{Abstract}
In Digital Pathology (DP), labeled data is generally very scarce due to the requirement that medical experts provide annotations. We address this issue by learning transferable features from weakly labeled data, which are collected from various parts of the body and are organized by non-medical experts. In this paper, we show that features learned from such weakly labeled datasets are indeed transferable and allow us to achieve highly competitive patch classification results on the colorectal cancer (CRC) dataset~\cite{kather2016multi} and the PatchCamelyon (PCam) dataset~\cite{veeling2018rotation} while using an order of magnitude less labeled data.

\section{Introduction}

\begin{figure}[H]
\centering
{\setlength{\tabcolsep}{2pt}
\scalebox{1.3}{
\fbox{\includegraphics[width=3.5in]{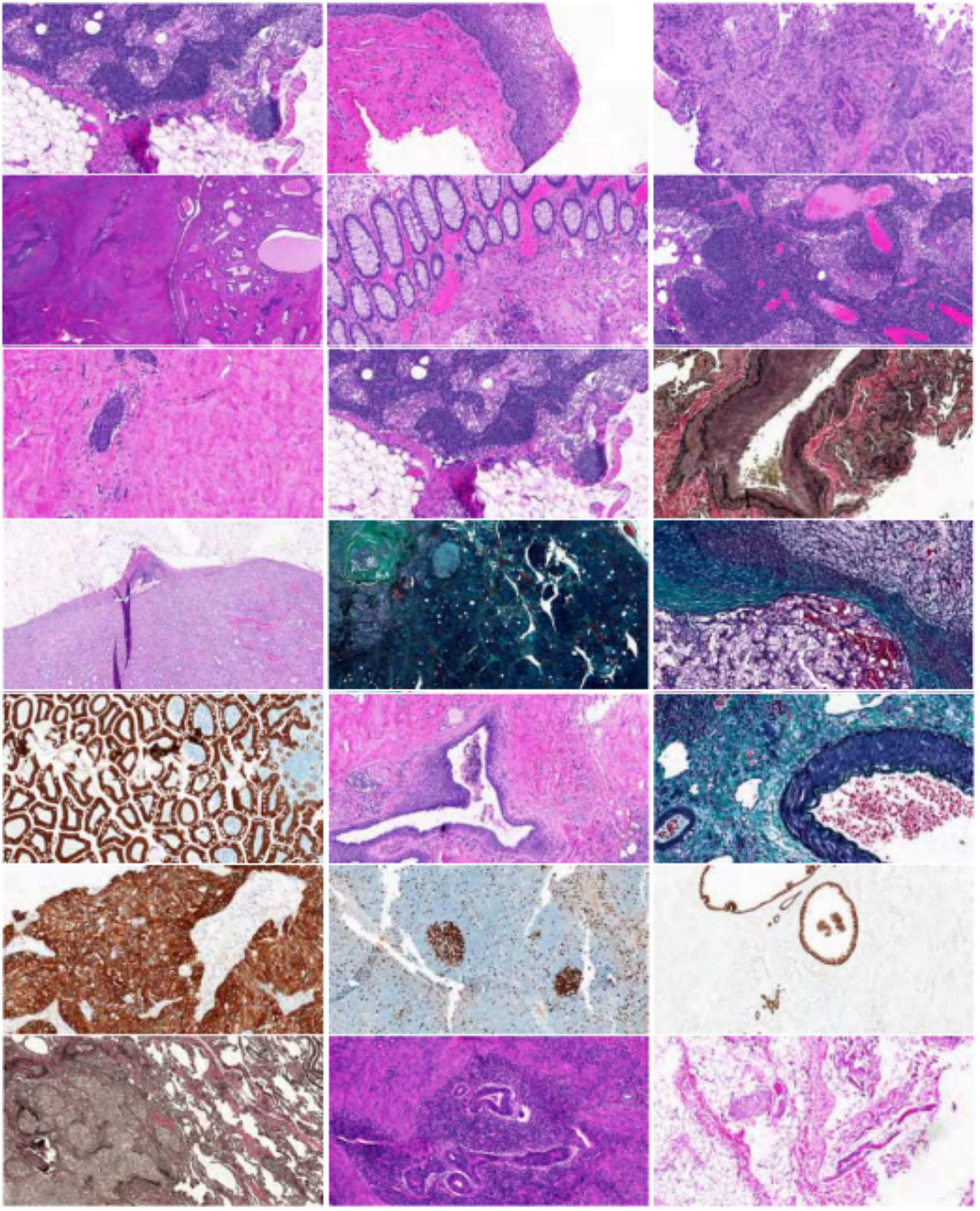}}
}
}
\caption{This figure shows all 24 whole slide images used to generate the KimiaPath24 dataset\cite{babaie2017classification}.
}
\label{fig:kimia24}
\end{figure}

In recent years, we have witnessed significant success in visual recognition for Digital Pathology~\cite{bejnordi2017diagnostic, aresta2019bach}. The main driving force of this success is due to the availability of labeled data by medical experts. However, creating such datasets is costly in terms of time and money.
Conversely, it is relatively easy to obtain weakly labeled images as it does not require annotation from medical experts. One example of such a dataset is KimiaPath24~\cite{babaie2017classification} (see Figure~\ref{fig:kimia24}). These images are collected from different parts of the body by non-medical experts based on visual similarity and are organized into 24 tissue groups.

In this work, our goal is to transfer knowledge from weakly labeled data to a dataset where labeled data is scarce.
We compare three approaches. As a baseline, we train on the target dataset from scratch, i.e., from a randomly initialized model. We contrast this with models pre-trained on weakly labeled data by using two strategies:  First, we learn features via classification of groups with a standard cross-entropy loss. Second, we explore metric learning as an alternative method for feature learning.
In order to evaluate our method, we simulate data scarcity by varying the amount of data available from the richly annotated CRC~\cite{kather2016multi} and PCam~\cite{veeling2018rotation} datasets.

Our contributions are as follows. First, we show that features learned from weakly labeled data are indeed useful for training models on other histology datasets.
Second, we show that with weakly labeled data, one can use an order of magnitude less data to achieve competitive patch classification results rivaling models trained from scratch with 100\% of the data.
Third, we further explore a proxy-based metric learning approach to learn features on the weakly labeled dataset. 
Fourth, we show that transfer learning is more effective with features learned 
from class-labeled images in the Digital Pathology domain (Kimia24) than class-labeled images in the natural image domain (ImageNet).
Finally, we also achieve the state-of-the-art results for both CRC and PCam when our models are trained with both weakly labeled data and 100\% of the original annotated data.\looseness=-1



\section{Related Work}

Recently in the field of DP, there are a few works that try to perform transfer learning from one dataset to another. Khan et al.~\cite{khan2019improving} attempt to improve prostate cancer detection by pre-training on a breast cancer dataset. Medela et al.~\cite{medela2019few} perform few-shot learning for lung, breast, and colon tissues by pre-training on the CRC dataset. These works demonstrate that there are some transferable features among various organs. However, the source domain on which they pre-train the model are still annotated by medical experts.

Apart from using classification to learn features from weakly labeled data, we explore the use of metric learning as an alternative approach. Metric Learning is a machine learning task where we learn a function to capture similarity or dissimilarity between inputs.
Metric Learning has many applications such as person re-identification~\cite{luo2019bag}, product retrieval~\cite{song2016deep}, clothing  retrieval~\cite{liu2016deepfashion} and face recognition~\cite{wen2016discriminative}. In Digital Pathology, Medela et al.~\cite{medela2019few} use metric learning in the form of Siamese networks for few-shot learning tasks. Teh et al.~\cite{teh:MIDLAbstract2019a} also use Siamese networks to boost patch classification performance in conjunction with cross-entropy loss. In Siamese networks, pairs of images are fed to the network, and the network attracts or repels these images based on some established notion of similarity (e.g., class label). A shortcoming for Siamese networks is its sampling process, which grows quadratically with respect to the number of examples in the dataset \cite{movshovitz2017no, wu2017sampling,wang2019multi}. Due to this shortcoming, we choose to explore ProxyNCA~\cite{movshovitz2017no}, which offers faster convergence and better performance.

\section{Description of datasets}

To validate our hypothesis, we use one weakly labeled dataset and two target datasets where patch classification is the targeted task. First, we pre-train two different models (Classification and ProxyNCA) on the weakly labeled dataset. After pre-training, we train and evaluate our models on the target datasets.

\subsection{Weakly labeled dataset}

We use the KimiaPath24 dataset \cite{babaie2017classification} as our weakly labeled dataset, where we learn transferable features. A medical non-expert has collected this data from various parts of the body and has organized them into 24 groups based on visual distinction. These images have a resolution of \SI{0.5}{\micro\metre} per pixel with a size of $1000 \times 1000$ pixels. There are a total of 23,916 images in this dataset.\\

\subsection{Target Dataset A: Colorectal Cancer (CRC) Dataset}
The CRC dataset\cite{kather2016multi} consists of seven types of cell textures: tumor epithelium, simple stroma, complex stroma, immune cells, debris and mucus, mucosal glands, adipose tissue, and background. There are 625 images per class, and each image has a resolution of \SI{0.49}{\micro\metre} per pixel with dimensions $150 \times 150$ pixels.

\begin{figure}[H]
\centering
{\setlength{\tabcolsep}{2pt}
\scalebox{1.30}{
\fbox{
\begin{tabular}{cccc}
\includegraphics[width=1.0in,height=1.0in]{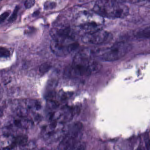}&
\includegraphics[width=1.0in,height=1.0in]{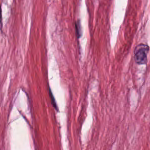}&
\includegraphics[width=1.0in,height=1.0in]{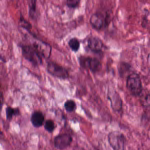}&
\includegraphics[width=1.0in,height=1.0in]{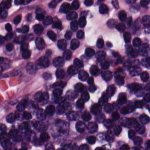} \\
Tumour Epithelium & Simple Stroma & Complex Stroma & Immune Cells\\
 & & &\\
\includegraphics[width=1.0in,height=1.0in]{fig_isbi2020/colon/05.png}&
\includegraphics[width=1.0in,height=1.0in]{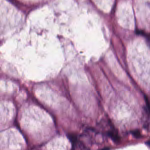}&
\includegraphics[width=1.0in,height=1.0in]{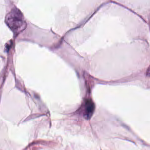}&
\includegraphics[width=1.0in,height=1.0in]{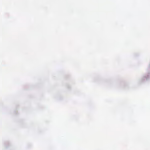}\\
Debris and Mucus & Mucosal Glands & Adipose Tissue & Background\\
\end{tabular}
}
}
}
\caption{Examples of tissue samples from the CRC dataset.}
\label{fig:colon}
\end{figure}

\begin{figure}[H]
\centering
{\setlength{\tabcolsep}{2pt}
\scalebox{1.30}{
\fbox{
\begin{tabular}{ccccc}
\parbox[t]{3mm}{\multirow{1}{*}{\rotatebox[x=-10in]{90}{\vspace{10pt}\textbf{Normal}\hspace{-40pt}}}}&
\includegraphics[width=1.0in,height=1.0in]{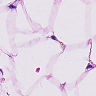}&
\includegraphics[width=1.0in,height=1.0in]{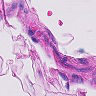}&
\includegraphics[width=1.0in,height=1.0in]{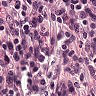}&
\includegraphics[width=1.0in,height=1.0in]{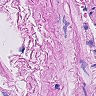} \\
\parbox[t]{3mm}{\multirow{1}{*}{\rotatebox[x=-10in]{90}{\vspace{10pt}\textbf{Tumor}\hspace{-40pt}}}}&
\includegraphics[width=1.0in,height=1.0in]{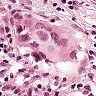}&
\includegraphics[width=1.0in,height=1.0in]{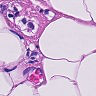}&
\includegraphics[width=1.0in,height=1.0in]{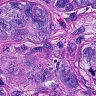}&
\includegraphics[width=1.0in,height=1.0in]{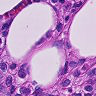}\\
\end{tabular}
}
}
}
\caption{Examples of tissue samples from the PCam dataset.
}

\label{fig:pcam}
\end{figure}

\subsection{Target Dataset B: Patch Camelyon (PCam) Dataset}
PCam~\cite{veeling2018rotation} is a subset of the CAMELYON16 dataset~\cite{bejnordi2017diagnostic}, which is a breast cancer dataset. There are a total of 327,680 images with a resolution of \SI{0.97}{\micro\metre} per pixel and dimensions $96 \times 96$ pixels. There are two categories presented in this dataset: tumor and normal, where an equal number of patches are present in both categories.

\section{Deep metric learning}

We propose to use ProxyNCA (Proxy-Neighborhood Component Analysis)~\cite{movshovitz2017no}, which is a proxy-based metric learning technique.
One benefit of ProxyNCA over standard Siamese Networks is a reduction in the number of compared examples, which is achieved by comparing examples against the class proxies.

\begin{figure}[H]
\centering
{\setlength{\tabcolsep}{2pt}
\scalebox{1.10}{
\begin{tabular}{cccc}
\fbox{\includegraphics[width=2.5in,height=1.3in]{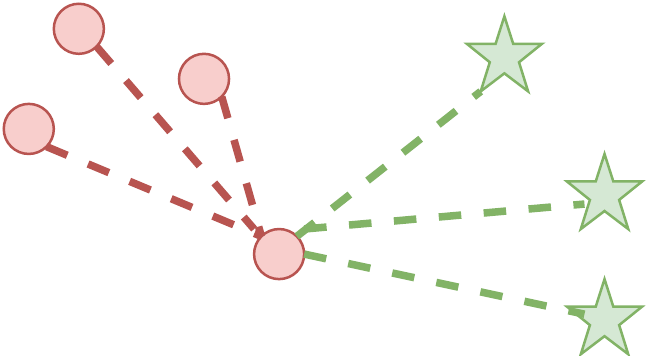}}&
\fbox{\includegraphics[width=2.5in,height=1.3in]{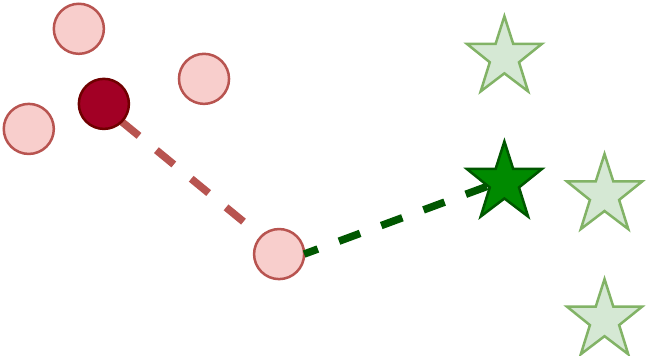}}&\\
\end{tabular}
}
}
\caption{A visualization how ProxyNCA works. [Left panel] Standard NCA compares one example with respect to all other examples (8 different pairings). [Right panel] In ProxyNCA, we only compare to the class proxies (2 different pairings) The above images are reproduced from~\cite{movshovitz2017no}.
}
\label{fig:pnca}
\end{figure}

Figure~\ref{fig:pnca} visualizes the computational savings that can be gained by comparing examples to class proxies rather than other examples. In ProxyNCA, class proxies are stored as parameters with the same dimension as the input embedding space and are updated by minimizing the proxy loss $l$.

ProxyNCA requires an input example $x_i$, a backbone Model $M$, an embedding Layer $E$, a corresponding embedding $\alpha_i$, a proxy function $p(\alpha_i)$, which returns a proxy of the same class as $\alpha_i$, and a proxy function $z(\alpha_i)$, which returns all proxies of different class than $\alpha_i$. Its goal is to minimize the distance between $\alpha_i$ and $p(\alpha_i)$ and at the same time maximize the distance between $\alpha_i$ and $z(\alpha_i)$. Let $d(a,b)$ denote Euclidean distance between $a$ and $b$, ${||a||}_2$ be the $L^2$-Norm of vector $a$ and $\lambda$ be the learning rate. We describe how to train ProxyNCA in Algorithm \ref{alg:proxynca}.

\begin{algorithm}
\caption{ProxyNCA Training}\label{alg:proxynca}
\begin{algorithmic}[1]
\STATE Randomly initialize all proxies
\FOR{$j = 1 \ldots $ number of mini-batches}
    \FOR{$i = 1 \ldots $ mini-batch size}
        \STATE $\alpha_i = E(M(x_i))$
        \STATE $l = - \log \left(\frac{\exp^{-d\left(\frac{\alpha_i}{{||\alpha_i||}_2}, \frac{p\left(\alpha_i\right)}{{||p\left(\alpha_i\right)||}_2}\right)}}{\sum_{z \in Z(\alpha_i)} \exp^{-d\left(\frac{\alpha_i}{{||\alpha_i||}_2}, \frac{z}{{||z||}_2}\right)}}\right)$
        \STATE $\theta = \theta - \lambda \frac{\partial l}{\partial \theta}$
    \ENDFOR
\ENDFOR
\end{algorithmic}
\end{algorithm}

After training on weakly labeled data, we discard everything except for the backbone model. We then use this backbone model along with a new embedding layer to train on the target dataset.

\section{Experiments}\label{sec:experiment}

In this section, we describe our experimental setup and results. We evaluate our models using prediction accuracy. For the colon cancer dataset, we follow the same experimental setup as \cite{kather2016multi} where we train and evaluate our models using 10-fold cross-validation. For the PCam dataset~\cite{veeling2018rotation}, we train our model using images in the training set and evaluate our performance on the test set following the setup from \cite{veeling2018rotation} and \cite{teh:MIDLAbstract2019a} where we report on different amounts ($R\%$) of labeled target data used in fine-tuning. $R\%$ of data is sampled evenly for all classes in the target dataset.

\subsection{Experimental Setup}

In all of our experiments, we use the Adam~\cite{kingma2014adam} optimizer to train our model. Following \cite{movshovitz2017no}, we use an exponential learning rate decay schedule with a factor of 0.94. We also perform channel-wise data normalization with the mean and standard deviation of the respective dataset.  \\

\noindent{\textbf{Weakly labeled dataset:}}
We train our weakly labeled dataset similarly for both target datasets with one subtle difference. For the CRC dataset, we perform random cropping of size $150 \times 150$ pixels directly on KimiaPath24 images since the resolution of both datasets are about the same. However, for the PCam dataset, we resize the KimiaPath24 images from $1000 \times 1000$ pixels to $514 \times 514$ pixels to match the resolution of the target dataset. Finally, we perform random cropping of size $96 \times 96$ pixels on the resized images. The initial learning rate is set to $1e^{-4}$, and we train our models for a fixed 100 epochs.\\

\noindent{\textbf{Target dataset:}} During training, we pad the images by 12.5\% via image reflection prior to random cropping to preserve the image resolution. We set the learning rate to $1e^{-4}$ and train all of our models with a fixed number of epochs (200 for the CRC dataset and 100 for the PCam dataset). We repeat the experiment ten times with different random seeds and report the mean over trials in Table~\ref{table:exp_1} and \ref{table:exp_2}. For the experiments that involve pre-training (Classification and ProxyNCA), we initialize our model with weights pre-trained on the weakly labeled dataset.   \\

\noindent{\textbf{Data augmentation:}} In addition to random cropping, we also perform the following data augmentation at each training stage: a) Random Horizontal Flip b) Random Rotation c) Color Jittering - where we set the hue, saturation, brightness and contrast threshold to 0.4.  All data transformations are implemented via the TorchVision package. \\

\noindent{\textbf{Backbone Model:}} In all of our experiments, we use a modified version of ResNet34~\cite{he2016deep}. Due to low resolution of the target datasets ($150 \times 150$ px and $96 \times 96$ px), we remove the max-pooling layer from ResNet34. \\

\noindent{\textbf{Additional Info:}} For the experiment with 100\% of training data on the PCam dataset, we follow the same experimental setup as \cite{teh:MIDLAbstract2019a} where the number of training epochs is ten and learning rate is reduced by a factor of ten after epoch five.

\subsection{Results}
Table~\ref{table:exp_1} and \ref{table:exp_2} show our experimental results. We train our model by varying the amount of target domain data on four different pre-training settings (Random, ImageNet, Classification and ProxyNCA). With the ImageNet setting, we initialize the model with weights pre-trained on the ImageNet 2012 dataset~\cite{imagenet_cvpr09}.

\begin{table}[h]
\centering
\setlength{\tabcolsep}{32pt}
\begin{tabular}{|cccc|} \hline
$N_c$ & $R$\% & Initialization/Method & Accuracy (\%) \\ \hline
12 & 2 & Random & 61.62 $\pm$ 3.79\\
  &  & ImageNet & 79.16 $\pm$ 2.83\\
  &  & Classification & \textbf{83.30 $\pm$ 2.88}\\
  &  & ProxyNCA & 82.82 $\pm$ 2.70\\  \hline
25 & 4 & Random & 64.12 $\pm$ 3.73\\
 &  & ImageNet & 83.42 $\pm$ 1.57\\
 &  & Classification & \textbf{87.30 $\pm$ 1.40}\\
 &  & ProxyNCA & 87.10 $\pm$ 1.23\\  \hline
50 & 8 & Random & 77.34 $\pm$ 2.11\\
 &  & ImageNet & 88.06 $\pm$ 1.73\\
 &  & Classification & 89.38 $\pm$ 1.29\\
 &  & ProxyNCA & \textbf{89.80 $\pm$ 1.88}\\  \hline
100 & 16 & Random & 84.58 $\pm$ 1.19\\
 &  & ImageNet & 90.08 $\pm$ 1.72\\
 &  & Classification & 91.58 $\pm$ 1.08\\
 &  & ProxyNCA & \textbf{91.96 $\pm$ 0.99}\\ \hline
625 & 100 & Random & 89.84 $\pm$ 1.29\\
 &  & ImageNet & 91.62 $\pm$ 1.38\\
 &  & Classification & 92.10 $\pm$ 0.80\\
 &  & ProxyNCA & \textbf{92.46 $\pm$ 1.22}\\
 &  & RBF-SVM \cite{kather2016multi}& 87.40 \\
\hline
\end{tabular}
\bigskip
    \caption{Accuracy of our model trained with $R$\% of data in three different pre-trained settings on the CRC dataset. $N_c$ denotes the number of examples per class.  We compare our approach with  the best model in \cite{kather2016multi} which is a RBF-SVM that uses five different concatenated features (Lower-order histogram, Higher-order histogram, Local Binary Patterns, Gray-Level Co-occurrence Matrix and Ensemble of Decision Trees).}
    \label{table:exp_1}
\end{table}

By pre-training our models on weakly labeled data, we achieve test accuracies of $89.80\%$ and $89.77\%$ on the CRC and PCam datasets respectively with an order of magnitude less training data. Both results rival the test accuracies of randomly initialized models~(89.84\% and 88.98\%), which are trained with $100\%$ of the data. With 100\% of training data, our models attain test accuracies of $92.70\%$ and $90.47\%$, outperforming previous state-of-the-art results: $87.40\%$ and $90.36\%$ on CRC and PCam respectively. We note that for PCam, the previous SOTA falls within the error bars of ours.

\begin{table}[h]
\centering
\setlength{\tabcolsep}{32pt}
\begin{tabular}{|cccc|} \hline
$N_c$ & $R$\% & Initialization/Method & Accuracy (\%) \\ \hline
1,000 & 0.76 & Random & 79.37 $\pm$ 1.33 \\
  &  & ImageNet & 83.91 $\pm$ 0.84\\
 & & Classification & 85.29 $\pm$ 1.08\\
 & & ProxyNCA & \textbf{86.69 $\pm$ 0.46}\\  \hline
2,000 & 1.53 & Random & 81.26 $\pm$ 1.47\\
  &  & ImageNet & 86.12 $\pm$ 0.60\\
 & & Classification & 86.55 $\pm$ 1.29\\
 & & ProxyNCA & \textbf{87.38 $\pm$ 0.78}\\  \hline
3,000 & 2.29 & Random & 84.09 $\pm$ 1.24\\
  &  & ImageNet & 86.52 $\pm$ 0.63\\
 &  & Classification & 87.03 $\pm$ 0.62\\
 &  & ProxyNCA & \textbf{87.69 $\pm$ 0.72}\\ \hline
13,107 & 10 & Random & 88.47 $\pm$ 0.59\\
  &  & ImageNet & 87.22 $\pm$ 0.47\\
 &  & Classification & 89.64 $\pm$ 0.39\\
 &  & ProxyNCA & \textbf{89.77 $\pm$ 0.50}\\ \hline
131,072 & 100 & Random & 88.98 $\pm$ 1.06\\
 &  & ImageNet & 88.91 $\pm$ 0.53\\
 &  & Classification & 89.85 $\pm$ 0.64\\
 &  & ProxyNCA & \textbf{90.47 $\pm$ 0.59}\\
 &  & P4M\cite{veeling2018rotation} & 89.97\\
 &  & Pi+\cite{teh:MIDLAbstract2019a} & 90.36 $\pm$ 0.41\\
\hline
\end{tabular}
\bigskip
    \caption{Accuracy of our model trained with $R$\% of data in three different pre-trained settings on the PCam dataset. $N_c$ denotes the number of examples per class. We compare our approach to \cite{veeling2018rotation}, which uses a DenseNet with group convolutional layers and \cite{teh:MIDLAbstract2019a} which uses contrastive and self-perturbation loss together with cross entropy loss.
    }
    \label{table:exp_2}
\end{table}

In addition, we also show that models pre-trained on weakly labeled data outperform models pre-trained on the ImageNet 2012 dataset, especially in the lower data regime. In Table~\ref{table:exp_2}, we observe that the performance of models trained under ImageNet setting underperform the randomly initialized model when $R$ is $10\%$ and $100\%$. We speculate that as we collect more images per class, the advantage of the ImageNet pre-trained model gradually vanishes.
This trend is not surprising due to the vast differences in the domain between two datasets (natural images vs.~Digital Pathology images).

We further observe that ProxyNCA outperforms classification on the PCam dataset. However, on the CRC dataset, this trend persists when the number of classes per sample is 50 or larger.
When the number of samples per class is too small, the results are highly varied, which makes the comparison more difficult.

\section{Conclusion}\label{sec:conclude}

We show that useful features can be learned from weakly labeled data, where the images are collected from different parts of the body by non-medical experts based on visual similarity. We show that such features are transferable to the CRC dataset as well as the PCam dataset and are able to achieve competitive results with an order of magnitude less training data. Although evaluation is facilitated in the simulated ``low data'' regime, our approach holds promise for transfer to digital pathology datasets for which the number of actual annotations by medical experts is very small.

\newpage
\chapter{Learning with less labels in Digital Pathology via Scribble Supervision from Natural Images}\label{isbi2022}

\section{Prologue}
This work has been accepted to The International Symposium on Biomedical Imaging (2022). Eu Wern Teh is the first author of this publication, and Graham W. Taylor is the co-author. 

\section{Abstract}
A critical challenge of training deep learning models in the Digital Pathology (DP) domain is the high annotation cost by medical experts. One way to tackle this issue is via transfer learning from the natural image domain (NI), where the annotation cost is considerably cheaper.
Cross-domain transfer learning from NI to DP is shown to be successful via class labels~\cite{teh2020learning}. One potential weakness of relying on class labels is the lack of spatial information, which can be obtained from spatial labels such as full pixel-wise segmentation labels and scribble labels. We demonstrate that scribble labels from NI domain can boost the performance of DP models on two cancer classification datasets (Patch Camelyon Breast Cancer and Colorectal Cancer dataset). Furthermore, we show that models trained with scribble labels yield the same performance boost as full pixel-wise segmentation labels despite being significantly easier and faster to collect.

\section{Introduction}

\begin{figure}[H]
  
  \centering
  \includegraphics[width=4.30in,   ]{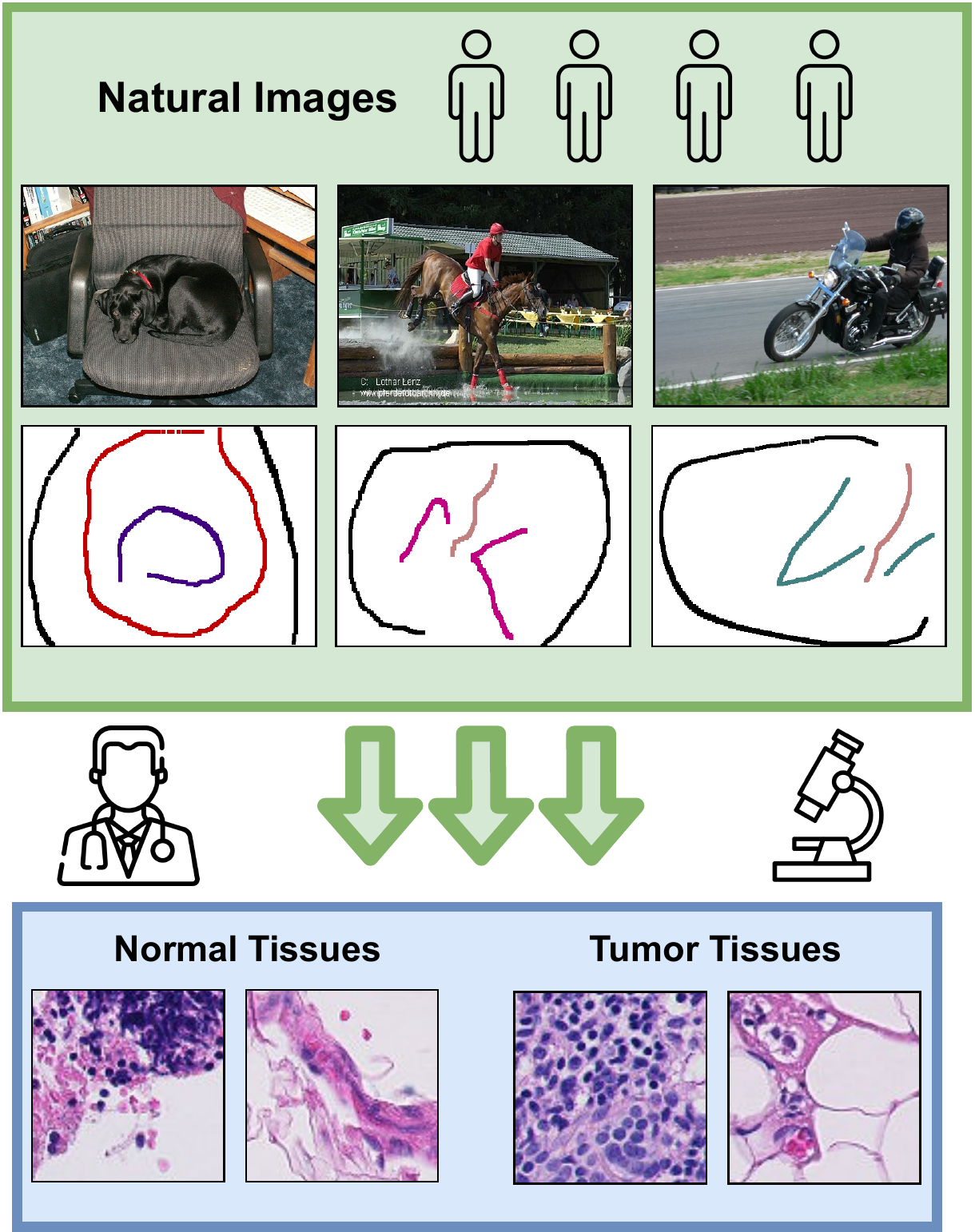}
  \caption{
  A demonstration of our proposed approach in cross-domain transfer learning. We first train a model by using scribble labels from the natural image (NI) domain. We transfer knowledge from the NI domain to the Digital Pathology (DP) domain by initializing the DP models with the pre-trained weights. Lastly, we train these DP models using labels provided by medical experts on cancer classification tasks.
  }
  \label{fig:main2022}
\end{figure}

Digital Pathology (DP) is a field that involves the analysis of microscopic images.
Supervised learning has made progress in DP for cancer classification and segmentation tasks in recent years; however, it requires a large number of labels to be effective~\cite{srinidhi2020deep}.
Unfortunately, labels from medical experts are scarce and extremely costly. On the other hand, it is relatively cheap to obtain labels from a layperson in the natural image (NI) domain.
We leverage inexpensive annotations in the NI domain to help models to perform better in DP tasks (Fig~\ref{fig:main2022}). 
We also investigate the effectiveness of spatial NI labels as they vary by level of detail and thus annotator effort.

Transfer learning via pre-training allows a model to learn features from a source task and transfer them to a target task to improve generalization.~\cite{huh2016makes}.
Features learned from image class labels in the NI domain are shown to be useful in boosting the performance of DP models, especially in the low-data regime~\cite{teh2020learning,kupferschmidt2021strength}.
However, class labels are devoid of spatial information, which can be crucial in transfer learning, and this information can be obtained from spatial labels (e.g., full pixel-wise segmentation labels and scribble labels).
In terms of the level of detail in labels, scribble labels are far weaker than full pixel-wise segmentation labels. However, collecting full pixel-wise segmentation labels is extremely laborious and time-consuming, making it less attractive than scribble labels. Despite the lack of details in scribble labels, we discover that models trained with scribble labels from the NI domain perform equally well in transfer learning compared to models trained with full pixel-wise segmentation labels. This discovery allows us to reliably use scribble annotations, which is significantly faster to obtain than the collection of full pixel-wise segmentation labels.

\section{Related Work}

Deep Learning models are effective in a fully supervised environment, but they require tremendous amounts of human labels~\cite{goodfellow2016deep}. However, human labels are often scarce in the Digital Pathology (DP) domain due to high annotation costs~\cite{srinidhi2020deep}.
Transfer learning in the form of pre-training is an effective way to combat label scarcity~\cite{huh2016makes}. Models are commonly pre-trained with class labels from natural image (NI) datasets~\cite{imagenet_cvpr09,mahajan2018exploring}, and these pre-trained models have helped to boost the performance of many computer vision tasks~\cite{huh2016makes}.

Conventionally, spatial labels are obtained to solve spatial tasks such as semantic segmentation~\cite{chen2017deeplab}, object detection~\cite{Redmon_2017_CVPR}, and human pose estimation~\cite{he2017mask}. Among all spatial labels, collecting full pixel-wise segmentation labels is considered as one of the most time-consuming tasks~\cite{lin2014microsoft}.
Scribble labels were introduced to reduce the reliance on full pixel-wise segmentation~\cite{lin2016scribblesup}. The collection of scribble labels is significantly faster than full pixel-wise segmentation labels. On a low-data regime, joint training of scribble labels and full pixel-wise segmentation labels are shown to be better than models trained only on full pixel-wise segmentation labels. Other works have shown that optimizing graph models can further improve scribble-aided segmentation models~\cite{tang2018regularized,tang2018normalized}.

\section{Methods}

We use a 2D Cross-Entropy Loss as described in Equations~\ref{eq:1_isbi2022} and~\ref{eq:2} to train our models using the full pixel-wise segmentation labels and the scribble labels. Both equations describe the loss for a single image, $x$, and the corresponding spatial mask, $y$, each of dimension $I \times J$, $y_{i,j} \in \{0,1,2,...K\}$. The function, $f$ denotes the segmentation model, $K$ indicates the number of classes present during training, and 0 represents the ignore label.
\begin{align}
L_{i,j}  =  -\frac{1}{I*J}\sum_{i}^{I}\sum_{j}^{J}\log\frac{\text{exp}(f(x_{i,j})[y_{i,j}])}{\sum_{k=1}^{K}\text{exp}(f(x_{i,j})[k])}\label{eq:1_isbi2022}
\end{align}

\begin{align}\label{eq:2}
        L_{i,j}^{'}= \begin{cases}
          L_{i,j}, & \text{if}\; y_{i,j} \neq 0,\\
            0, & \text{if}\; y_{i,j} = 0,
        \end{cases}
\end{align}

\begin{figure}[H]
  \centering
  \includegraphics[width=5.3in, ]{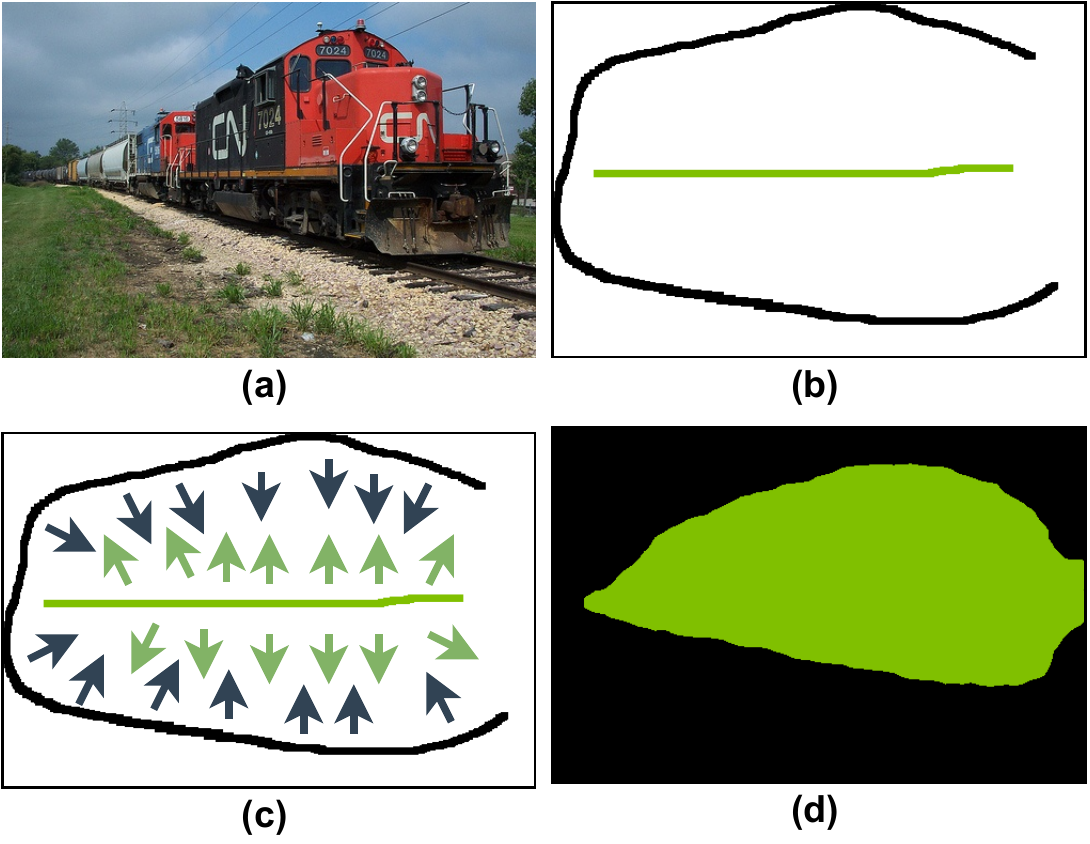}
  \caption{An illustration of how scribble supervised segmentation works. Panels (a) and (b) show the input image and the corresponding scribble mask. Panel (c) depicts the expansion direction of the model's pixel prediction. Panel (d) shows the model's pixel prediction after training.
  }
  \label{fig:voc_scrib}
\end{figure}

The ignore-label (white region in Fig.~\ref{fig:voc_scrib}(b)) plays an important role in scribble-supervised segmentation training.
If we treat the white region as a separate class, the model will not expand its prediction of the real classes. By incurring zero loss in the ignore-label region, the model is free to predict anything. During training, the model's predictions tend to start near the scribble annotation, and they gradually expand outward as training progresses. In Fig.~\ref{fig:voc_scrib}(c), we see that both the background and locomotive classes attempt to fill up the empty ignore-labels region. After training, an equilibrium is reached, and the final prediction is shown in Fig.\ref{fig:voc_scrib}~(d).

\section{Experiments}\label{sec:exp}

We hypothesize that models pre-trained on scribble labels can help DP models achieve higher performance when compared to models pre-trained on class labels. We validate our hypothesis by using one natural image dataset and two digital pathology datasets. We first pre-train our models on the natural image dataset before training on the digital pathology datasets. We report our results by using accuracy as the performance metric on the digital pathology datasets.

\subsection{Pre-training on Pascal VOC 2012 dataset}

\begin{figure}[H]
  \centering
  \includegraphics[width=5.30in, ]{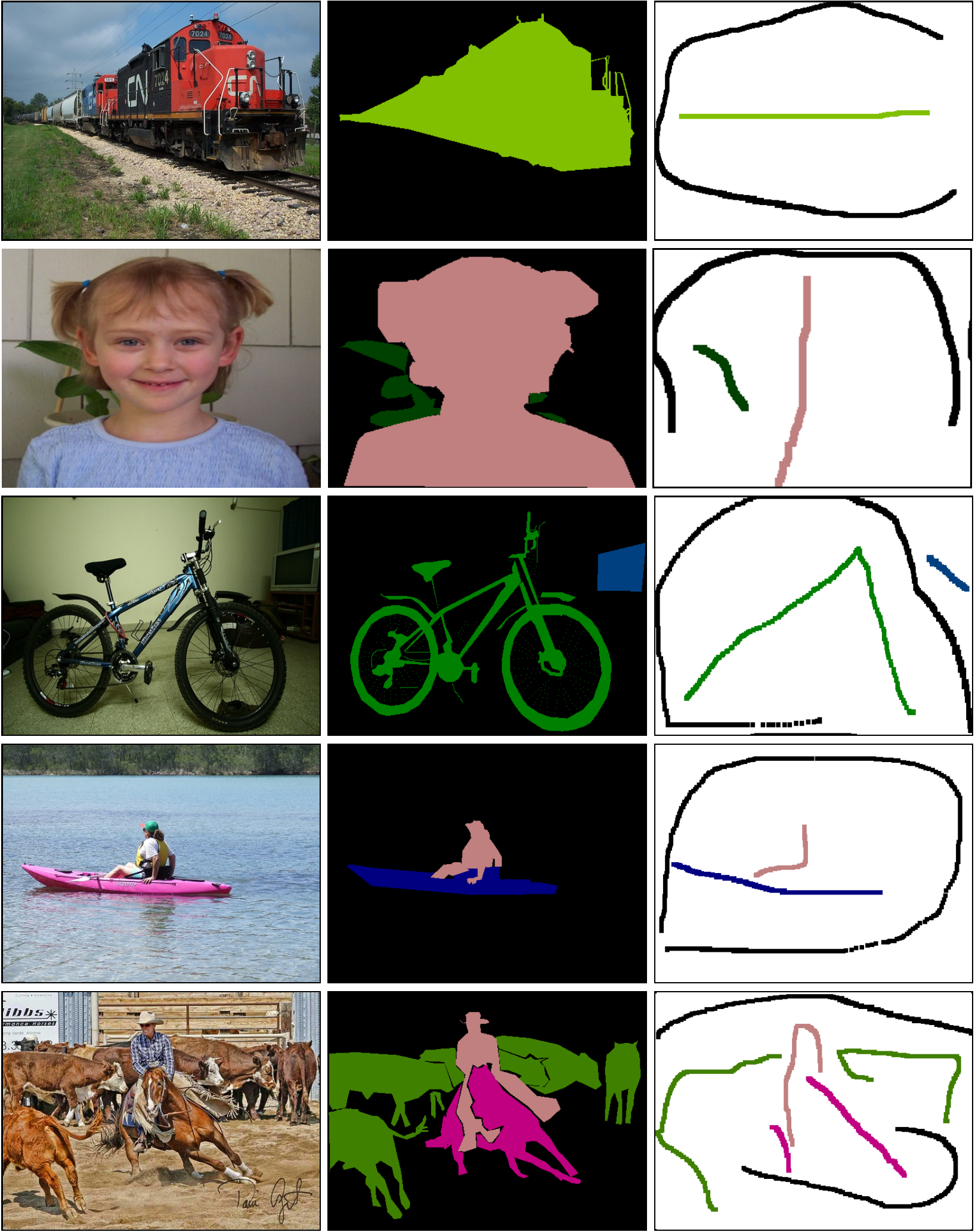}
  \caption{Examples of full pixel-wise segmentation masks and the corresponding scribble masks.
  }
  \label{fig:voc_scrib_sample}
\end{figure}

We use the Pascal VOC 2012 dataset as our natural image dataset~\cite{pascal-voc-2012}. It consists of 21 classes: aeroplane, bicycle, bird, boat, bottle, bus, car, cat, chair, cow, table, dog, horse, motorcycle, person, plant, sheep, couch, locomotive, television, and background (Fig.~\ref{fig:voc_scrib_sample}). It has a total of 10,582 training images and 1,449 validation images. Additionally, we add scribble labels obtained from ~\cite{lin2016scribblesup} to our experiments.

We pre-train our models using three different settings: multi-label classification, full pixel-wise segmentation, and scribble segmentation.
We use a randomly initialized ResNet-34 model to train the multi-label classification model and
a randomly initialized DeepLabV2 model with ResNet-34 backbone for the full pixel-wise segmentation and scribble segmentation pre-training~\cite{he2016deep,chen2017deeplab}. 
For the multi-label classification, we modify the ResNet-34 model to output 21 classes for each image and train the model with a binary cross-entropy loss per output. For the full pixel-wise and scribble segmentation, we modify DeepLabV2 with ResNet-34 backbone to output 21 classes for each pixel, and we penalize the model with a 2D cross-entropy loss.
We train all models for 40,000 iterations using Stochastic Gradient Descent with a learning rate of 2.5e-4, a momentum of 0.9, and a weight decay of 5e-4. We also use a polynomial decay rate of $\lambda(1-\frac{\text{iter}}{\text{max\_iter}})^{0.9}$, where $\lambda$ is the base learning rate~\cite{chen2017deeplab}. We augment the dataset via random cropping ($321\times321$), random horizontal flipping (0.5), and random re-scaling (0.5 to 1.5).
After pre-training, we initialize the downstream models with the corresponding ResNet-34's weights.

\subsection{Training and Evaluation on Digital Pathology downstream tasks}

We evaluate transfer learning performance using two downstream datasets: the Patch Camelyon and the Colorectal cancer dataset following~\cite{teh2020learning,kupferschmidt2021strength}. After pre-training a model on the Pascal VOC 2012 dataset, we initialized the model with the corresponding pre-trained weights and trained on the downstream dataset.
Following \cite{teh2020learning}, we train our downstream tasks with the Adam optimizer with a learning rate of 1e-4 using the exponential learning rate decay scheduler with a factor of 0.94. We train all our models on the PCam dataset for 100 epochs and 200 epochs for the CRC dataset. All downstream training is repeated ten times with random seeds.
Finally, we evaluate the model on the test set of the downstream dataset by using accuracy as a performance metric. The mean accuracy and standard deviation are reported in Table~\ref{table:exp_colon_pcam_isbi2022}.

\begin{figure}[H]
  \centering
  \includegraphics[width=5.30in, ]{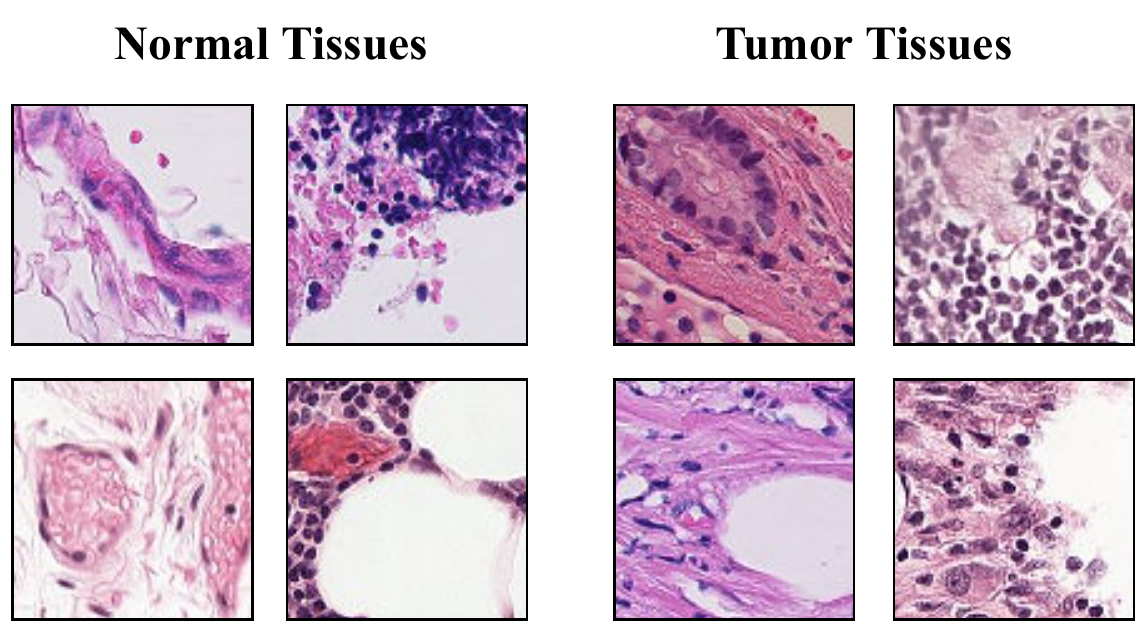}
  \caption{Tissue samples from the Patch Camelyon dataset.}
  \label{fig:pcam_isbi2022}
\end{figure}

The Patch Camelyon dataset (PCam) consists of 262,144 training images and 32,678 test images~\cite{veeling2018rotation}. It is a subset of the CAMELYON16 breast cancer dataset. Each patch has a resolution of 0.97 $\mu$m and a dimension of 96$\times$96 pixels. There are two classes present in this dataset: normal and tumor tissues (Fig.~\ref{fig:pcam_isbi2022}).

\begin{figure}[H]
  \centering
  \includegraphics[width=5.30in, ]{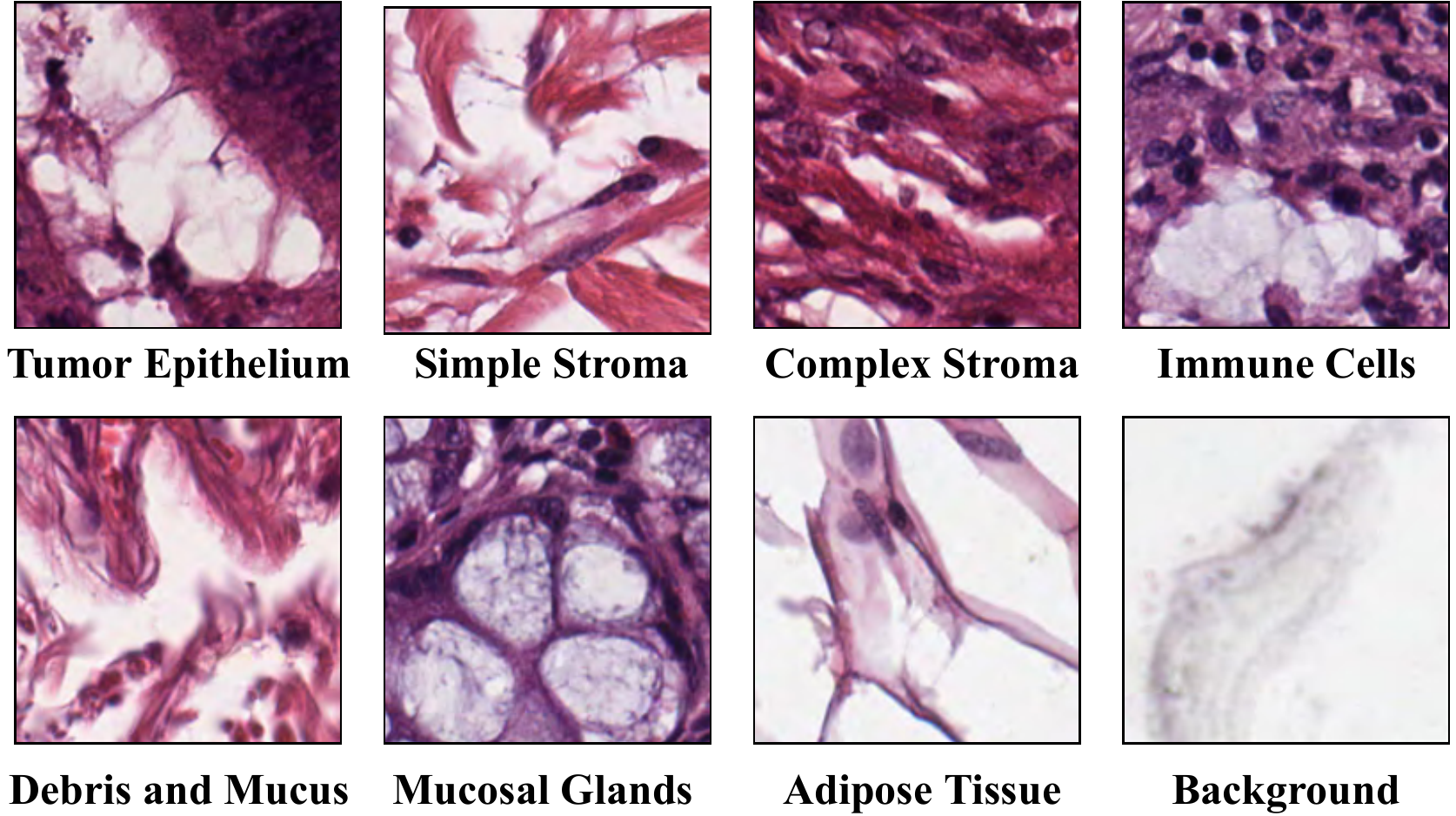}
  \caption{Tissue samples from the Colorectal cancer dataset.}
  \label{fig:crc}
\end{figure}

\begin{table}[H]
\centering
\setlength{\tabcolsep}{10pt}
\begin{tabular}{|l|ccc|ccc|} \hline
&\multicolumn{3}{c|}{CRC dataset}&\multicolumn{3}{c|}{PCam dataset} \\ \hline
 Initialization/Method & $N_c$ & $R$\%  & Accuracy (\%)& $N_c$ & $R$\% & Accuracy (\%) \\ \hline
Random &12 & 2  & 61.62 $\pm$ 3.79 & 1,000 & 0.76 & 79.37 $\pm$ 1.33 \\
Multi-label Classification &  &   & 62.98 $\pm$ 3.70  &  &   & 76.61 $\pm$ 1.83\\
Full Segmentation &  &   & 67.34 $\pm$ 3.80 & &  & 82.11 $\pm$ 1.28\\
Scribble Segmentation &  &   & \textbf{67.40 $\pm$ 3.71} & &  & \textbf{82.13 $\pm$ 1.49}\\\hline
Random & 25 & 4 &  64.12 $\pm$ 3.73 & 2,000 & 1.53  & 81.26 $\pm$ 1.47\\
Multi-label Classification & &  &  69.16 $\pm$ 3.87 &   &   & 80.95 $\pm$ 1.18\\
Full Segmentation & &  & \textbf{73.34 $\pm$ 2.17}&  &  & \textbf{86.11 $\pm$ 1.13}\\
Scribble Segmentation & &  & 72.60 $\pm$ 2.82 &  &  & 86.04 $\pm$ 1.08\\\hline
Random & 50 & 9 &  77.34 $\pm$ 2.11 & 3,000 & 2.29  & 84.09 $\pm$ 1.24\\
Multi-label Classification & &  &  78.00 $\pm$ 1.25 &   &   & 82.50 $\pm$ 1.10\\
Full Segmentation & &  &  \textbf{81.40 $\pm$ 2.05} & &   & \textbf{87.80 $\pm$ 0.67}\\
Scribble Segmentation & &  &  81.02 $\pm$ 2.57 & &   & 87.68 $\pm$ 0.64\\\hline
\end{tabular}
\bigskip
    \caption{Accuracy of our model trained with $R$\% of data in four different pre-trained settings on the CRC dataset and PCam dataset. $N_c$ denotes the number of examples per class. Models for classification and segmentation are pre-trained on the Pascal VOC 2012 dataset~\cite{pascal-voc-2012}. }
    \label{table:exp_colon_pcam_isbi2022}
\end{table}

The Colorectal cancer (CRC) dataset consists of 5,000 images~\cite{kather2016multi}. Each patch has a resolution of 0.49 $\mu$m and a dimension of 150$\times$150 pixels. It consists of eight classes: tumor epithelium, simple stroma, complex stroma, immune cells, debris and mucus, mucosal glands, adipose tissue, and background (Fig.~\ref{fig:crc}). We follow the same experimental setup as~\cite{teh2020learning}, where we train and evaluate our models with 10-fold cross-validation.

\subsection{Discussion}

Table~\ref{table:exp_colon_pcam_isbi2022} reveals that models pre-trained on spatial labels outperform the randomly initialized models and those pre-trained on Pascal class labels. On average, models pre-trained on spatial labels outperform class labels by $3.5\%$ in the Colorectal cancer dataset and  $5.3\%$ in the Patch Camelyon dataset.
These results show that spatial information matters in transfer learning from the NI to DP domain.

Models pre-trained on scribble labels perform equally well when compared to the models pre-trained on full pixel-wise segmentation labels. The average difference of mean and standard deviation between these two settings are 0.23$\%$ and 1.95$\%$ respectively. These results indicate that we can rely on scribble labels for cross-domain transfer learning.

Additionally, models pre-trained on Pascal \emph{class labels} slightly outperform the random baseline on the Colorectal cancer dataset, but they perform poorly on the Patch Camelyon dataset.
We speculate that it takes more images ($>10K$) and more than diversity ($>20$ classes) in the class labels to be useful for transfer learning.

\section{Conclusion}\label{sec:con}

We demonstrate that spatial information matters in transfer learning from the natural image (NI) domain to the Digital Pathology (DP) domain. Surprisingly, on both the CRC and PCam datasets, we showed that full pixel-wise segmentation labels are not needed for pre-training a model on the NI domain, and models pre-trained on scribble labels are sufficient to ensure a performance boost in the DP domain.
\newpage
\chapter{Understanding the impact of image and input resolution on deep digital pathology patch classifiers}\label{midl2022}

\section{Prologue}
This work has been accepted to the Canadian Conference on Computer and Robot Vision (2022). Eu Wern Teh is the first author of this work, and Graham W. Taylor is the co-author.

\section{Abstract}
We consider annotation efficient learning in Digital Pathology (DP), where expert annotations are expensive and thus scarce. We explore the impact of image and input resolution on DP patch classification performance. We use two cancer patch classification datasets PCam and CRC, to validate the results of our study. Our experiments show that patch classification performance can be improved by manipulating both the image and input resolution in annotation-scarce and annotation-rich environments.
 We show a strong positive correlation between the image and input resolution and the patch classification accuracy on both datasets.
By exploiting the image and input resolution, our final models trained on $\textless$ 1\% of data successfully surpass models trained on 100\% of data in the original image resolution on the PCam dataset.

\section{Introduction}

Digital Pathology (DP) is a medical imaging field where microscopic images are analyzed to perform Digital Pathology tasks (e.g., cancer diagnosis). The appearance of hardware for a complete DP system popularized the usage of whole slide image (WSI) \cite{pantanowitz2010digital}. These whole slide images are high-resolution images, usually gigapixel in size. Without resizing WSIs, these high-resolution images will not fit in the memory of off-the-shelf GPUs. On the other hand, resizing WSIs destroys fine-grained information crucial in DP tasks, such as cancer classification. A common practice is to divide WSIs into small patches to address this issue.
Most of the DP solutions for patch classification use single-scale images, where a fixed zoom-level of WSI is used~\cite{veeling2018rotation,kather2019predicting,aresta2019bach,srinidhi2020deep}. Therefore, a standard augmentation strategy such as Random-resized-crop will not work in patch classification  (more details are provided in Appendix~\ref{sec:rrc}).

The Deep Learning field has focused on innovating architectures ranging from AlexNet to ResNet to Transformers \cite{krizhevsky2012imagenet,he2016deep,dosovitskiy2020image}. ``The bigger the model, the better" approach works when a massive quantity of human annotations is available. However, in an annotation-scarce environment, models with higher capacity often perform worse due to overfitting~\cite{goodfellow2016deep}.
We take a step back from designing complex methods and explore image and input resolution factors, which are often ignored as part of an input processing pipeline.
We show that we can improve patch classification with a fixed zoom level by increasing the image and input resolution of image patches in DP. By increasing the image and input resolution, we allow our model to focus on fine-grained information without losing coarse-grained information.

\section{Related work}~\label{sec:rel_midl2022}

\noindent{\textbf{Annotation-efficient learning:}} In the Digital Pathology (DP) domain, the availability of human labels is often scarce due to the high annotation costs \cite{srinidhi2020deep}. This data scarcity is a challenge for supervised deep learning models as these models often require tremendous amounts of human labels to be effective \cite{goodfellow2016deep}. An effective way of combating label scarcity is via transfer learning in the form of pre-training \cite{huh2016makes}.  
Transfer learning from a pre-trained ImageNet model is shown to be effective in the DP domain, especially in the low-data regime~\cite{hegde2019similar,teh2020learning,kupferschmidt2021strength}.
\\

\begin{figure*}[htb]
  \centering
  \includegraphics[width=6.40in,   ]{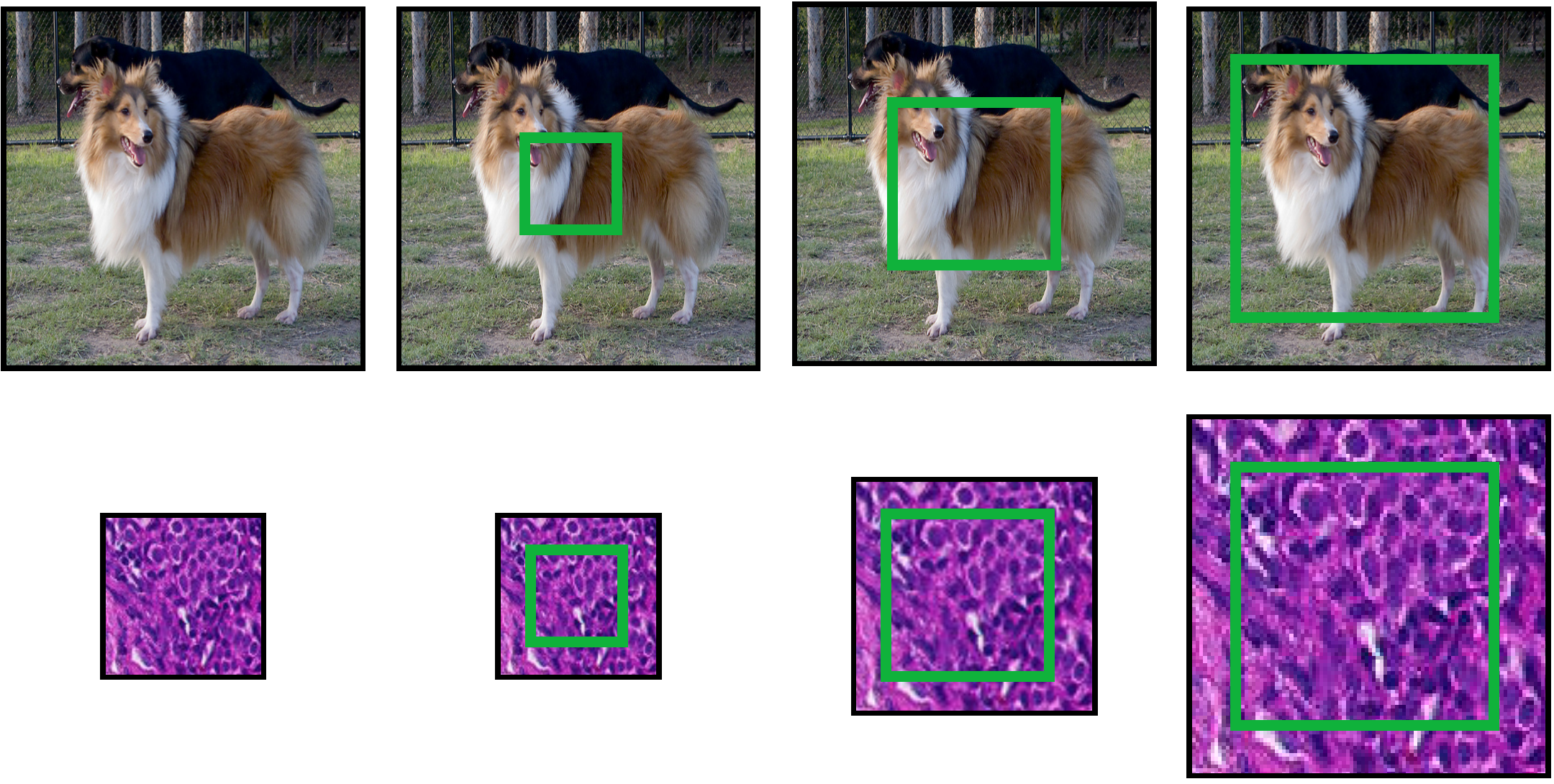}
  \caption{
  An illustration of the differences between input and image resolutions. In both rows of images, input resolution increases from left to right. However, only the bottom row has an increase in image resolution.
  }
  \label{fig:main_diff}
\end{figure*}

\noindent{\textbf{The effect of input resolution on model performance:}} There is an important distinction between image resolution and input resolution (Figure~\ref{fig:main_diff}). Image resolution refers to the width and height of an image in pixels, while input resolution refers to the input width and height being fed to a model. Random-resized-crop augmentation is a standard input augmentation technique used in the Natural Image (NI) domain for image classification~\cite{touvron2019fixing,tan2019efficient,he2016deep}. During Random-resized-crop augmentation, the image resolution remains the same throughout training, while the crop area changes. The crop area is randomly scaled between 0.08 to 1.00 of the image resolution, followed by another random re-scaling between 0.75 to 1.33. Finally, the crop area is resized to the given input resolution.

In the NI domain,  classification performance can be improved by increasing the input resolution (crop size)~\cite{touvron2019fixing,tan2019efficient}.
Touvron et al. discover that different training and testing crop size affects model performance~\cite{touvron2019fixing}. The most optimal train and test crop size for the ImageNet dataset are 384$\times$384 and 448$\times$448. Tang et al. push the limits of their EfficientNet model by increasing the crop size of the images~\cite{tan2019efficient}. Teh et al. show that image crop size has a large influence on image retrieval performance~\cite{teh2020proxynca++}.
It is reasonable to use Random-resized-crop in the NI domain, as NI images come with a wide range of different image resolutions (Figure~\ref{fig:imagenet}). However, in the DP domain, Random-resized-crop is ineffective for patch classification because image patches have the same image resolution (fixed zoom-level)  throughout the dataset (Appendix~\ref{sec:rrc}).

\section{Methodology and Control Factors}


\textbf{Supervised-training:} A ResNet-34 architecture is used as our model in all of our experiments\footnote{With the exception of experiments in Table~\ref{table:exp_colon_pcam_small} where we explore ResNet-152 and ResNeXt-101 architectures \cite{xie2017aggregated}.} where we trained them with cross-entropy loss as described in Equation~\ref{eq:1} \cite{he2016deep}. $B$ denotes the batch size, $x_i$ denotes a single image with a corresponding image label, $y_i$, and $K$ denotes the number of classes in the respective dataset. $f(x_i)$ represents a deep model that accepts an image, $x_i$ and returns a vector of size $K$ denoting the classification scores.

\begin{align}
L  =  -\frac{1}{B}\sum_{i}^{B}\log\frac{\text{exp}(f(x_{i})[y_{i}])}{\sum_{k=1}^{K}\text{exp}(f(x_{i})[k])}\label{eq:1}
\end{align}

\noindent{\textbf{Model initialization:}} In Section~\ref{sec:scarce} and \ref{sec:rich}, we use randomly initialized weights in our experiments. In Section~\ref{sec:limit}, we use both randomly initialized and ImageNet pre-trained ResNet-34 model to explore the limits of annotation-efficient learning for patch classification in DP. \\

\begin{figure*}[htb]
  \centering
  \includegraphics[width=6.40in,   ]{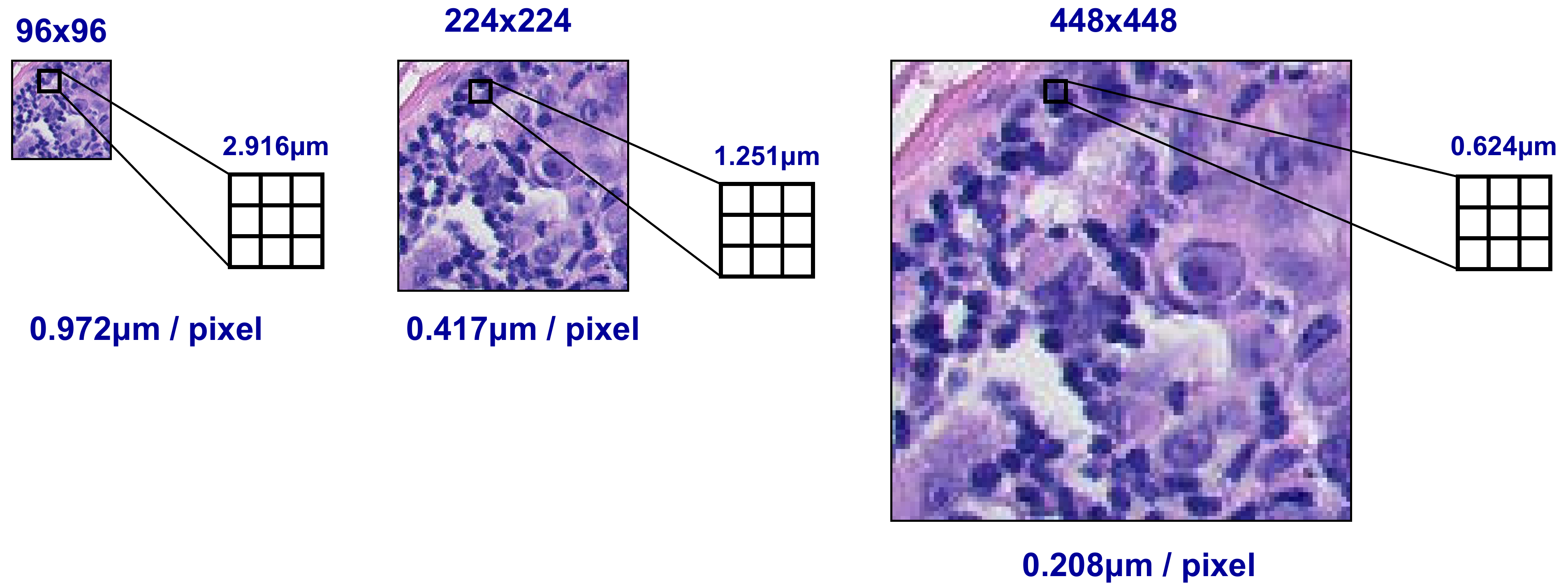}
  \caption{
  An illustration of image resolution and the corresponding receptive fields of a 3$\times$3 convolution layer at various image resolutions. The leftmost image is resized twice (224$\times$224  and 448$\times$448) via bilinear interpolation from an original image at 96$\times$96 resolution with $0.972\mu m$/pixel. As image resolution increases, the receptive field decreases, allowing finer-grained features to be captured by the convolution layer.
  }
  \label{fig:main_midl2022}
\end{figure*}

\noindent{\textbf{Data augmentation:}} We perform data augmentation via random rotation, random cropping, random horizontal flipping, and color jittering following Teh et al.~\cite{teh2020learning} (Algorithm~\ref{alg:code}). Each image is resized with matching height, $h$ and width, $w$ via bilinear interpolation. As image resolution increases, the receptive field of each convolution layer in the model increases, resulting in model awareness towards finer-grained details in the images (Figure~\ref{fig:main_midl2022}). 
Additionally, we also perform experiments without data augmentation in Section~\ref{sec:scarce} (Algorithm~\ref{alg:code_no}). \\

\noindent{\textbf{Dataset demography:}} We use the Patch Camelyon (PCam) and the Colorectal cancer (CRC) dataset in our experiments. For each dataset, we create two subsets: scarce and full. The scarce subset consists of $\textless 1$ \% of the dataset, while the full subset consists of $100\%$ of the dataset.

The PCam dataset consists of 262,144 training images, 32,768 validation images, and 32,768 test images ~\cite{veeling2018rotation}. It comprises two classes: normal tissue and tumor tissue. Each image has a dimension of 96$\times$96 pixels with a resolution of 0.972$\mu$m per pixel. On the scarce subset of PCam, we randomly extract 2000 images from the training set and 2000 images from the validation set. The 2000 validation images are used in both scarce and full experiments to determine early stopping. There is approximately 0.76\% of images in PCam-Scarce training set when compared to the full training set.

The CRC dataset consists of 100,000 images, and we split the dataset into 70\% training set, 15\% validation set, and 15\% test set following ~\cite{kather2019predicting}. It consists of nine classes: adipose tissue, background, debris, lymphocytes, mucus, smooth muscle, normal colon tissue, cancer-associated stroma, and colorectal adenocarcinoma epithelium. Each image has a dimension of 224$\times$224 pixels with a resolution of 0.5$\mu$m per pixel. On the scarce subset of CRC, we randomly extract 693 images (77 images per class) from the training set and 693 images (77 images per class) from the validation set. The 693 validation images are used in both scarce and full experiments to find the best epoch at which to early stop. There is approximately 0.99\% of images in CRC-Scarce training set when compared to the full training set. \\

\noindent{\textbf{Additional Experimental Settings:}} We train all our models on the scarce subset for 100 epochs and 20 epochs for the full subset of the corresponding dataset. We use a validation set to select the best epoch, and the best model is being evaluated with a test set. All models are trained with the Adam optimizer with a learning rate of $1e^{-4}$, a weight decay of $5e^{-3}$, and a batch size of 32. We train all models five times with different seeds and report the mean accuracy with one standard deviation of uncertainty. 

\section{Experiments}

We examine the effects of image and input resolution in three different settings: Annotation-scarce environments (\S~\ref{sec:scarce}), Annotation-rich environments (\S~\ref{sec:rich}), and Transfer Learning settings (\S~\ref{sec:limit}).

\subsection{The effects of image and input resolution on annotation-scarce dataset}~\label{sec:scarce}

\begin{figure*}[htb]
  \centering
  \includegraphics[width=6.40in,   ]{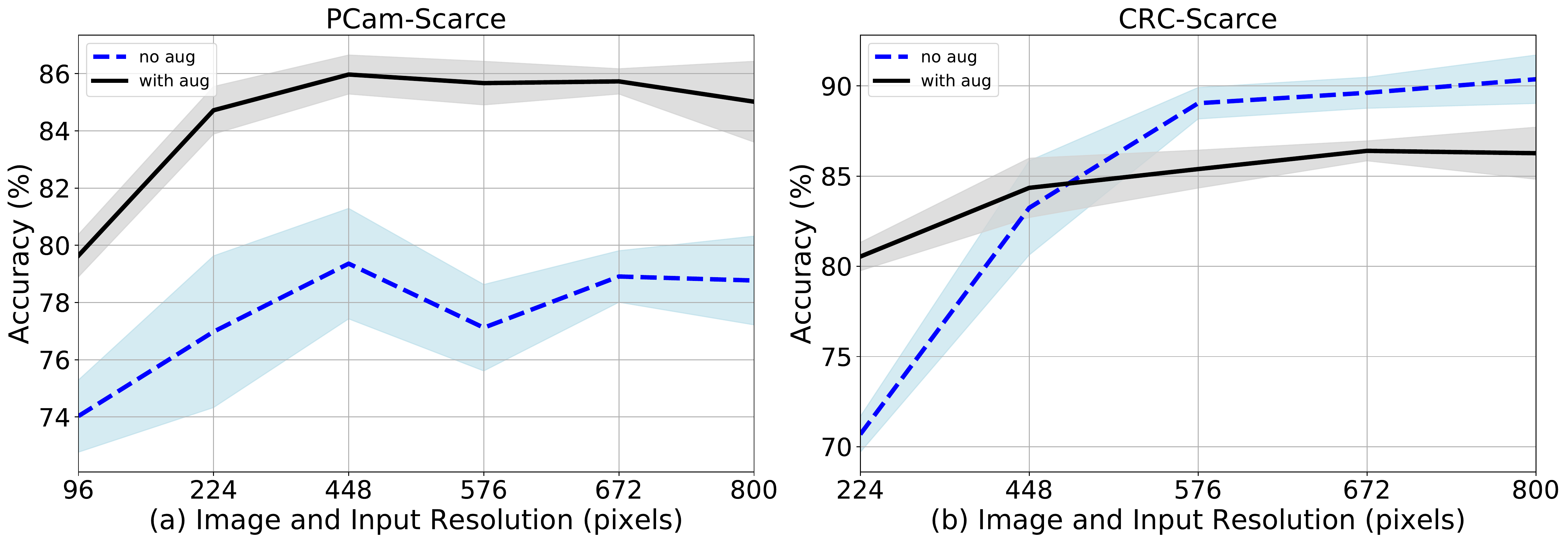}
  \caption{Image and Input resolution effects on models trained with $\textless$ 1\% of the original dataset.
  The shaded areas represent one standard deviation of uncertainty. The dotted line represents models trained without data augmentation.
  }
  \label{fig:pcam_lcrc_1p}
\end{figure*}

We analyze the impact of image and input resolution on two annotation-scarce datasets: PCam-Scarce and CRC-Scarce with and without data augmentation (Figure~\ref{fig:pcam_lcrc_1p}). On models with augmentation, there is a strong positive correlation of 0.703 (PCam-Scarce) and 0.944 (CRC-Scarce) between the image and input resolution and patch accuracy. Similarly, there is also a strong correlation of 0.771 (PCam-Scarce) and 0.938 (CRC-Scarce) on models without augmentation. Models' performance suffers in general without data augmentation on the PCam-Scarce dataset. Nevertheless, on the CRC-Scarce dataset, models without data augmentation surpass models with data augmentation as image and input resolution approach 576$\times$576 and beyond. This result shows that an increase of image and input resolution alone may be sufficient to regularize the network on the CRC-Scarce dataset.

\begin{table}[h]
\centering
\setlength{\tabcolsep}{20pt}
\scalebox{1.0}{
\begin{tabular}{|lcc|c|c|} \hline
\multicolumn{3}{|c|}{}&\multicolumn{1}{c|}{PCam-Scarce}&\multicolumn{1}{c|}{CRC-Scarce} \\ \hline
Models & Res. & GFLOPs & Accuracy (\%) & Accuracy (\%)  \\ \hline
ResNet-34 & 96 &  0.68 & \cellcolor{gray!24}79.64$\pm$0.74 & 72.17$\pm$1.65  \\
ResNet-34 & 224 &  3.68 & 84.72$\pm$0.83 & \cellcolor{gray!24}80.54$\pm$0.78  \\
ResNet-34 & 448 &  14.73 & \textbf{85.97$\pm$0.68} & 84.35$\pm$1.64  \\
ResNet-34 & 576 &  24.35 & 85.67$\pm$0.76 & 85.39$\pm$1.05  \\
ResNet-34 & 672 &  33.14 & 85.73$\pm$0.44 & \textbf{86.40$\pm$0.55}  \\
ResNet-34 & 800 &  46.97 & 85.02$\pm$1.41 & 86.27$\pm$1.44 \\
\hline
ResNet-152 & 96 &  2.14 & \cellcolor{gray!24}77.47$\pm$1.57 & 55.68$\pm$1.00  \\
ResNet-152 & 224 &  11.63 & 79.97$\pm$1.10 & \cellcolor{gray!24}65.34$\pm$1.44 \\
ResNet-152 & 448 &  46.51 & 85.72$\pm$0.57 & 85.37$\pm$0.37 \\
ResNet-152 & 576 &  76.88 & 85.48$\pm$1.32 & 85.55$\pm$0.94 \\
ResNet-152 & 672 &  104.64 & \textbf{85.92$\pm$0.30} & \textbf{86.21$\pm$1.20} \\
ResNet-152 & 800 &  148.30 & 85.19$\pm$0.51 & 85.55$\pm$1.69 \\
\hline
ResNeXt-101 & 96 &  3.04 & \cellcolor{gray!24}79.34$\pm$0.60 & 61.53$\pm$0.89 \\
ResNeXt-101 & 224 &  16.55 & 80.06$\pm$0.77 & \cellcolor{gray!24}71.21$\pm$1.67 \\
ResNeXt-101 & 448 &  66.21 & 85.72$\pm$0.52 & 84.10$\pm$1.30 \\
ResNeXt-101 & 576 &  109.45 & 85.58$\pm$1.21 & \textbf{86.67$\pm$0.71} \\
ResNeXt-101 & 672 &  148.98 & \textbf{86.10$\pm$0.17} & 86.21$\pm$1.12 \\
ResNeXt-101 & 800 &  211.14 & 84.98$\pm$1.09 & 86.30$\pm$1.23 \\
\hline
\end{tabular}
}
    \caption{Patch classification accuracy of our models trained on the PCam-Scarce dataset and CRC-Scarce dataset. All models are trained with data augmentation. Column 2 denotes both the image and input resolution. The shaded cells represent the default configurations on both datasets.}
    \label{table:exp_colon_pcam_small}
\end{table}

Table~\ref{table:exp_colon_pcam_small} shows our experimental results on the PCam-Scarce and CRC-Scarce dataset at various image and input resolutions. In addition to the ResNet-34 architecture, we also experiment with ResNet-152 and ResNeXt-101 (32\texttimes 8d) architectures. ResNet-152 is a direct upgrade of ResNet-34, with more convolution layers. ResNeXt is an enhanced version of the corresponding ResNet architecture with increased parallel residual blocks and more channels  \cite{xie2017aggregated}. ResNeXt-101 (32\texttimes 8d) has 32 parallel residual blocks with 8\texttimes\ more channels compared to the vanilla ResNet-101 architecture. Additionally, we also report the number of floating-point operations (GFLOPS) for each model at various image and input resolutions\footnote{We exclude the final linear classifier in this calculation as it is held constant across architectures.}.

In the low resolution setting (96$\times$96), model performance decreases when larger and more complex architectures (ResNet-152 and ResNeXt-101) are used on the annotation-scarce dataset (Table~\ref{table:exp_colon_pcam_small}). 
On the other hand, increasing the image and input resolution greatly improve performances for all models. 
This result suggests that increasing the image and input resolution may indeed be regularizing high capacity models in the low-annotation setting. However, increasing the image and input resolution also leads to increased computational cost, which grows quadratically with those factors.

\begin{figure*}[htb]
  \centering
  \includegraphics[width=6.40in,   ]{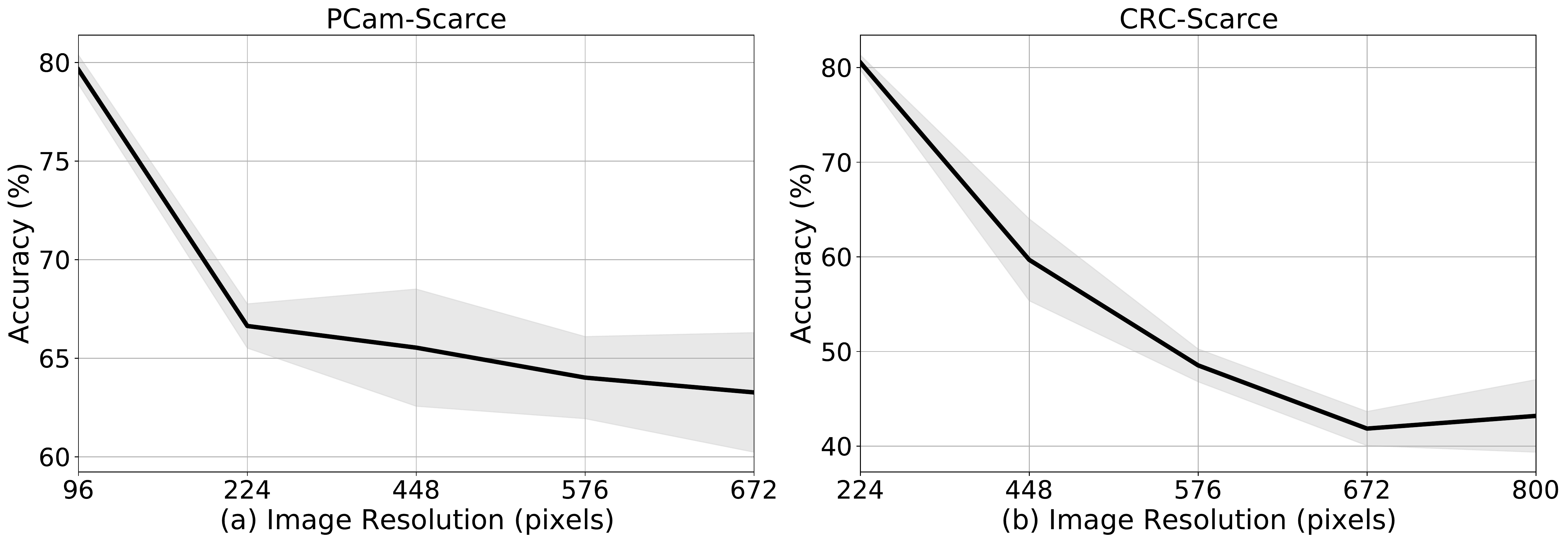}
  \caption{Effect of image resolution on models trained with \textless 1\% of the original dataset. Input resolution is held constant in this experiment. The shaded areas represent one standard deviation of uncertainty. All models train with data augmentation.
  }
  \label{fig:pcam_lcrc_1p_alt}
\end{figure*}

Figure~\ref{fig:pcam_lcrc_1p_alt} shows that varying the image resolution alone is not sufficient. There is a strong negative correlation of 0.835 (Pcam-Scarce) and 0.970 (CRC-Scarce) between image resolution and patch classification accuracy. As the image resolution increases, a model can capture finer-grained information. Nonetheless, if we do not increase the input resolution proportionally, the model loses global information, causing a significant drop in performance. \\

\subsection{The effects of image and input resolution on an annotation-rich dataset}~\label{sec:rich}

\begin{figure*}[htb]
  \centering
  \includegraphics[width=6.40in,   ]{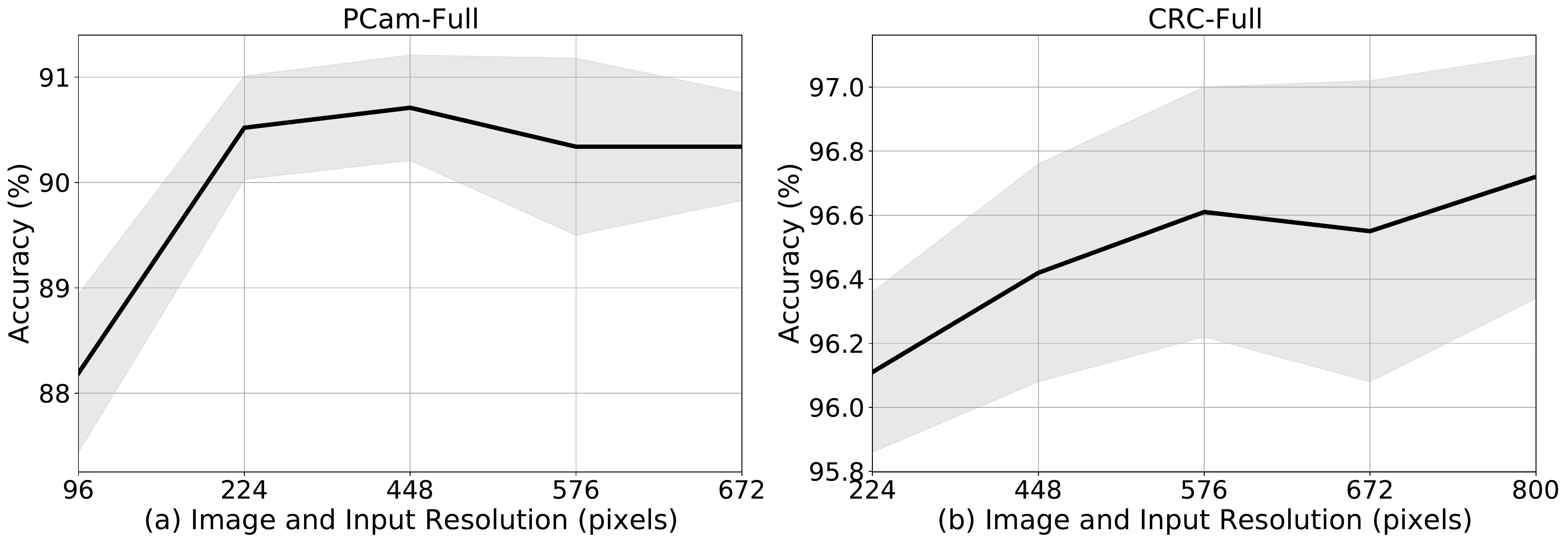}
  \caption{Image and Input resolution effects on models trained with 100\% of the original dataset. The shaded areas represent one standard deviation of uncertainty. All models are trained with data augmentation.
  }
  \label{fig:pcam_lcrc_full}
\end{figure*}

We examine the impact of image and input resolution on two annotation-rich datasets: PCam-Full and CRC-Full.
Figure~\ref{fig:pcam_lcrc_full} shows that as we increase the image and input resolution, the models' performance generally increases.
There is a strong positive correlation of 0.649 (PCam-Full) and 0.961 (CRC-Full) between the image and input resolution and patch classification accuracy.
For the PCam-Full dataset, peak performance occurs at 576$\times$576 resolution. The model achieves peak performance at 800$\times$800 resolution for the CRC-Full dataset. There is a gain of 2.52\% in accuracy for the PCam-Full dataset and a gain of 0.62\% in accuracy for the CRC-Full dataset when compared to the accuracy of models trained with the original image size of the respective datasets (PCam-Full: 96$\times$96, CRC-Full: 224$\times$224). 

\subsection{Annotation-efficient learning in the transfer learning setting}~\label{sec:limit}

\begin{figure*}[htb]
  \centering
  \includegraphics[width=6.40in,   ]{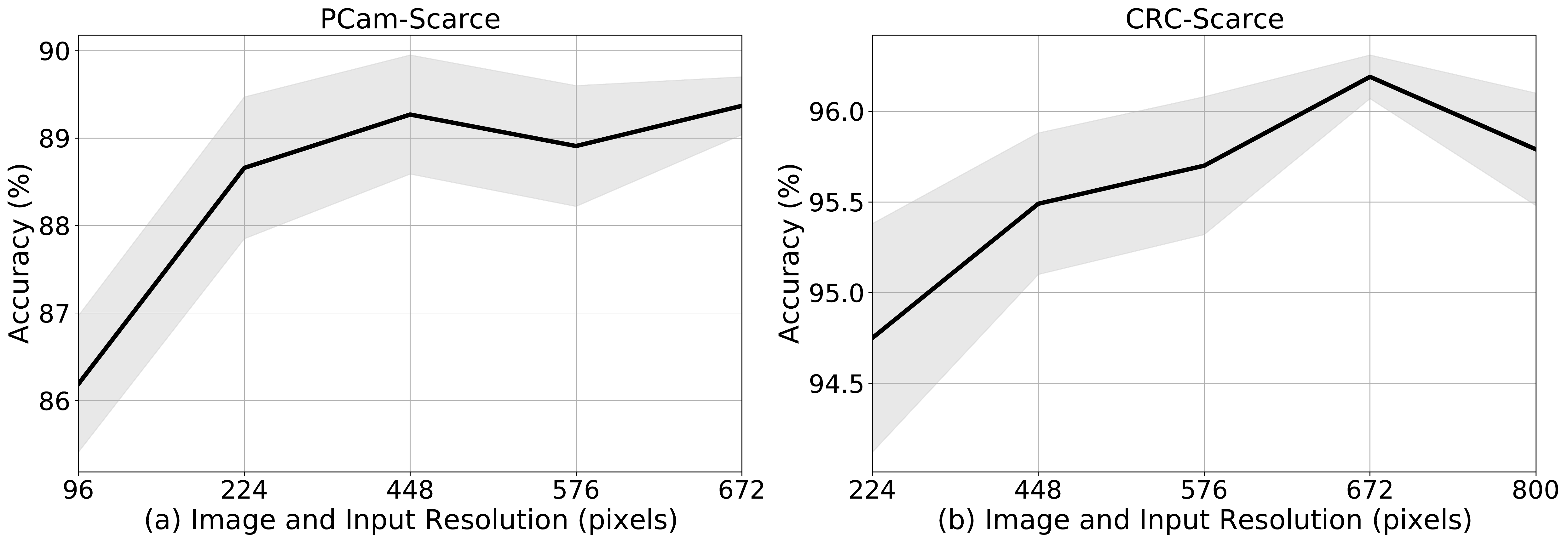}
  \caption{Image and Input resolution effects on models trained with $\textless$ 1\% of the original dataset.
  The shaded areas represent one standard deviation of uncertainty. PCam-Scarce models are trained with data augmentation, but CRC-Scarce models are trained without data augmentation. All models are initialized with ImageNet pre-trained weights.
  }
  \label{fig:pcam_lcrc_1p_imgnet}
\end{figure*}

Transfer learning in the form of pre-training is a common strategy to tackle annotation efficient learning (\S~\ref{sec:rel_midl2022}). In the previous experiments we avoided transfer learning to isolate the contribution of image and input resolution on performance. In this experiment, we study the interaction between image and input resolutions and pre-trained features (Table~\ref{table:exp_colon_pcam} and Figure~\ref{fig:pcam_lcrc_1p_imgnet}). Previously, we observed that on the PCam-Scarce and CRC-Scarce datasets, there is a gain of up to 6.33\% and 9.83\% in patch classification accuracy by increasing the image and input resolution.  Here, we observe an additional gain of 3.30\% and 5.42\% in patch classification accuracy when we initialize the weights of the ResNet-34 model by supervised pre-training on the ImageNet 2012 dataset. On the PCam-Scarce dataset, our final model surpasses the model trained on 100\% of the dataset in the original image size (96$\times$96). For the CRC-Scarce dataset, our final model is only 2.22\% away from the model trained on 100\% of the dataset in the original image size (224$\times$224). 
Additionally, there is a strong positive correlation of 0.810 (PCam-Scarce) and 0.875 (CRC-Scarce) between the image and input resolution and the patch classification accuracy on models initialized with ImageNet pre-trained weights (Figure~\ref{fig:pcam_lcrc_1p_imgnet}).
This result shows that pre-trained models work well with the image and input resolution factors, yielding an additive gain in performance.

\begin{table}[h]
\centering
\setlength{\tabcolsep}{4pt}
\scalebox{1.00}{
\begin{tabular}{|ccccc|ccccc|} \hline
\multicolumn{5}{|c|}{PCam dataset}&\multicolumn{5}{c|}{CRC dataset} \\ \hline
Subset & Init. & Aug. & Res. & Accuracy (\%) &   Subset & Init. & Aug. & Res. & Accuracy (\%)  \\ \hline
 Scarce &  Random & \greencheck & 96 & 79.64$\pm$0.74 &  Scarce &  Random & \greencheck & 224 & 80.54$\pm$0.78 \\
 Scarce &  Random & \greencheck & 448 & 85.97$\pm$0.68 &  Scarce &  Random & \textcolor{red}{\ding{53}} & 800 & 90.37$\pm$1.34 \\
 Scarce &  ImageNet & \greencheck & 448 & \textbf{89.27$\pm$0.68} &  Scarce &  ImageNet & \textcolor{red}{\ding{53}} & 800 & 95.79$\pm$0.31 \\
\hline
 Full &  ImageNet & \greencheck & 96 & 88.88$\pm$0.92 &  Full &  ImageNet & \greencheck & 224 & \textbf{98.01$\pm$0.14} \\
\hline
\end{tabular}
}
    \caption{Accuracy of our models trained on the PCam dataset and CRC dataset. We also show dataset subset (column 1 and 6), model initialization (columns 2 and 7), the use of data augmentation during training (columns 3 and 8) as well as the image and input resolution (columns 4 and 9) in this table.
    }
    \label{table:exp_colon_pcam}
\end{table}

\newpage

\section{ImageNet 2012 statistics}~\label{sec:imagenetstats}

\begin{figure*}[htb]
  \centering
  \includegraphics[width=6.40in,   ]{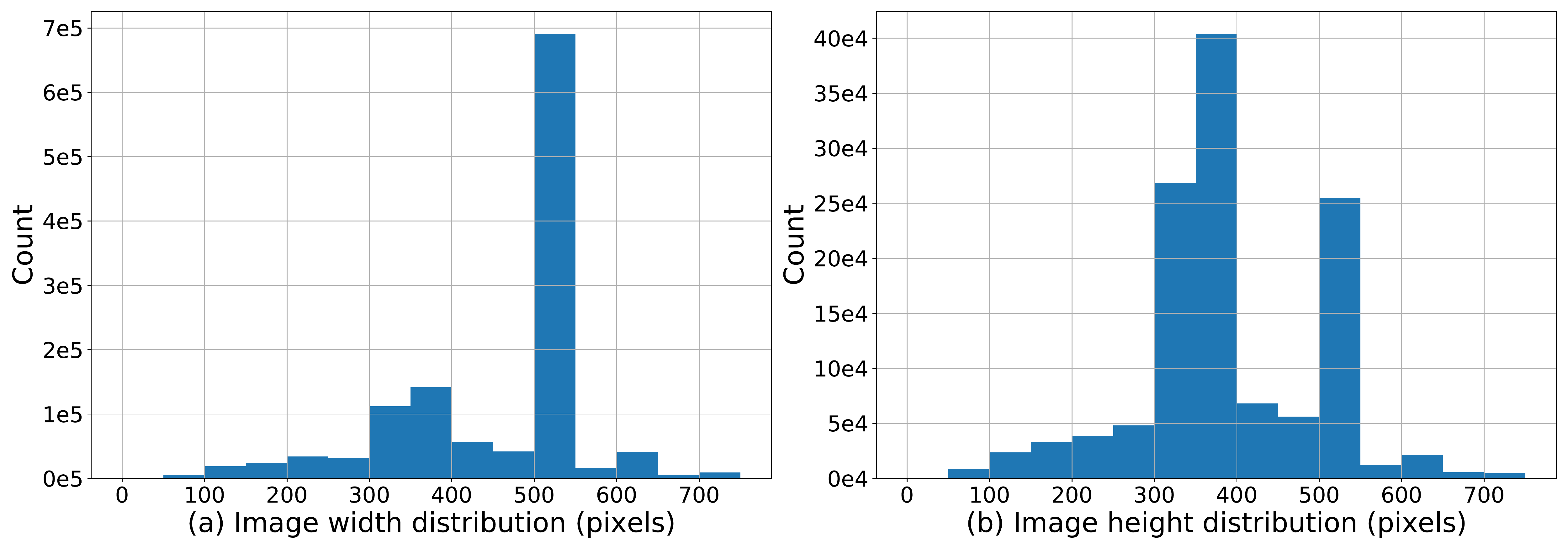}
  \caption{Image resolution distribution of ImageNet 2012 dataset. The average image width and height are 472 and 405, with a standard deviation of 208 and 179.
  }
  \label{fig:imagenet}
\end{figure*}

Figure~\ref{fig:imagenet} shows the image resolution distribution of the ImageNet 2012 dataset. The average image width and height are 472 and 405, with a standard deviation of 208 and 179. $54\%$ of the images have an image width of 500 to 550, and $52\%$ have an image height of 300 to 400.

\section{Data augmentation for Digital Pathology}

Algorithm~\ref{alg:code} shows the data augmentation strategy specifically designed for patch classification in Digital Pathology, following \cite{teh2020learning}. Table~\ref{table:image_res} shows the image and input resolution and the corresponding pad size used in our experiments.
The pad size is approximately 12.5\% to 14.3\% of the corresponding image resolution.

\begin{algorithm}[htb]
\caption{Data Augmentation, PyTorch-like}
\label{alg:code}
\definecolor{codeblue}{rgb}{0.25,0.5,0.5}
\definecolor{codekw}{rgb}{0.85, 0.18, 0.50}
\lstset{
  backgroundcolor=\color{white},
  basicstyle=\fontsize{7.5pt}{7.5pt}\ttfamily\selectfont,
  columns=fullflexible,
  breaklines=true,
  captionpos=b,
  commentstyle=\fontsize{7.5pt}{7.5pt}\color{codeblue},
  keywordstyle=\fontsize{7.5pt}{7.5pt}\color{codekw},
}
\begin{lstlisting}[language=python]
# h,w: height, width; p: padding size
import torchvision.transforms as t

transform = {'train':t.Compose([
                        t.Resize((h, w)),
                        t.Pad(p, padding_mode='reflect'),
                        t.RandomRotation([0, 360]),
                        t.RandomCrop((h,w)),
                        t.RandomHorizontalFlip(0.5),
                        t.ColorJitter(
                            hue= 0.4,
                            saturation=0.4,
                            brightness=0.4,
                            contrast=0.4),
                        t.ToTensor(),
                        ]),
                 'test':t.Compose([
                        t.Resize((h, w)),
                        t.ToTensor(),
                        ])}
\end{lstlisting}
\end{algorithm}

\begin{table}[h]
\centering
\setlength{\tabcolsep}{17pt}
\scalebox{1.0}{
\begin{tabular}{|l|c|c|c|c|c|c|} \hline
Image Resolution (pixels)   & 96 & 224 & 448 & 576 & 672 & 800 \\
\hline
Image Padding (pixels) & 12 & 32 & 64 & 80 & 96 & 114 \\
\hline
\end{tabular}
}
    \caption{Image resolution and the corresponding pad size.}
    \label{table:image_res}
\end{table}

\section{Random-resized-crop}~\label{sec:rrc}

Algorithm~\ref{alg:code_rrc} shows a typical data augmentation strategy used in the Natural Image domain. We compare our data augmentation strategy (Algorithm~\ref{alg:code}) with respect to Algorithm~\ref{alg:code_rrc} on PCam-Scarce dataset with the image resolution of 96$\times$96 and a pad size of 12. The mean accuracy of Algorithm~\ref{alg:code} and Algorithm~\ref{alg:code_rrc} are 79.64$\pm$0.74 and 75.25$\pm$0.91. There is a drop of 4.39\% of mean accuracy by switching from Algorithm~\ref{alg:code} to Algorithm~\ref{alg:code_rrc}, showing the ineffectiveness of Random-resized-crop on patch classification problem in the Digital Pathology domain.

\begin{algorithm}[htb]
\caption{Data Augmentation, PyTorch-like}
\label{alg:code_rrc}
\definecolor{codeblue}{rgb}{0.25,0.5,0.5}
\definecolor{codekw}{rgb}{0.85, 0.18, 0.50}
\lstset{
  backgroundcolor=\color{white},
  basicstyle=\fontsize{7.5pt}{7.5pt}\ttfamily\selectfont,
  columns=fullflexible,
  breaklines=true,
  captionpos=b,
  commentstyle=\fontsize{7.5pt}{7.5pt}\color{codeblue},
  keywordstyle=\fontsize{7.5pt}{7.5pt}\color{codekw},
}
\begin{lstlisting}[language=python]
# h,w: height, width; p: padding size
import torchvision.transforms as t

transform = {'train':t.Compose([
                        t.RandomResizedCrop((h, w)),
                        t.RandomHorizontalFlip(0.5),
                        t.ToTensor(),
                        ]),
                 'test':t.Compose([
                        t.Resize((h+p, w+p)),
                        t.CenterCrop((h, w)),
                        t.ToTensor(),
                        ])}

\end{lstlisting}
\end{algorithm}

\section{Experiments without Data Augmentation}

We also perform experiments without data augmentation (Algorithm~\ref{alg:code_no}) on PCam-Scarce and CRC-Scarce (Figure~\ref{fig:pcam_lcrc_1p}).

\begin{algorithm}[htb]
\caption{Without Data Augmentation, PyTorch-like}
\label{alg:code_no}
\definecolor{codeblue}{rgb}{0.25,0.5,0.5}
\definecolor{codekw}{rgb}{0.85, 0.18, 0.50}
\lstset{
  backgroundcolor=\color{white},
  basicstyle=\fontsize{7.5pt}{7.5pt}\ttfamily\selectfont,
  columns=fullflexible,
  breaklines=true,
  captionpos=b,
  commentstyle=\fontsize{7.5pt}{7.5pt}\color{codeblue},
  keywordstyle=\fontsize{7.5pt}{7.5pt}\color{codekw},
}
\begin{lstlisting}[language=python]
# h,w: height, width;
import torchvision.transforms as t

transform = {'train':t.Compose([
                        t.Resize((h, w)),
                        t.ToTensor(),
                        ]),
                 'test':t.Compose([
                        t.Resize((h, w)),
                        t.ToTensor(),
                        ])}

\end{lstlisting}
\end{algorithm}

\section{Conclusion}

Input and image resolution are important meta-parameters that have been overlooked to-date in DP classification research. In this paper, we show that tuning the image and input resolution can yield impressive gains in performance. Across annotation-scarce and annotation-rich environments, we demonstrate a strong positive correlation between the image and input resolution and the patch classification accuracy. By increasing the image and input resolution, our models can capture finer-grained information without losing coarse-grained information, yielding a gain of 6.33\% (PCam-Scarce) and 9.83\% (CRC-Scarce) in patch classification accuracy. 
We highlighted other important practical considerations such as the interaction of the data augmentation strategy with resolution and choice of dataset.

\newpage
\chapter{ProxyNCA++: Revisiting and Revitalizing ProxyNCA}\label{eccv2020}

\section{Prologue}
This work has been accepted to The European Conference on Computer Vision (2020). Eu Wern Teh is the first author of this publication, and the co-authors are Terrance DeVries and Graham W. Taylor. 

\section{Abstract}
We consider the problem of distance metric learning (DML), where the task is to learn an effective similarity measure between images.
We revisit ProxyNCA and incorporate several enhancements. We find that low temperature scaling is a performance-critical component and explain why it works. Besides, we also discover that Global Max Pooling works better in general when compared to Global Average Pooling. Additionally, our proposed fast moving proxies also addresses small gradient issue of proxies, and this component synergizes well with low temperature scaling and Global Max Pooling.
Our enhanced model, called ProxyNCA++, achieves a 22.9 percentage point average improvement of Recall@1 across four different zero-shot retrieval datasets compared to the original ProxyNCA algorithm.
Furthermore, we achieve state-of-the-art results on the CUB200, Cars196, Sop, and InShop datasets, achieving Recall@1 scores of 72.2, 90.1, 81.4, and 90.9, respectively. 

\section{Introduction}

Distance Metric Learning (DML) is the task of learning effective similarity measures between examples. It is often applied to images, and has found numerous applications such as visual products retrieval \cite{liu2016deepfashion,song2016deep,Bell2015}, person re-identification \cite{Zheng_2019_CVPR,Wang:2018:LDF:3240508.3240552}, face recognition \cite{Schroff_2015_CVPR}, few-shot learning~\cite{Vinyals:2016:MNO:3157382.3157504,koch2015siamese}, and clustering \cite{7471631}. In this paper, we focus on DML's application on zero-shot image retrieval \cite{movshovitz2017no,wu2017sampling,song2016deep,jacob2019metric}, where the task is to retrieve images from previously unseen classes.

Proxy-Neighborhood Component Analysis (ProxyNCA)~\cite{movshovitz2017no} is a proxy-based DML solution that consists of updatable proxies, which are used to represent class distribution. It allows samples to be compared with these proxies instead of one another to reduce computation. 
After the introduction of ProxyNCA, there are very few works that extend ProxyNCA~\cite{xuan2018deep,sanakoyeu2019divide}, making it less competitive when compared with recent DML solutions~\cite{wang2019multi,jacob2019metric,zhai2019}.

\begin{figure}[H]
  \centering
  \includegraphics[width=6.00in, ]{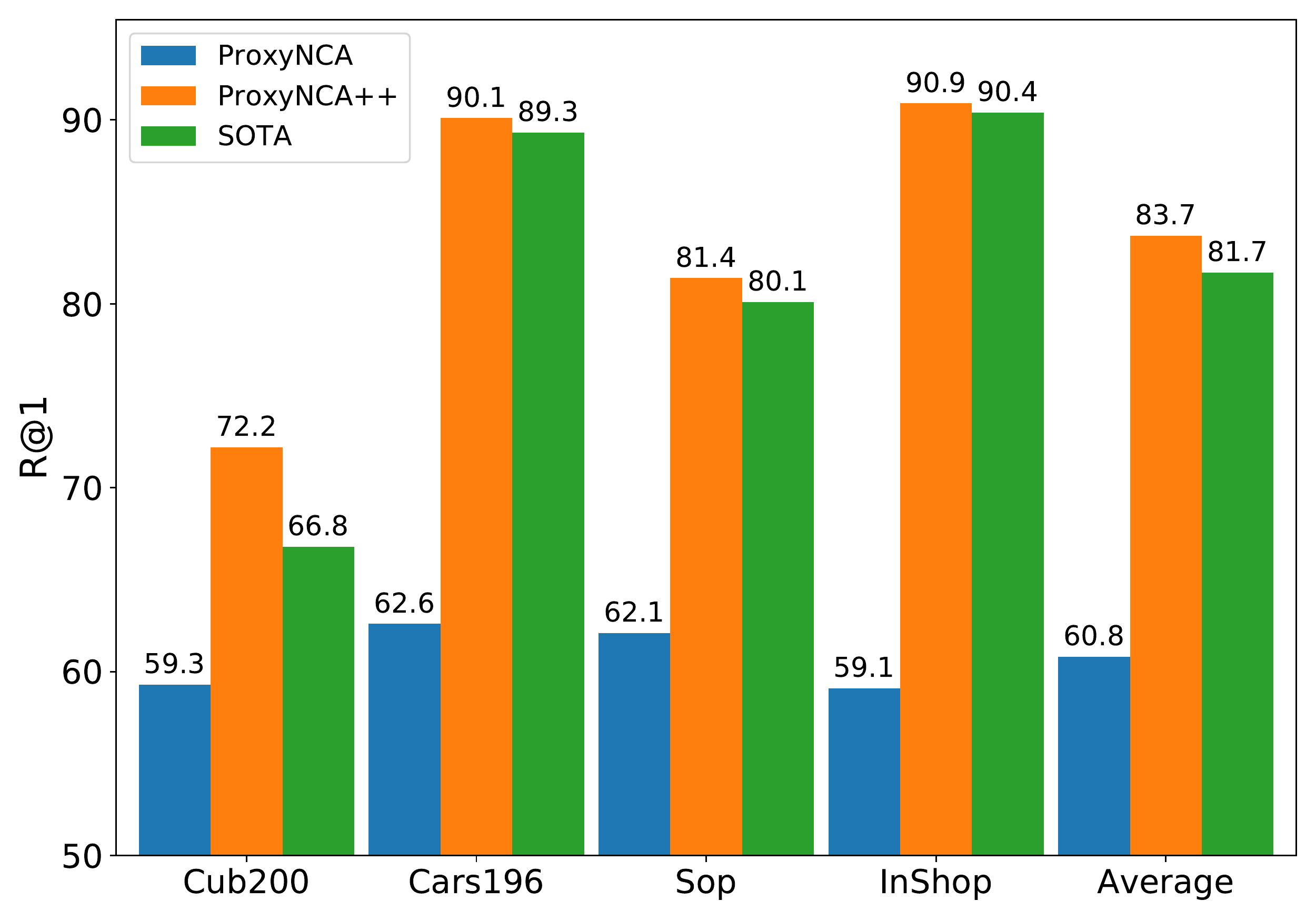}
  \caption{A summary of the average performance on Recall@1 for all datasets. With our proposed enhancements, we improve upon the original ProxyNCA by $22.9$pp, and outperform current state-of-the-art models by $2.0$pp on average.}
  \label{fig:summ}
\end{figure}

Our contributions are the following: First, we point out the difference between NCA and ProxyNCA, and propose to use proxy assignment probability which aligns ProxyNCA with NCA~\cite{goldberger2005neighbourhood}. Second, we explain why low temperature scaling works and show that it is a performance-critical component of ProxyNCA. Third, we explore different global pooling strategies and find out that Global Max Pooling (GMP) outperforms the commonly used Global Average Pooling (GAP), both for ProxyNCA and other methods. Fourth, we suggest using faster moving proxies that compliment well with both GMP and low temperature scaling, which also address the small gradient issue due to $L^2$-Normalization of proxies. 
Our enhanced ProxyNCA, which we called ProxyNCA++, has a $22.9$ percentage points of improvement over ProxyNCA on average for Recall@1 across four different zero-shot retrieval benchmarks (performance gains are highlighted in Figure~\ref{fig:summ}). In addition, we also achieve state-of-the-art performance on all four benchmark dataset across all categories.

\section{Related Work}
The core idea of Distance Metric Learning (DML) is to learn an embedding space where similar examples are attracted, and dissimilar examples are repelled. 
To restrict the scope, we limit our review to methods that consider image data. There is a large body of work in DML, and it can be traced back to the 90s, where Bromley et al.~\cite{Bromley:1993:SVU:2987189.2987282} designed a Siamese neural network to verify signatures. Later, DML was used in facial recognition, and dimensionality reduction in the form of a contrastive loss \cite{chopra2005learning,hadsell2006dimensionality},
where pairs of similar and dissimilar images are selected, and the distance between similar pairs of images is minimized while the distance between dissimilar images is maximized.

Like contrastive loss, which deals with the actual distance between two images, triplet loss optimizes the relative distance between positive pair (an anchor image and an image similar to anchor image) and negative pair (an anchor image and an image dissimilar to anchor image)~\cite{Chechik:2010:LSO:1756006.1756042}. In addition to contrastive and triplet loss, there is a long line of work which proposes new loss functions, such as angular loss~\cite{Wang_2017_ICCV}, histogram loss~\cite{NIPS2016_6464}, margin-based loss~\cite{wu2017sampling}, and hierarchical triplet loss~\cite{Ge_2018_ECCV}. Wang et al.~\cite{wang2019multi}~categorize this group as paired-based DML.

One weakness of paired-based methods is the sampling process. First, the number of possible pairs grows polynomially with the number of data points, which increases the difficulty of finding an optimal solution. Second, if a pair or triplet of images is sampled randomly, the average distance between two samples is approximately $\sqrt{2}$-away \cite{wu2017sampling}. In other words, a randomly sampled image is highly redundant and provides less information than a carefully chosen one.

In order to overcome the weakness of paired-based methods, several works have been proposed in the last few years. Schroff et al.~\cite{Schroff_2015_CVPR} explore a curriculum learning strategy where examples are selected based on the distances of samples to the anchored images. They use a semi-hard negative mining strategy to select negative samples where the distances between negative pairs are at least greater than the positive pairs. However, such a method usually generates very few semi-hard negative samples, and thus requires very large batches (on the order of thousands of samples) in order to be effective.
Song et al.~\cite{song2016deep} propose to utilize all pair-wise samples in a mini-batch to form triplets, where each positive pair compares its distance with all negative pairs. Wu et al.~\cite{wu2017sampling} proposed a distance-based sampling strategy, where examples are sampled based on inverse $n$-dimensional unit sphere distances from anchored samples. Wang et al.~\cite{wang2019multi} propose a mining and weighting scheme, where informative pairs are sampled by measuring positive relative similarity, and then further weighted using self-similarity and negative relative similarity.

Apart from methods dedicated to addressing the weakness of pair-based DML methods, there is another line of work that tackles DML via class distribution estimation. The motivation for this camp of thought is to compare samples to proxies, and in doing so, reduce computation.  One method that falls under this line of work is the Magnet Loss~\cite{rippel2015metric} in which samples are associated with a cluster centroid, and at each training batch, samples are attracted to cluster centroids of similar classes and repelled by cluster centroids of different classes. Another method in this camp is ProxyNCA~\cite{movshovitz2017no}, where proxies are stored in memory as learnable parameters. During training, each sample is pushed towards its proxy while repelling against all other proxies of different classes. ProxyNCA is discussed in greater detail in Section \ref{sec:proxynca}.

Similar to ProxyNCA, Zhai et al.~\cite{zhai2019} design a proxy-based solution that emphasizes on the Cosine distance rather than the Euclidean squared distance. They also use layer norm in their model to improve robustness against poor weight initialization of new parameters and introduces class balanced sampling during training, which improves their retrieval performance. In our work, we also use these enhancements in our architecture.

Recently, a few works in DML have explored ensemble techniques. 
Opitz et al.~\cite{Opitz_2017_ICCV} train an ensemble DML by reweighting examples using online gradient boosting. 
The downside of this technique is that it is a sequential process. 
Xuan et al.~\cite{xuan2018deep} address this issue by proposing an ensemble technique where ensemble models are trained separately on randomly combined classes.
Sanakoyeu et al.~\cite{sanakoyeu2019divide} propose a unique divide-and-conquer strategy where the data is divided periodically via clustering based on current combined embedding during training. Each cluster is assigned to a consecutive chunk of the embedding, called learners, and they are randomly updated during training. Apart from ensemble techniques, there is recent work that attempts to improve DML in general. Jacob et al.~\cite{jacob2019metric} discover that DML approaches that rely on Global Average Pooling (GAP) potentially suffer from the scattering problem, where features learned with GAP are sensitive to outlier. To tackle this problem, they propose HORDE, which is a high order regularizer for deep embeddings that computes higher-order moments of features.

\section{Methods}

\begin{figure}
  \centering
  \includegraphics[width=5.4in]{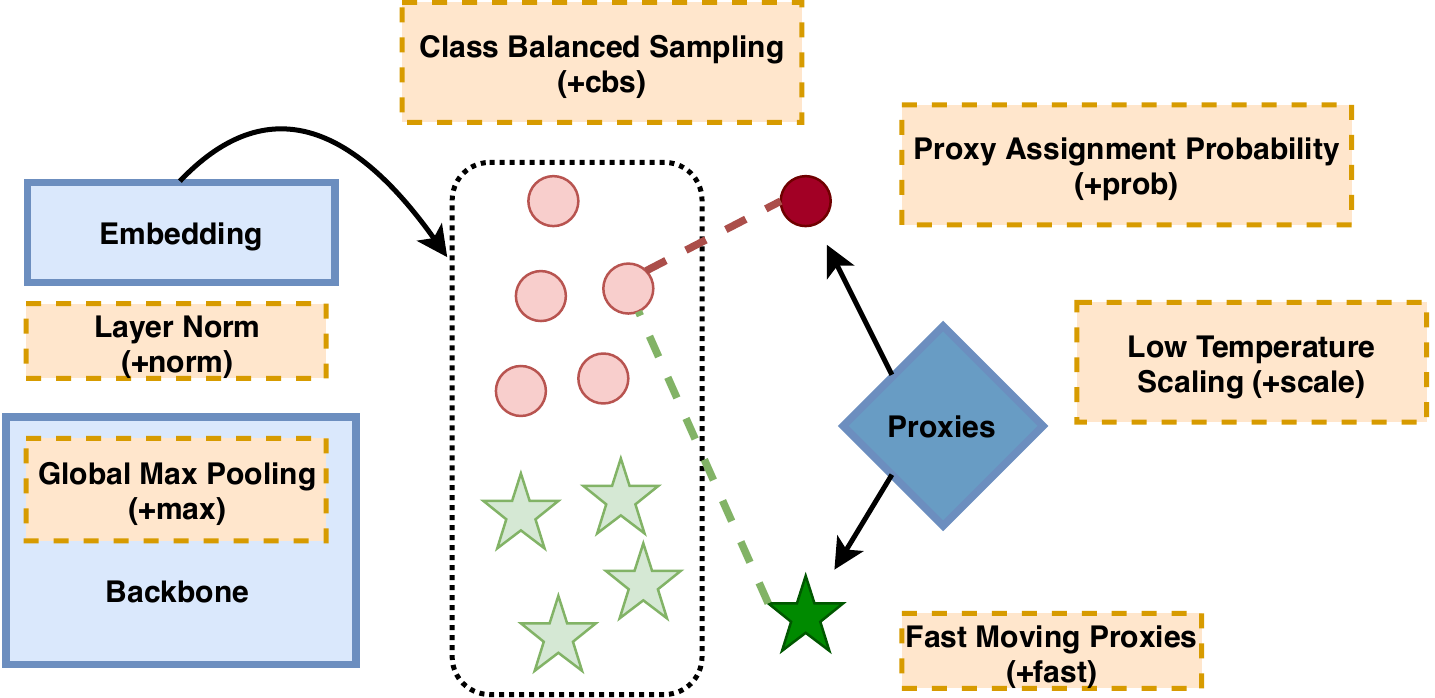}
  \caption{
  We show an overview of our architecture, ProxyNCA++, which consists of the original building blocks of ProxyNCA and six enhancements, which are shown in the dashed boxes. ProxyNCA consists of a pre-trained backbone model, a randomly initialized embedding layer, and randomly initialized proxies. The six enhancements in ProxyNCA++ are proxy assignment probability (+prob), low temperature scaling (+scale), class balanced sampling (+cbs), layer norm (+norm), global max pooling (+max) and fast-moving proxies (+fast).
  }

  \label{fig:pnca_pp}
\end{figure}

In this section, we revisit NCA and ProxyNCA and discuss six enhancements that improve the retrieval performance of ProxyNCA. The enhanced version, which we call ProxyNCA++, is shown in Figure~\ref{fig:pnca_pp}.

\subsection{Neighborhood Component Analysis (NCA)}
Neighborhood Component Analysis (NCA) is a DML algorithm that learns a Mahalanobis distance for k-nearest neighbors (KNN).  Given two points, $x_i$ and $x_j$, Goldberg et al.~\cite{goldberger2005neighbourhood} define $p_{ij}$ as the assignment probability of $x_i$ to $x_j$:
 \begin{equation}\label{eq:4}
 p_{ij} = \frac{-d(x_i, x_j)}{\sum_{k \not\in i} -d(x_i, x_k)}
\end{equation}

\noindent where $d(x_i, x_k)$ is Euclidean squared distance computed on some learned
embedding. In the original work, it was parameterized as a linear mapping, but
nowadays, the method is often used with nonlinear mappings such as feedforward or
convolutional neural networks. Informally, $p_{ij}$ is the probability that
points $i$ and $j$ are said to be ``neighbors''.

The goal of NCA is to maximize the probability that points assigned to the same
class are neighbors, which, by normalization, minimizes the probability that
points in different classes are neighbors:

\begin{equation}\label{eq:nca}
 L_{\text{NCA}} = -\log\left(\frac{\sum_{j \in C_i}^{}\text{exp}(-d(x_i, x_j))}{\sum_{k \not\in C_i}^{}\text{exp}(-d(x_i, x_k))}\right) .\
\end{equation}

Unfortunately, the computation of NCA loss grows polynomially with the number of samples in the dataset. To speed up computation, Goldberg et al.~use random sampling and optimize the NCA loss with respect to the small batches of samples.

\subsection{ProxyNCA}
\label{sec:proxynca}
ProxyNCA is a DML method which performs metric learning in the space of class distributions. It is motivated by NCA, and it attempts to address the computation weakness of NCA by using proxies. 
In ProxyNCA, \emph{proxies} are stored as learnable parameters to faithfully represent classes by prototypes in an embedding space. During training, instead of comparing samples with one another in a given batch, which is quadratic in computation with respect to the batch size, ProxyNCA compares samples against proxies, where the objective aims to attract samples to their proxies and repel them from all other proxies. 

Let $C_i$ denote a set of points that belong to the same class, $f(a)$ be a proxy function that returns a corresponding class proxy, and ${||a||}_2$ be the $L^2$-Norm of vector $a$. For each sample $x_i$, we minimize the distance  $d(x_i, f(x_i))$ between the sample, $x_i$ and its own proxy, $f(x_i)$ and maximize the distance  $d(x_i, f(z))$ of that sample with respect to all other proxies $Z$, where $ f(z) \in Z$ and $z \not \in C_i$.

\begin{equation}\label{eq:1_eccv2020}
 L_{\text{ProxyNCA}} = -\log\left(\frac{\text{exp}\left(-d(\frac{x_i}{||x_i||_2}, \frac{f(x_i)}{||f(x_i)||_2})\right)}{\sum_{f(z)\in Z}^{}\text{exp}\left(-d(\frac{x_i}{||x_i||_2}, \frac{f(z)}{||f(z)||_2})\right)}\right) .\
\end{equation}
\subsection{Aligning with NCA by optimizing proxy assignment probability}

Using the same motivation as NCA (Equation~\ref{eq:4}), we propose to optimize the proxy assignment probability, $P_{i}$. Let  $A$ denote the set of all proxies. For each $x_i$, we aim to maximize $P_{i}$.

\begin{align}\label{eq:pp}
 &P_{i} = \frac{\text{exp}\left(-d(\frac{x_i}{||x_i||_2}, \frac{f(x_i)}{||f(x_i)||_2})\right)}{\sum_{f(a)\in A}^{}\text{exp}\left(-d(\frac{x_i}{||x_i||_2}, \frac{f(a)}{||f(a)||_2})\right)} \\
 &  L_{\text{ProxyNCA++}} = -\log(P_{i})
\end{align}

Since $P_i$ is a probability score that must sum to one, maximizing $P_i$ for a proxy also means there is less chance for $x_i$ to be assigned to other proxies. In addition, maximizing $P_i$ also preserves the original ProxyNCA properties where $x_i$ is attracted toward its own proxy $f(x_i)$ while repelling proxies of other classes, $Z$. It is important to note that in ProxyNCA, we maximize the distant ratio between $-d(x_i, y_j)$ and $\sum_{f(z)\in Z}^{}-d(x_i, f(z))$, while in ProxyNCA++, we maximize the proxy assignment probability, $P_{i}$, a subtle but important distinction. Table~\ref{table:prob_det} shows the effect of proxy assignment probability to ProxyNCA and its enhancements.

\subsection{About Temperature Scaling}

Temperature scaling is introduced in \cite{hinton2015distilling}, where Hinton et al. use a high temperature ($T > 1$) to create a softer probability distribution over classes for knowledge distillation purposes. Given a logit $y_i$ and a temperature variable $T$, a temperature scaling is defined as $q_{i} = \frac{\exp(y_i/T)}{\sum_j \exp(y_j/T)}$. By incorporating temperature scaling to the loss function of ProxyNCA++ in Equation~\ref{eq:pp}, the new loss function has the following form:

\begin{align}\label{eq:ppp}
 &L_{\text{ProxyNCA++}} = -\log\left(\frac{\text{exp}\left(-d(\frac{x_i}{||x_i||_2}, \frac{f(x_i)}{||f(x_i)||_2})*\frac{1}{T}\right)}{\sum_{f(a)\in A}^{}\text{exp}\left(-d(\frac{x_i}{||x_i||_2}, \frac{f(a)}{||f(a)||_2})*\frac{1}{T}\right)} \right)
\end{align}

\begin{figure}
  \centering
  \includegraphics[width=6.4in]{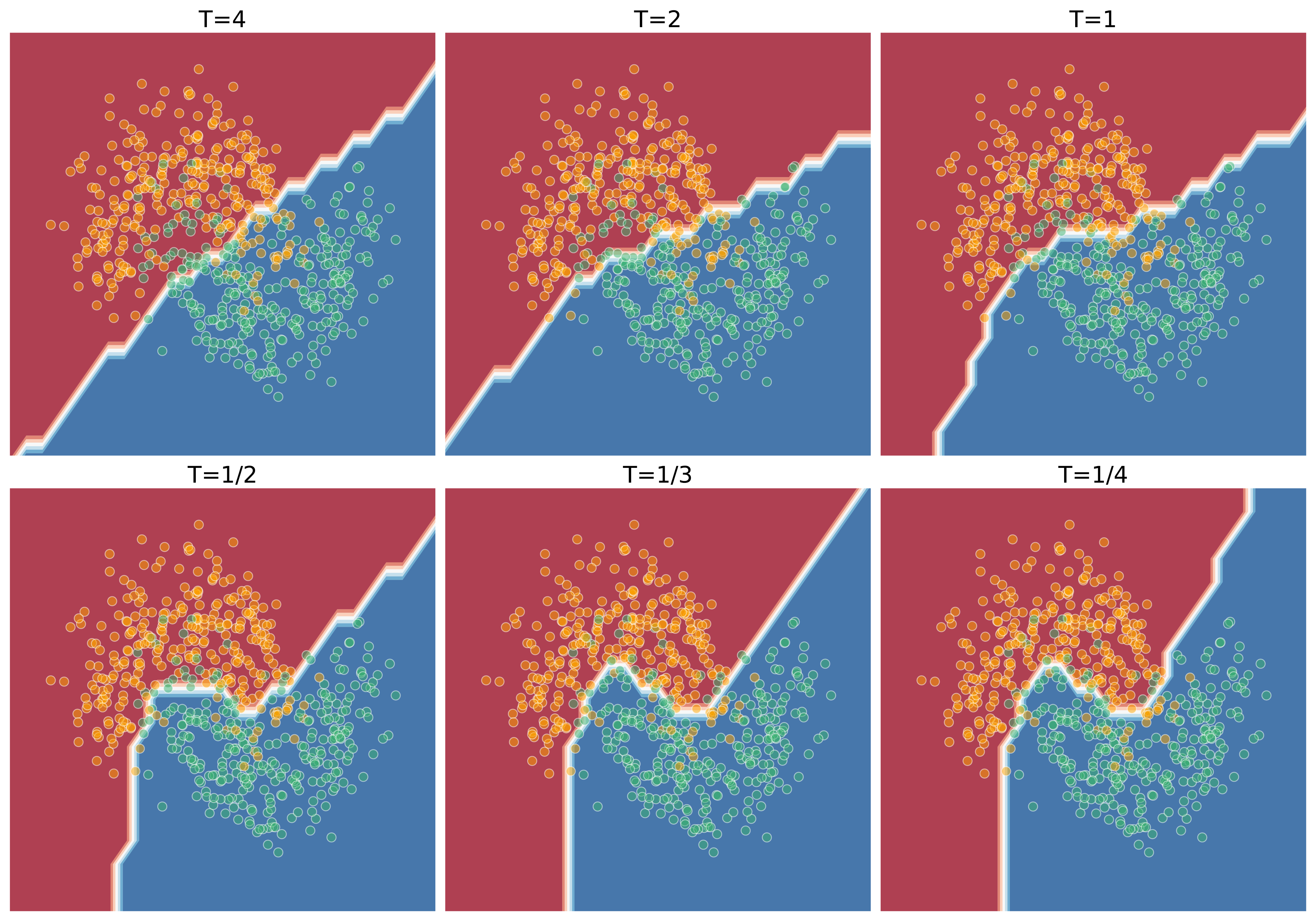}
  \caption{The effect of temperature scaling on the decision boundary of a Softmax Classifier trained on the two moons synthetic dataset.}
  \label{fig:temp_scale}
\end{figure}

When $T = 1$, we have a regular Softmax function. As $T$ gets larger, the output of the softmax function will approach a uniform distribution. On the other hand, as $T$ gets smaller, it leads to a peakier probability distribution. Low temperature scaling ($T < 1$) is used in \cite{wu2018improving} and \cite{zhai2019}. In this work, we attempt to explain why low-temperature scaling works by visualizing its effect on synthetic data. In Figure~\ref{fig:temp_scale}, as $T$ gets smaller, the decision boundary is getting more refined and can classify the samples better. In other words, as $T$ becomes smaller, the model can overfit to the problem better and hence generating better decision boundaries.

\begin{figure}
  \centering
  \includegraphics[width=6.4in]{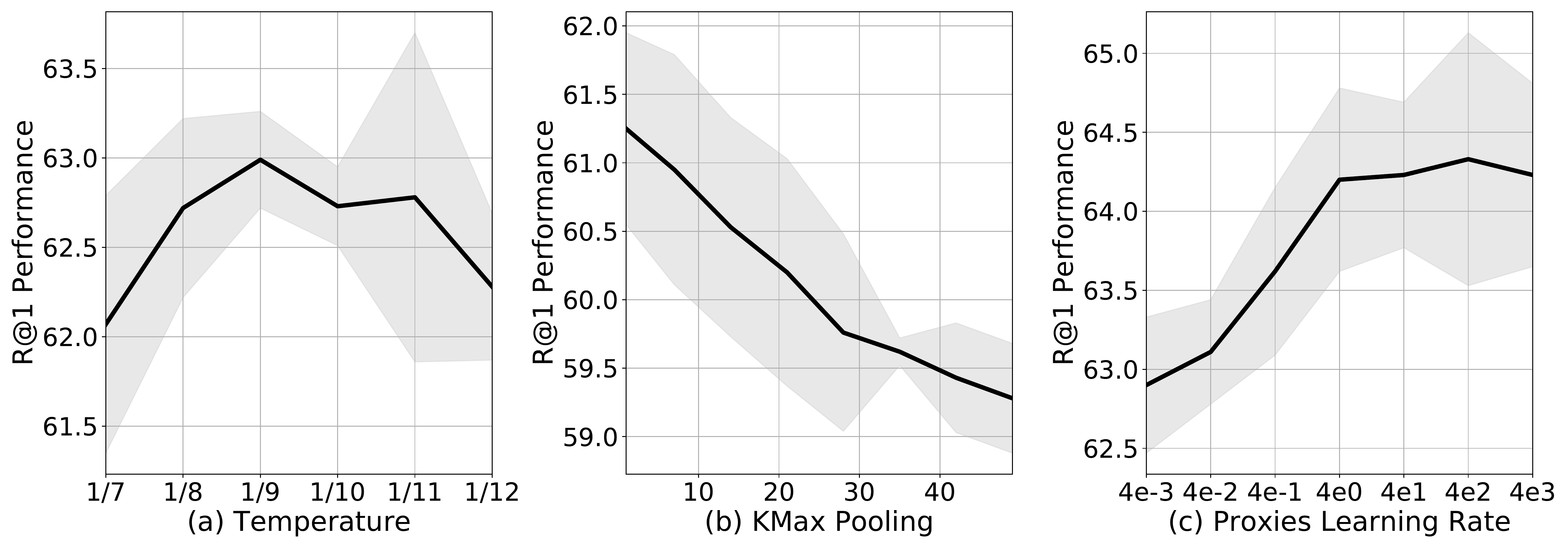}
  \caption{We show three plots of R@1 with different (a) temperature scales , (b) $k$ values for $K$-Max Pooling and (c) proxy learning rates on on CUB200~\cite{wah2011caltech}. The shaded areas represent one standard deviation of uncertainty. }
  \label{fig:ts_kmax}
\end{figure}

In Figure~\ref{fig:ts_kmax} (a), we show a plot of R@1 score with respect to temperature scale on the CUB200 dataset. The highest test average R@1 happens at $T=\frac{1}{9}$. Lowering $T$ beyond this point will allow the model to overfit more to the training set and to make it less generalizable. Hence, we see a drop in test performance.
Table~\ref{table:tempscale_det} shows the effect of low temperature scaling to ProxyNCA and its enhancements.

\subsection{About Global Pooling}
In DML, the de facto global pooling operation used by the community is Global Average Pooling (GAP).  In this paper, we investigate the effect of global pooling of spatial features on zero-shot image retrieval. We propose the use of Global $K$-Max Pooling~\cite{durand2016weldon} to interpolate between GAP and Global Max Pooling (GMP). Given a convolution feature map of $M \times M$ dimension with $E$ channels, $g \in \mathbb{R}^{M \times M \times E}$ and a binary variable, $h_i \in \{0,1\}$, Global K-Max Pooling is defined as:

\begin{equation}\label{eq:kmax}
 \text{Global }k\text{-Max}(g_{\epsilon}) = \max\limits_{h} \frac{1}{k} \sum_{i=1}^{M^2} h_i \cdot g_{\epsilon} \text{\ , s.t.} \sum_{i=1}^{M^2} h_i = k, \forall \epsilon \in E
\end{equation}

When $k=1$, we have GMP, and when $k=M^2$, we have GAP. Figure~\ref{fig:ts_kmax} (b) is a plot of Recall@1 with different $k$ value of Global K-Max Pooling on the CUB200 dataset. There is a negative correlation of 0.98 between $k$ and Recall@1 performance, which shows that a lower $k$ value results in better retrieval performance.

\subsection{About Fast moving proxies}

In ProxyNCA, the proxies, the embedding layer, and the backbone model all share the same learning rate. We hypothesize that the proxies should be moving faster than the embedding space in order to represent the class distribution better. However, in our experiments, we discovered that the gradient of proxies is smaller than the gradient of the embedding layer and backbone model by three orders of magnitude, and this is caused by the  $L^2$-Normalization of proxies. 
To mitigate this problem, we use a higher learning rate for the proxies. 

From our ablation studies in Table~\ref{table:tempscale_det}, we observe that fast moving proxies synergize better with low temperature scaling and Global Max Pooling. We can see a 1.4pp boost in R@1 if we combine fast proxies and low temperature scaling. There is also a 2.1pp boost in the retrieval performance if we combine fast proxies, low temperature scaling, and Global Max Pooling. Figure~\ref{fig:ts_kmax} (c) is a plot of Recall@1 with different proxy learning rates on CUB200.

\subsection{Layer Norm (Norm) and Class Balanced Sampling (CBS)}
The use of layer normalization \cite{vaswani2017attention} without affine parameters is explored by Zhai et al.~\cite{zhai2019}. Based on our experiments, we also find that this enhancement helps to boost performance. Besides, we also use a class balanced sampling strategy in our experiments, where we have more than one instance per class in each training batch. To be specific, for every batch of size $N_b$, we only sample $N_c$ classes from which we then randomly select $\lfloor N_b/N_c\rfloor$ examples. This sampling strategy commonly appears in pair-based DML approaches~\cite{wang2019multi,wu2017sampling,song2016deep} as a baseline and
Zhai et al. is the first paper that uses it in a proxy-based DML method.

\section{Experiments}

We train and evaluate our model on four zero-shot image retrieval datasets: the Caltech-UCSD Birds dataset~\cite{wah2011caltech} (CUB200), the Stanford Cars dataset~\cite{KrauseStarkDengFei-Fei_3DRR2013} (Cars196), the Stanford Online Products dataset~\cite{song2016deep} (Sop), and the In Shop Clothing Retrieval dataset~\cite{liu2016deepfashion} (InShop). The composition in terms of number of images and classes of each dataset is summarized in Table~\ref{table:dataset}.

\subsection{Experimental Setup}

For each dataset, we use the first half of the original training set as our training set and the second half of the original training set as our validation set. In all of our experiments, we use a two-stage training process. We first train our models on the training set and then use the validation set to perform hyper-parameter tuning (e.g., selecting the best epoch for early stopping, learning rate, etc.). Next, we train our models with the fine-tuned hyper-parameters on the combined training and validation sets (i.e., the complete original training set).

\begin{table}
\centering
\setlength{\tabcolsep}{8pt}
\begin{tabular}{|lrrccc|ccc|}
\hline
 &  &  &  & batch size & batch size & & & \\
 & images & classes & avg & (ours) & (MS~\cite{wang2019multi}) & Base lr & Proxy lr & cbs\\ \hline
\small{CUB200}& 11,788 & 200 & 58 & 32 & 80      & 4e-3 & 4e2 &4 \\
\small{Cars196}& 16,185 & 196 & 82 & 32 & -      & 4e-3 & 4e2 &4\\
\small{Sop}& 120,053 & 22,634 & 5 & 192 & 1000   & 2.4e-2 & 2.4e2 &3\\
\small{InShop}& 52,712 & 11,967 & 4 & 192 & -    & 2.4e-2 & 2.4e3 &3\\
\hline
\end{tabular}
\caption{We show the composition of all four zero-shot image retrieval datasets considered in this work. In addition, we also report the learning rates, the batch size, and cbs (class balanced sampling) instances for each dataset during training. The number of classes for the Sop and InShop datasets is large when compared to CUB200 and Cars196 dataset. However, the number of instances per class is very low for the Sop and InShop datasets. In general, ProxyNCA does not require a large batch size when compared to pairs-based DML methods. To illustrate this, we also show the batch sizes used in \cite{wang2019multi}, which is current state-of-the-art among pairs-based methods. Their technique requires a batch size, which is several times larger compared to ProxyNCA++.}
\label{table:dataset}
\end{table}

We use the same learning rate for both stages of training.
We also set the number of proxies to be the same as the number of classes in the training set. For our experiments with fast proxies, we use a different learning rate for proxies (see Table~\ref{table:dataset} for details). We also use a temperature value of  $\frac{1}{9}$ across all datasets.

In the first stage of training, we use the ``reduce on loss plateau decay'' annealing~\cite{goodfellow2016deep} to control the learning rate of our model based on the recall performance (R@1) on the validation set. We set the patience value to four epochs in our experiments. We record the epochs where the learning rate is reduced and also save the best epochs for early stopping on the second stage of training.

In all of our experiments, we leverage the commonly used ImageNet~\cite{ILSVRC15} pre-trained Resnet50~\cite{he2016deep} model as our backbone (see Table~\ref{table:arch} for commonly used backbone architectures). Features are extracted after the final convolutional block of the model and are reduced to a spatial dimension of $1\times 1$ using a global pooling operation. This procedure results in a 2048 dimensional vector, which is fed into a final embedding layer. In addition, we also experiment with various embedding sizes. We observe a gain in performance as we increase the size of the embedding. It is important to note that not all DML techniques yield better performance as embedding size increases. For some techniques such as \cite{wang2019multi,song2016deep}, a larger embedding size hurts performance.

\begin{table}
\centering
\setlength{\tabcolsep}{42pt}
\begin{tabular}{|lcc|}
\hline
Architecture & Abbreviation & Top-1 Error~(\%) \\ \hline
 \small{Resnet18~\cite{he2016deep}}& R18 &30.24 \\
\small{GoogleNet~\cite{I1}} &I1 &  30.22  \\
\small{Resnet50~\cite{he2016deep}}&R50&  23.85 \\
\small{InceptionV3~\cite{I3}}& I3 & 22.55 \\
\hline
\end{tabular}
\caption{Commonly used backbone architectures for zero-shot image retrieval, with associated ImageNet Top-1 Error \% for each architecture.}
\label{table:arch}
\end{table}

During training, we scale the original images to a random aspect ratio (0.75 to 1.33) before applying a crop of random size (0.08 to 1.0 of the scaled image). After cropping, we resize the images to 256$\times$256. We also perform random horizontal flipping for additional augmentation. During testing, we resize the images to 288$\times$288 and perform a center crop of size 256$\times$256.

\subsection{Evaluation}

 We evaluate retrieval performance based on two evaluation metrics: (a) Recall@K (R@K) and (b) Normalized Mutual Information, $\text{NMI}(\Omega, \mathbb{C}) = \frac{2*I(\Omega, \mathbb{C})}{H(\Omega) + H(\mathbb{C})}$, where $\Omega$ represents ground truth label, $\mathbb{C}$ represents the set of clusters computed by K-means, $I$ stands for mutual information and $H$ stands for entropy. The purpose of NMI is to measure the purity of the cluster on unseen data.

Using the same evaluation protocols detailed in \cite{movshovitz2017no,wang2019multi,jacob2019metric,liu2016deepfashion}, we evaluate our model using unseen classes on four datasets.
The InShop dataset~\cite{liu2016deepfashion} is slightly different than all three other datasets. There are three groupings of data: training set, query set, and gallery set. The query and gallery set have the same classes, and these classes do not overlap with the training set. Evaluation is done based on retrieval performance on the gallery set.

Tables~\ref{table:cub},~\ref{table:cars}, ~\ref{table:sop}, and ~\ref{table:inshop} show the results of our experiments~\footnote[1]{For additional experiments on different crop sizes, please refer to the corresponding supplementary materials in the appendix}.
For each dataset, we report the results of our method, averaged over five runs. We also report the standard deviation of our results to account for uncertainty.
Additionally, we also show the results of ProxyNCA++ trained with smaller embedding sizes (512, 1024).
Our ProxyNCA++ model outperforms ProxyNCA and all other state-of-the-art methods in all categories across all four datasets. Note, our model trained with a 512-dimensional embedding also outperform all other methods in the same embedding space except for The InShop dataset~\cite{liu2016deepfashion}, where we tie in the R@1 category.

\begin{table}[H]
\centering
\setlength{\tabcolsep}{6pt}
\begin{tabular}{|*8c|}
\hline
R@k & 1 & 2 & 4 & 8 & NMI & Arch & Emb\\ \hline
\small{ProxyNCA\cite{movshovitz2017no}} & 49.2 & 61.9 & 67.9 & 72.4 & 59.5& \small{I1} & \small{128}\\
\small{Margin\cite{wu2017sampling}} & 63.6 & 74.4 & 83.1 & 90.0 & 69.0 & \small{R50}& \small{128}\\
\small{MS~\cite{wang2019multi}}& 65.7 & 77.0 & 86.3 & 91.2 & - & \small{I3}& \small{512}\\
\small{HORDE~\cite{jacob2019metric}}& 66.8 & 77.4 & 85.1 & 91.0 & - & \small{I3}& \small{512}\\
\small{NormSoftMax~\cite{zhai2019}}& 61.3 & 73.9 & 83.5 & 90.0 & - & \small{R50}& \small{512}\\
\small{NormSoftMax~\cite{zhai2019}}& 65.3 & 76.7 & 85.4 & 91.8 & - & \small{R50}& \small{2048}\\
\hline
\small{ProxyNCA} & 59.3$\pm$0.4 & 71.2$\pm$0.3 & 80.7$\pm$0.2 & 88.1$\pm$0.3 & 63.3$\pm$0.5 & \small{R50}& \small{2048}\\
\small{ProxyNCA++} & 69.0$\pm$0.8 & 79.8$\pm$0.7 & 87.3$\pm$0.7 & 92.7$\pm$0.4 & 73.9$\pm$0.5 & \small{R50}& \small{512}\\
\small{ProxyNCA++} & 70.2$\pm$1.6 & 80.7$\pm$1.4 & 88.0$\pm$0.9 & 93.0$\pm$0.4 & 74.2$\pm$1.0 & \small{R50}& \small{1024}\\
\small{ProxyNCA++} & 69.1$\pm$0.5 & 79.6$\pm$0.4 & 87.3$\pm$0.3 & 92.7$\pm$0.2 & 73.3$\pm$0.7 & \small{R50}& \small{2048}\\
\small{(-max, -fast)}& & & & & & &\\
\small{ProxyNCA++} & \textbf{72.2$\pm$0.8} & \textbf{82.0$\pm$0.6} & \textbf{89.2$\pm$0.6} & \textbf{93.5$\pm$0.4} & \textbf{75.8$\pm$0.8} & \small{R50}& \small{2048}\\
\hline
\end{tabular}
\caption{Recall@k for k = 1,2,4,8 and NMI on CUB200-2011~\cite{wah2011caltech}.}
\label{table:cub}
\end{table}

\begin{table}[H]
\centering
\setlength{\tabcolsep}{6pt}
\begin{tabular}{|*8c|}
\hline
R@k & 1 & 2 & 4 & 8 & NMI & Arch & Emb\\ \hline
\small{ProxyNCA~\cite{movshovitz2017no}} & 73.2 & 82.4 & 86.4 & 88.7 & 64.9 &\small{I1}&\small{128}\\
\small{Margin~\cite{wu2017sampling}} & 79.6 & 86.5 & 91.9 & 95.1 & 69.1 & \small{R50}&\small{128}\\
\small{MS~\cite{wang2019multi}}& 84.1 & 90.4 & 94.0 & 96.1 & - &\small{I3}&\small{512}\\
\small{HORDE~\cite{jacob2019metric}}& 86.2 & 91.9 & 95.1 & 97.2 & - & \small{I3}&\small{512}\\
\small{NormSoftMax~\cite{zhai2019}}& 84.2 & 90.4 & 94.4 & 96.9 & - & \small{R50}&\small{512}\\
\small{NormSoftMax~\cite{zhai2019}}& 89.3 & 94.1 & 96.4 & 98.0 & - & \small{R50}&\small{2048}\\
\hline
\small{ProxyNCA}& 62.6$\pm$9.1 & 73.6$\pm$8.6 & 82.2$\pm$6.9 & 88.9$\pm$4.8 & 53.8$\pm$7.0 & \small{R50}&\small{2048}\\
\small{ProxyNCA++}& 86.5$\pm$0.4 & 92.5$\pm$0.3 & 95.7$\pm$0.2 & 97.7$\pm$0.1 & 73.8$\pm$1.0 & \small{R50}&\small{512}\\
\small{ProxyNCA++}& 87.6$\pm$0.3 & 93.1$\pm$0.1 & 96.1$\pm$0.2 & 97.9$\pm$0.1 & 75.7$\pm$0.3 & \small{R50}&\small{1024}\\
\small{ProxyNCA++}& 87.9$\pm$0.2 & 93.2$\pm$0.2 & 96.1$\pm$0.2 & 97.9$\pm$0.1 & 76.0$\pm$0.5 & \small{R50}&\small{2048}\\
\small{(-max, -fast)}& & & & & & & \\
\small{ProxyNCA++}& \textbf{90.1$\pm$0.2} & \textbf{94.5$\pm$0.2} & \textbf{97.0$\pm$0.2} & \textbf{98.4$\pm$0.1} & \textbf{76.6$\pm$0.7} & \small{R50}&\small{2048}\\
\hline
\end{tabular}
\caption{Recall@k for k = 1,2,4,8 and NMI on CARS196~\cite{KrauseStarkDengFei-Fei_3DRR2013}.}
\label{table:cars}
\end{table}

\begin{table}[H]
\centering
\setlength{\tabcolsep}{8pt}
\begin{tabular}{|*7c|}
\hline
R@k & 1 & 10 & 100 & 1000 & Arch & Emb \\ \hline
\small{ProxyNCA~\cite{movshovitz2017no}} & 73.7 & - & - & - & \small{I1}& \small{128}\\
\small{Margin~\cite{wu2017sampling}} & 72.7 & 86.2 & 93.8 & 98.0 &\small{R50}& \small{128}\\
\small{MS~\cite{wang2019multi}}& 78.2 & 90.5 & 96.0 & 98.7 & \small{I3}& \small{512}\\
\small{HORDE~\cite{jacob2019metric}}& 80.1 & 91.3 & 96.2 & 98.7 & \small{I3}& \small{512}\\
\small{NormSoftMax~\cite{zhai2019}}& 78.2 & 90.6 & 96.2 & - &  \small{R50}& \small{512}\\
\small{NormSoftMax~\cite{zhai2019}}& 79.5 & 91.5 & 96.7 & - &  \small{R50}& \small{2048}\\
\hline
\small{ProxyNCA}& 62.1$\pm$0.4 & 76.2$\pm$0.4 & 86.4$\pm$0.2 & 93.6$\pm$0.3 &  \small{R50}& \small{2048}\\
\small{ProxyNCA++}& 80.7$\pm$0.5 & 92.0$\pm$0.3 & 96.7$\pm$0.1 & 98.9$\pm$0.0 &  \small{R50}& \small{512}\\
\small{ProxyNCA++}& 80.7$\pm$0.4 & 92.0$\pm$0.2 & 96.7$\pm$0.1 & 98.9$\pm$0.0 &  \small{R50}& \small{1024}\\
\small{ProxyNCA++(-max, -fast)}& 72.1$\pm$0.2 & 85.4$\pm$0.1 & 93.0$\pm$0.1 & 96.7$\pm$0.2 &  \small{R50}& \small{2048}\\
\small{ProxyNCA++}& \textbf{81.4$\pm$0.1} & \textbf{92.4$\pm$0.1} & \textbf{96.9$\pm$0.0} & \textbf{99.0$\pm$0.0} &  \small{R50}& \small{2048}\\
\hline
\end{tabular}
\caption{Recall@k for k = 1,10,100,1000 and NMI on Stanford Online Products~\cite{song2016deep}.}
\label{table:sop}
\end{table}

\begin{table}[H]
\centering
\setlength{\tabcolsep}{5pt}
\begin{tabular}{|*8c|}
\hline
R@k & 1 & 10 & 20 & 30 & 40 &  Arch & Emb \\ \hline
\small{MS~\cite{wang2019multi}}& 89.7 & 97.9 & 98.5 & 98.8 & 99.1 & \small{I3}& \small{512}\\
\small{HORDE~\cite{jacob2019metric}}& 90.4 & 97.8 & 98.4 & 98.7 & 98.9 &  \small{I3}& \small{512}\\
\small{NormSoftMax~\cite{zhai2019}}& 88.6 & 97.5 & 98.4 & 98.8 & -  & \small{R50}& \small{512}\\
\small{NormSoftMax~\cite{zhai2019}}& 89.4 & 97.8 & 98.7 & 99.0 & -  & \small{R50}& \small{2048}\\
\hline
\small{ProxyNCA}& 59.1$\pm$0.7 & 80.6$\pm$0.6 &  84.7$\pm$0.3 & 86.7$\pm$0.4 & 88.1$\pm$0.5  & \small{R50}& \small{2048}\\
\small{ProxyNCA++}& 90.4$\pm$0.2 & 98.1$\pm$0.1  &  98.8$\pm$0.0 & 99.0$\pm$0.1 & 99.2$\pm$0.0  & \small{R50}& \small{512}\\
\small{ProxyNCA++}& 90.4$\pm$0.4 & 98.1$\pm$0.1 &  98.8$\pm$0.1 & 99.1$\pm$0.1 & 99.2$\pm$0.1  & \small{R50}& \small{1024}\\
\small{ProxyNCA++}& 82.5$\pm$0.3 & 93.5$\pm$0.1 &  95.4$\pm$0.2 & 96.3$\pm$0.0 & 96.8$\pm$0.0  & \small{R50}& \small{2048}\\
\small{(-max, -fast)}&  &  &   &  & &  & \\
\small{ProxyNCA++}& \textbf{90.9$\pm$0.3} & \textbf{98.2$\pm$0.0} &  \textbf{98.9$\pm$0.0} & \textbf{99.1$\pm$0.0}& \textbf{99.4$\pm$0.0} & \small{R50}& \small{2048}\\
\hline
\end{tabular}
\caption{Recall@k for k = 1,10,20,30,40 on the In-Shop Clothing Retrieval dataset~\cite{song2016deep}.}
\label{table:inshop}
\end{table}

\subsection{Ablation Study}\label{sec:ablation}

In Table~\ref{table:cub_det}, we perform an ablation study on the performance of our proposed methods using the CUB200 dataset. The removal of the low temperature scaling component gives the most significant drop in R@1 performance (-10.8pt). This is followed by Global Max Pooling (-3.2pt), Layer Normalization (-2.6pt), Class Balanced Sampling (-2.6pt), Fast proxies (-1.9pt) and Proxy Assignment Probability (-1.1pt).

We compare the effect of the Global Max Pooling (GMP) and the Global Average Pooling (GAP) on other metric learning methodologies \cite{Schroff_2015_CVPR,wu2017sampling,wang2019multi,jacob2019metric} in Table \ref{table:max_other} on CUB200 dataset. The performance of all other models improves when GAP is replaced with GMP, with the exception of HORDE~\cite{jacob2019metric}. In HORDE, Jacob et al.~\cite{jacob2019metric} include both the pooling features as well as the higher-order moment features in the loss calculation. We speculate that since this method is designed to reduce the effect of outliers, summing max-pooled features canceled out the effect of higher-order moment features, which may have lead to sub-optimal performance.

\begin{table}[H]
\centering
\setlength{\tabcolsep}{8pt}
\begin{tabular}{|l|*5c|}
\hline
R@k & 1 & 2 & 4 & 8 & NMI \\
\hline
\small{ProxyNCA++ (Emb: 2048)} & 72.2$\pm$0.8 & 82.0$\pm$0.6 & 89.2$\pm$0.6 & 93.5$\pm$0.4 & 75.8$\pm$0.8 \\
\hspace{0.3cm}\small{-scale}&61.4$\pm$0.4 & 72.4$\pm$0.5 & 81.5$\pm$0.3 & 88.4$\pm$0.5 & 64.8$\pm$0.4 \\
\hspace{0.3cm}\small{-max}& 69.0$\pm$0.6 & 80.3$\pm$0.5 & 88.1$\pm$0.4 & 93.1$\pm$0.1 & 74.3$\pm$0.4 \\
\hspace{0.3cm}\small{-norm}& 69.6$\pm$0.3 & 80.5$\pm$0.5 & 88.0$\pm$0.2 & 93.0$\pm$0.2 & 75.2$\pm$0.4 \\
\hspace{0.3cm}\small{-cbs}& 69.6$\pm$0.6 & 80.1$\pm$0.3 & 87.7$\pm$0.3 & 92.8$\pm$0.2 & 73.4$\pm$0.3 \\
\hspace{0.3cm}\small{-fast}& 70.3$\pm$0.9 & 80.6$\pm$0.4 & 87.7$\pm$0.5 & 92.5$\pm$0.3 & 73.5$\pm$0.9 \\
\hspace{0.3cm}\small{-prob}& 71.1$\pm$0.7 & 81.1$\pm$0.3 & 87.9$\pm$0.3 & 92.6$\pm$0.3 & 73.4$\pm$0.8 \\
\hline
\end{tabular}
\caption{An ablation study of ProxyNCA++ and its enhancements on CUB200~\cite{wah2011caltech}.  }
\label{table:cub_det}
\end{table}

\begin{table}[H]
\centering
\setlength{\tabcolsep}{32pt}
\begin{tabular}{|l|c|c|}
\hline
R@1 & without prob & with prob  \\ \hline
\small{ProxyNCA (Emb: 2048)}& 59.3 $\pm$ 0.4 & 59.0 $\pm$ 0.4\\
\hspace{0.3cm}\small{+scale}& 62.9 $\pm$ 0.4 & 63.4 $\pm$ 0.6 \\
\hspace{0.3cm}\small{+scale +norm} & 65.3 $\pm$ 0.7 & 65.7 $\pm$ 0.8\\
\hspace{0.3cm}\small{+scale +max} & 65.1 $\pm$ 0.3 & 66.2  $\pm$ 0.3\\
\hspace{0.3cm}\small{+scale +norm +cbs} & 67.2 $\pm$ 0.8 & 69.1 $\pm$ 0.5 \\
\hspace{0.3cm}\small{+scale +norm +cbs +max}& 68.8 $\pm$ 0.7& 70.3 $\pm$ 0.9 \\
\hspace{0.3cm}\small{+scale +norm +cbs +max +fast} & 71.1 $\pm$ 0.7 & 72.2 $\pm$  0.8 \\
\hline
\end{tabular}
\caption{An ablation study of the effect of Proxy Assignment Probability (+prob) to ProxyNCA and its enhancements on CUB200~\cite{wah2011caltech}.}
\label{table:prob_det}
\end{table}

\begin{table}[H]
\centering
\setlength{\tabcolsep}{32pt}
\begin{tabular}{|l|c|c|}
\hline
R@1 & without scale & with scale  \\ \hline
\small{ProxyNCA (Emb: 2048)}& 59.3 $\pm$ 0.4 & 62.9  $\pm$ 0.4\\
\hspace{0.3cm}\small{+cbs}& \color{red}{54.8 $\pm$ 6.2} & 64.0 $\pm$ 0.4 \\
\hspace{0.3cm}\small{+prob}& \color{red}{59.0 $\pm$ 0.4} & 63.4 $\pm$ 0.6 \\
\hspace{0.3cm}\small{+norm}& 60.2 $\pm$ 0.6 & 65.3 $\pm$ 0.7 \\
\hspace{0.3cm}\small{+max}& 61.3 $\pm$ 0.7 & 65.1  $\pm$ 0.3\\
\hspace{0.3cm}\small{+fast}& \color{red}{56.3 $\pm$ 0.8} & 64.3  $\pm$ 0.8\\
\hspace{0.3cm}\small{+max +fast} & 60.3 $\pm$ 0.5 & 67.2 $\pm$ 0.5 \\
\hspace{0.3cm}\small{+norm +prob +cbs} & 60.4 $\pm$ 0.7 & 69.1  $\pm$ 0.5\\
\hspace{0.3cm}\small{+norm +prob +cbs +max} & 61.2 $\pm$ 0.7 & 70.3  $\pm$ 0.9\\
\hspace{0.3cm}\small{+norm +prob +cbs +max +fast}& 61.4 $\pm$ 0.4 & 72.2 $\pm$ 0.8 \\
\hline
\end{tabular}
\caption{An ablation study of the effect of low temperature scaling to ProxyNCA and its enhancements on CUB200~\cite{wah2011caltech}. Without low temperature scaling, three out of six enhancements (in red) get detrimental results when they are applied to ProxyNCA.}
\label{table:tempscale_det}
\end{table}

\begin{table}[H]
\centering
\setlength{\tabcolsep}{15pt}
\begin{tabular}{|l|c|c|}
\hline
R@1 & Global Average Pooling & Global Max Pooling  \\ \hline
\small{ProxyNCA (Emb: 2048)}& 59.3 $\pm$ 0.4 & 61.3 $\pm$ 0.7\\
\hspace{0.3cm}\small{+cbs}& 54.8 $\pm$ 6.2 & 55.5 $\pm$ 6.2  \\
\hspace{0.3cm}\small{+prob} & 59.0 $\pm$ 0.4 & 61.2 $\pm$ 0.7 \\
\hspace{0.3cm}\small{+norm} & 60.2 $\pm$ 0.6 & 60.9 $\pm$ 0.9 \\
\hspace{0.3cm}\small{+scale}& 62.9 $\pm$ 0.4 & 65.1 $\pm$ 0.3\\
\hspace{0.3cm}\small{+fast} & 56.3 $\pm$ 0.8 & 60.3 $\pm$ 0.5 \\
\hspace{0.3cm}\small{+scale +fast} & 64.3 $\pm$ 0.8 & 67.2  $\pm$ 0.5\\
\hspace{0.3cm}\small{+norm +prob +cbs} & 60.4 $\pm$ 0.7 & 61.2  $\pm$ 0.7\\
\hspace{0.3cm}\small{+norm +prob +cbs +fast} & 56.2 $\pm$ 0.9 & 61.4 $\pm$ 0.4  \\
\hspace{0.3cm}\small{+norm +prob +cbs +scale} & 69.1 $\pm$  0.5 & 70.3 $\pm$ 0.9  \\
\hspace{0.3cm}\small{+norm +prob +cbs +scale +fast} & 69.0 $\pm$  0.6 & 72.2 $\pm$ 0.8 \\
\hline
\end{tabular}
\caption{An ablation study of ProxyNCA the effect of Global Max Pooling to ProxyNCA and its enhancements on CUB200~\cite{wah2011caltech}. We can see a 2.1pp improvement on average after replacing GAP with GMP. }
\label{table:cub_det3}
\end{table}

\begin{table}[H]
\centering
\setlength{\tabcolsep}{22pt}
\begin{tabular}{|l|cccc|}
\hline
Method & Pool & R@1 & Arch & Emb \\ \hline
\small{WithoutTraining} & avg & 45.0 & \small{R50}& \small{2048}\\
 & max & \textbf{53.1} & \small{R50}& \small{2048}\\
\hline
\small{Cross-Entropy Loss} & avg & 57.2 & \small{R50}& \small{2048}\\
 & max & \textbf{66.9} & \small{R50}& \small{2048}\\
\hline
\small{ProxyNCA++} & avg & 69.0 & \small{R50}& \small{2048}\\
 & max & \textbf{72.2} & \small{R50}& \small{2048}\\
\hline
\small{Margin~\cite{wu2017sampling}} & avg & 63.3 & \small{R50}& \small{128}\\
 & max & \textbf{64.3} & \small{R50}& \small{128}\\
\hline
\small{Triplet-Semihard sampling~\cite{Schroff_2015_CVPR}} & avg & 60.5 & \small{R50}& \small{128}\\
 & max & \textbf{61.6} & \small{R50}& \small{128}\\
\hline
\small{MS~\cite{wang2019multi}} & avg & 64.9 & \small{R50}& \small{512}\\
 & max & \textbf{68.5} & \small{R50}& \small{512}\\
 \hline
 \small{MS~\cite{wang2019multi}} & avg & 65.1 & \small{I3}& \small{512}\\
 & max & \textbf{66.1} & \small{I3}& \small{512}\\
\hline
\small{Horde (Contrastive Loss)~\cite{jacob2019metric}} & avg & \textbf{65.1} & \small{I3}& \small{512}\\
 & max & 63.1 & \small{I3}& \small{512}\\
\hline
\end{tabular}
\caption{Comparing the effect of Global Max Pooling and Global Average Pooling on the CUB200 dataset for a variety of methods. 
}
\label{table:max_other}
\end{table}

\section{A comparison with NormSoftMax~\cite{zhai2019}}

In this section, we compare the differences between ProxyNCA++ and NormSoftMax~\cite{zhai2019}. Both ProxyNCA++ and NormSoftMax are proxy-based DML solutions. By borrowing notation from Equation 6 in the main paper, ProxyNCA++ has the following loss function: 

\begin{align}\label{eq:ppp}
 &L_{\text{ProxyNCA++}} = -\log\left(\frac{\text{exp}\left(-d(\frac{x_i}{||x_i||_2}, \frac{f(x_i)}{||f(x_i)||_2})*\frac{1}{T}\right)}{\sum_{f(a)\in A}^{}\text{exp}\left(-d(\frac{x_i}{||x_i||_2}, \frac{f(a)}{||f(a)||_2})*\frac{1}{T}\right)} \right)
\end{align}

\noindent{And NormSoftMax} has the following loss function:

\begin{align}\label{eq:ns}
 &L_{\text{NormSoftMax}} = -\log\left(\frac{\text{exp}\left(\frac{x_i}{||x_i||_2}^\top \frac{f(x_i)}{||f(x_i)||_2}*\frac{1}{T}\right)}{\sum_{f(a)\in A}^{}\text{exp}\left(\frac{x_i}{||x_i||_2}^\top \frac{f(a)}{||f(a)||_2}*\frac{1}{T}\right)} \right)
\end{align}

The main difference between Equation~\ref{eq:ppp} and~\ref{eq:ns} is the distance function. In ProxyNCA++, we use a Euclidean squared distance function instead of cosine distance function. 
 
\begin{figure}[H]
  \centering
  \includegraphics[width=6.4in]{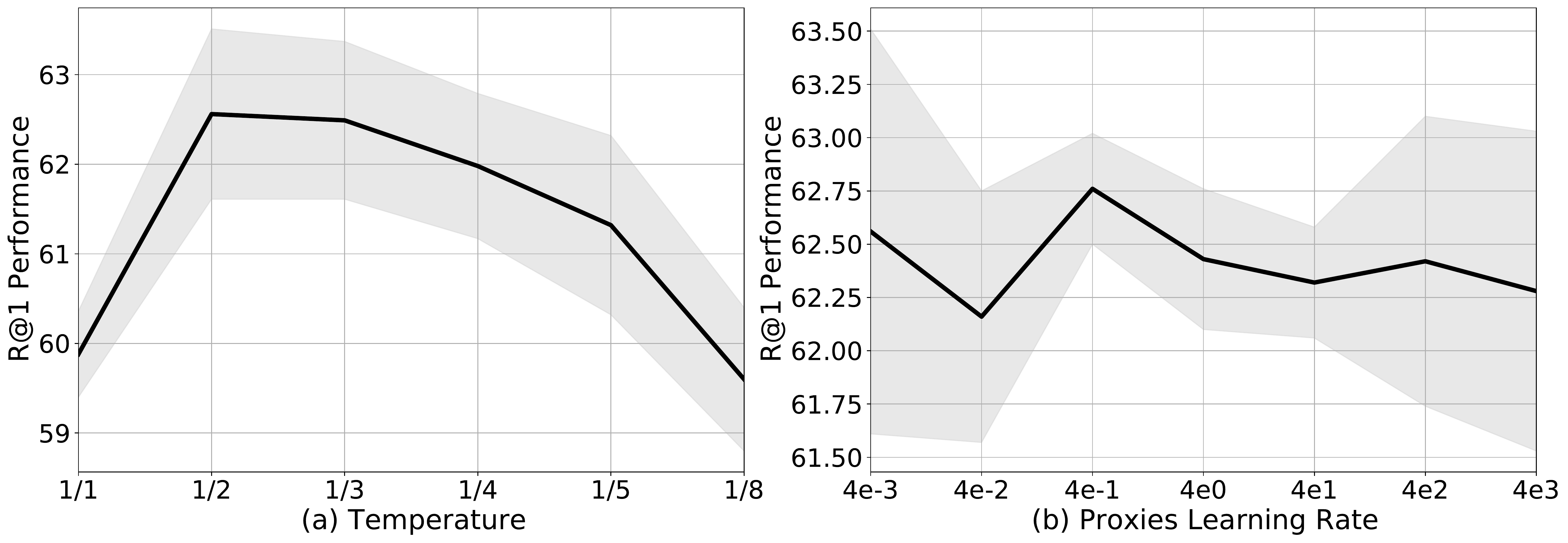}
  \caption{A sensitivity study of temperature scaling (left) and proxy learning rate (right) for NormSoftMax~\cite{zhai2019} without layer norm (norm), class balanced sampling (cbs), and fast proxies (fast). On the left panel, we show a plot of R@1 with different temperature scales on CUB200~\cite{wah2011caltech}. On the right panel, we show a plot of R@1 with different proxy learning rate with $T=1/2$ on CUB200~\cite{wah2011caltech}.
  The shaded areas represent one standard deviation of uncertainty. }
  \label{fig:n_ts}
\end{figure}

Based on our sensitivity studies on temperature scaling and proxy learning rate, we show that NormSoftMax perform best when the temperature scale, $T$ is set to $1/2$ and the proxy learning rate is set to $4e^{-1}$ (see Figure~\ref{fig:n_ts}).

We perform an ablation study of NormSoftMax in Table~\ref{table:cub_nsm} and \ref{table:cars_nsm}. On the CUB200 dataset, we show that the Global Max Pooling (GMP) component (max) improves NormSoftMax by $2.2$pp on R@1. However, the fast proxies component (fast) reacts negatively with NormSoftMax by decreasing its performance by $0.6$pp. By combining both the GMP and the fast proxies components into NormSoftMax, we see a small increase of R@1 performance (0.6pp).

On the CARS196 dataset, there is a slight increase in R@1 performance (0.3pp) by adding fast proxies component to NormSoftMax. When we add the GMP component to NormSoftMax, we observe an increase of R@1 by 1.1pp. Combining both the GMP and the fast proxies components, there is a 1.0pp increase in R@1 performance.

\begin{table}[H]
\centering
\setlength{\tabcolsep}{13pt}
\bigskip
\begin{tabular}{|l|*5c|}
\hline
R@k & 1 & 2 & 4 & 8 & NMI \\
\hline
\small{ProxyNCA++} & 72.2$\pm$0.8 & 82.0$\pm$0.6 & 89.2$\pm$0.6 & 93.5$\pm$0.4 & 75.8$\pm$0.8 \\
\hspace{0.3cm}\small{-max}& 69.0$\pm$0.6 & 80.3$\pm$0.5 & 88.1$\pm$0.4 & 93.1$\pm$0.1 & 74.3$\pm$0.4 \\
\hspace{0.3cm}\small{-fast}& 70.3$\pm$0.9 & 80.6$\pm$0.4 & 87.7$\pm$0.5 & 92.5$\pm$0.3 & 73.5$\pm$0.9 \\
\hspace{0.3cm}\small{-max -fast}& 69.1$\pm$0.5 & 79.6$\pm$0.4 & 87.3$\pm$0.3 & 92.7$\pm$0.2 & 73.3$\pm$0.7\\
\hline
\small{NormSoftMax} & 65.0$\pm$1.7 & 76.6$\pm$1.1 & 85.5$\pm$0.6 & 91.6$\pm$0.4 & 69.6$\pm$0.8 \\
\small{(+max, +fast)}& & & & &\\
\small{NormSoftMax} & 63.8$\pm$1.3 & 75.9$\pm$1.0 & 84.9$\pm$0.8 & 91.4$\pm$0.6 & 70.8$\pm$1.1 \\
\small{(+fast)}& & & & &\\
\small{NormSoftMax} & 67.6$\pm$0.4 & 78.4$\pm$0.2 & 86.7$\pm$0.4 & 92.2$\pm$0.3 & 71.2$\pm$0.9 \\
\small{(+max)}& & & & &\\
\small{NormSoftMax} & 64.4$\pm$1.1 & 76.1$\pm$0.7 & 85.0$\pm$0.7 & 91.4$\pm$0.3 & 70.0$\pm$1.1 \\
\hline
\end{tabular}
\caption{A comparison of ProxyNCA++ and NormSoftMax~\cite{zhai2019} on CUB200~\cite{wah2011caltech}.  All models are experimented with embedding size of 2048. For NormSoftMax~\cite{Zheng_2019_CVPR}, we use a temperature scaling of $T=1/2$, a proxy learning rate of $4e^{-1}$ (fast) and learning rates of $4e-3$ for the backbone and embedding layers. It is important to note that, NormSoftMax~\cite{zhai2019}  does not have max pooling and fast proxy component. }
\label{table:cub_nsm}
\end{table}

\begin{table}[H]
\centering
\setlength{\tabcolsep}{13pt}
\begin{tabular}{|l|*5c|}
\hline
R@k & 1 & 2 & 4 & 8 & NMI \\
\hline
\small{ProxyNCA++} & 90.1$\pm$0.2 & 94.5$\pm$0.2 & 97.0$\pm$0.2 & 98.4$\pm$0.1 & 76.6$\pm$0.7 \\
\hspace{0.3cm}\small{-max}& 87.8$\pm$0.6 & 93.2$\pm$0.4 & 96.3$\pm$0.2 & 98.0$\pm$0.1 & 76.4$\pm$1.3 \\
\hspace{0.3cm}\small{-fast}& 89.2$\pm$0.4 & 93.9$\pm$0.2 & 96.5$\pm$0.1 & 98.0$\pm$0.1 & 74.8$\pm$0.7\\
\hspace{0.3cm}\small{-max -fast}& 87.9$\pm$0.2 & 93.2$\pm$0.2 & 96.1$\pm$0.2 & 97.9$\pm$0.1 & 76.0$\pm$0.5\\
\hline
\small{NormSoftMax} & 86.0$\pm$0.1 & 92.0$\pm$0.1 & 95.5$\pm$0.1 & 97.6$\pm$0.1 & 68.6$\pm$0.6 \\
\small{(+max, +fast)}& & & & &\\
\small{NormSoftMax} & 85.3$\pm$0.4 & 91.6$\pm$0.3 & 95.5$\pm$0.2 & 97.6$\pm$0.1 & 72.2$\pm$0.7 \\
\small{(+fast)}& & & & &\\
\small{NormSoftMax} & 86.1$\pm$0.4 & 92.1$\pm$0.3 & 95.5$\pm$0.2 & 97.6$\pm$0.2 & 68.0$\pm$0.5  \\
\small{(+max)}& & & & &\\
\small{NormSoftMax} & 85.0$\pm$0.6 & 91.4$\pm$0.5 & 95.3$\pm$0.4 & 97.5$\pm$0.3 & 70.7$\pm$1.1 \\
\hline
\end{tabular}
\caption{A comparison of ProxyNCA++ and NormSoftMax~\cite{zhai2019} on CARS196~\cite{KrauseStarkDengFei-Fei_3DRR2013}.  All models are experimented with embedding size of 2048. For NormSoftMax~\cite{Zheng_2019_CVPR}, we use a temperature scaling of $T=1/2$, a proxy learning rate of $4e^{-1}$ (fast) and learning rates of $4e-3$ for the backbone and embedding layers. It is important to note that, NormSoftMax~\cite{zhai2019}  does not have max pooling and fast proxy component. }
\label{table:cars_nsm}
\end{table}

\section{Two moon classifier}

In Section 3.4 (About Temperature Scaling) in the main paper, we show a visualization of the effect of temperature scaling on the decision boundary of a softmax classifier on a two-moon synthetic dataset. In detail, we trained a two-layers linear model. The first layer has an input size of 2 and an output size of 100. This is followed by a ReLU unit. The second layer has an input size of 100 and an output size of 2. For the synthetic dataset, we use the scikit-learn's~\footnote[2]{\url{https://scikit-learn.org/stable/modules/generated/sklearn.datasets.make_moons.html}} moons data generator to generate 600 samples with noise of 0.3 and a random state of 0. 

\section{Regarding crop size of images}

Image crop size can have a large influence on performance. Current SOTA method~\cite{jacob2019metric} for embedding size 512 uses a crop size of $256\times256$, which we also use for our experiments (See Table~\ref{table:256}). We repeat these experiments with a crop-size of $227\times227$ to make it comparable with older SOTA method~\cite{wang2019multi} (See Table~\ref{table:227}). In this setting, we outperform SOTA for CARS and SOP. We tie on CUB, and we underperform on InShop. However, since no spread information is reported in SOTA~\cite{wang2019multi}, it is hard to make a direct comparison.

\begin{table}[H]
\centering
\setlength{\tabcolsep}{54pt}
\begin{tabular}{|l|c|c|}
\hline
R@k & SOTA~\cite{jacob2019metric} & Ours \\
\hline
\small{CUB} & 66.8 & 69.0$\pm$0.8\\
\small{CARS} & 86.2 & 86.5$\pm$0.4\\
\small{SOP} & 80.1 & 80.7$\pm$0.5\\
\small{InShop} & 90.4 & 90.4$\pm$0.2\\
\hline
\end{tabular}
\caption{A comparison of ProxyNCA++ and the current SOTA~\cite{jacob2019metric} in the embedding size of 512 and a crop size of $256\times256$.}
\label{table:256}
\end{table}

\begin{table}[H]
\centering
\setlength{\tabcolsep}{54pt}
\begin{tabular}{|l|c|c|}
\hline
R@k & SOTA~\cite{wang2019multi} & Ours \\
\hline
\small{CUB} & 65.7 & 64.7$\pm$1.6\\
\small{CARS} & 84.2 & 85.1$\pm$0.3\\
\small{SOP} & 78.2 & 79.6$\pm$0.6\\
\small{InShop} & 89.7 & 87.6$\pm$1.0\\
\hline
\end{tabular}
\caption{A comparison of ProxyNCA++ and the current SOTA~\cite{wang2019multi} in the embedding size of 512 and a crop size of $227\times227$.}
\label{table:227}
\end{table}

\section{Regarding the implementation of baseline}

We follow Algorithm 1 in the original paper~\cite{movshovitz2017no} when implementing our baseline.
In the original paper, there is an $\alpha$ variable that resembles temperature scaling. However, $\alpha$ choice is ambiguous and is used to prove the error bound theoretically. We replicate \cite{movshovitz2017no} on CUB, with embedding size of 64, crop size of 227, and GoogLeNet~\cite{I1} backbone. With the temperature scale, T=1/3, we obtain a R@1 of 49.70, which is close to the reported R@1 49.2. This indicates our implementation of ProxyNCA is correct.

The baseline ProxyNCA is implemented using the same training set up as the proposed ProxyNCA++. As mentioned in our paper (Sec 4.1), we split the original training set into the training (1st half) and validation set (2nd half). We did not perform an extensive sweep of hyperparameters. In our experiment, we first select the best hyperparameter for baseline ProxyNCA (i.e., learning rate [1e-3 to 5e-3]) before adding any enhancements corresponding to ProxyNCA++. We believe that it is possible to obtain better results for both ProxyNCA and ProxyNCA++ with a more extensive sweep of hyperparameters.

\section{Regarding the Global Max Pooling (GMP) vs. Global Average Pooling (GAP)}

\begin{figure}
  \centering
  \includegraphics[width=6.4in]{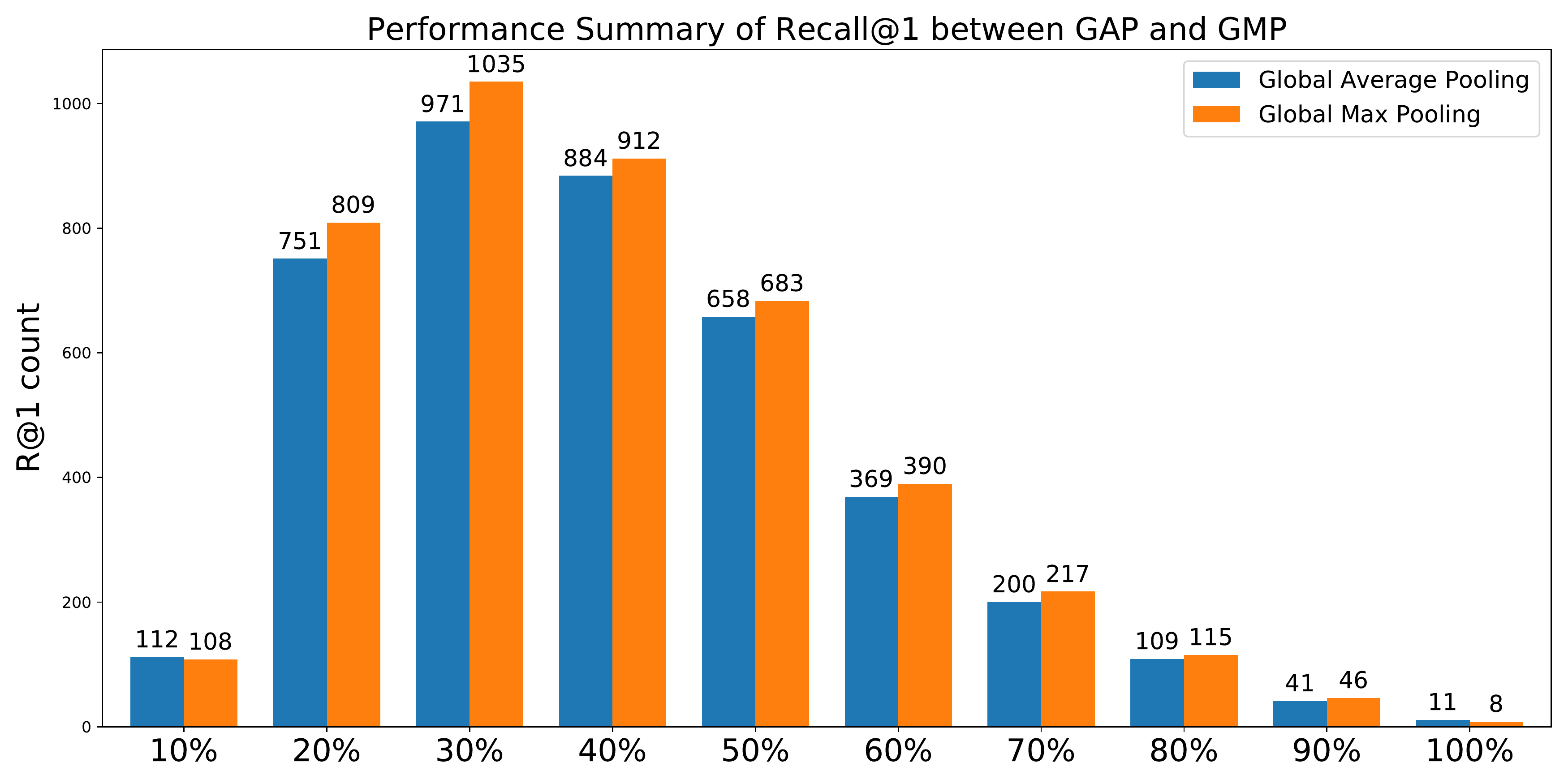}
  \caption{Performance summary (R@1) between GMP and GAP on various object sizes (in percent w.r.t. image size) in the CUB dataset.}
  \label{fig:gmp_gap_cub}
\end{figure}

In our paper, we show that GMP is better than GAP empirically. However, we could not find any consistent visual evidence as to why GMP works better. We initially hypothesized that GAP was failing for small objects. But after we controlled for object size, we did not observe any consistent visual evidence to support this hypothesis. In Figure~\ref{fig:gmp_gap_cub}, GMP consistently outperform GAP regardless of object size; this evidence disproved our initial hypothesis.  

\section{Regarding the computation complexity of ProxyNCA++}

The inference time to compute embeddings for ProxyNCA++ and baseline ProxyNCA will depend on the base architecture. In our experiments, we used a ResNet-50 model as a backbone, so inference time would be comparable to that of a ResNet-50 classifier. There are two differences which have a negligible effect on inference time: (a) The removal of the softmax classification layer, and (b) the addition of layer norm.

As for training time complexity, ProxyNCA++ is comparable to ProxyNCA both theoretically and in terms of runtime. Given a training batch size of B, we only need to compute the distance between each sample w.r.t the proxies, K. After that, we compute a cross-entropy of these distances, where we minimize the probability of a sample being assigned to its own proxy. Therefore the runtime complexity in a given batch is O(BK).

\section{Conclusion}

We revisit ProxyNCA and incorporate several enhancements. We find that low temperature scaling is a performance-critical component and explain why it works. Besides, we also discover that Global Max Pooling works better in general when compared to Global Average Pooling. Additionally, our proposed fast moving proxies also addresses small gradient issue of proxies, and this component synergizes well with low temperature scaling and Global Average pooling. 
The new and improved ProxyNCA, which we call ProxyNCA++, outperforms the original ProxyNCA by 22.9 percentage points on average across four zero-shot image retrieval datasets for Recall@1. In addition, we also achieve state-of-art results on all four benchmark datasets for all categories.
\newpage
\chapter{The GIST and RIST of Iterative Self-Training for Semi-Supervised Segmentation}\label{crv2022}

\section{Prologue}
This work has been accepted to the Canadian Conference on Computer and Robot Vision (2022). Eu Wern Teh is the first author of this publication, and the co-authors are Terrance DeVries, Brendan Duke, Ruowei Jiang, Parham Aarabi, and Graham W. Taylor.    

\section{Abstract}
We consider the task of semi-supervised semantic segmentation, where we aim to produce pixel-wise semantic object masks given only a small number of human-labeled training examples. We focus on iterative self-training methods in which we explore the behavior of self-training over multiple refinement stages. We show that iterative self-training leads to performance degradation if done na\"ively with a fixed ratio of human-labeled to pseudo-labeled training examples. We propose Greedy Iterative Self-Training (GIST) and Random Iterative Self-Training (RIST) strategies that
alternate between training on either human-labeled data or pseudo-labeled data at each refinement stage, resulting in a performance boost rather than degradation.
We further show that GIST and RIST can be combined with existing semi-supervised learning methods to boost performance. 

\section{Introduction}

Semantic segmentation is the task of producing pixel-wise semantic labels over a given image.
This is an important problem that has many useful applications such as medical imaging, robotics, scene-understanding, and autonomous driving. Supervised semantic segmentation models are effective, but they require tremendous amounts of pixel-wise labels, typically provided by a time consuming human annotation process. To overcome the need of collecting more pixel-wise labeled data, there has been
an increase in interest in
semi-supervised semantic segmentation in recent years~\cite{souly2017semi,hung2019adversarial, mittal2019semi,mendelsemi,french2019semi,olsson2020classmix,Ouali_2020_CVPR}.

Self-training is a classic semi-supervised learning method that uses pseudo-labels to guide its learning process. We define pseudo-labels as predictions generated by a given model in contrast to human-provided annotations. Self-training means using a model's own predictions as pseudo-labels in its loss during training.
Recently, there has been a resurgence of self-training methods in semi-supervised learning~\cite{lee2013pseudo,radosavovic2018data,zhai2019s4l,zoph2020rethinking,xie2020self}.
Despite the recent comeback of self-training methods, most recent self-training works are confined to only one refinement stage.
\footnote{Iterative self-training consists of multiple refinement stages, each consisting of K training iterations. At the beginning of each stage the model is initialized with weights from the previous stage, and pseudo-labels are regenerated. (see Sec.~\ref{sec2})}
We aim to investigate the behaviour of self-training after many (i.e.~$>3$) refinement stages for semi-supervised semantic segmentation.

\begin{figure}[H]
  \centering
  \includegraphics[width=6.40in, ]{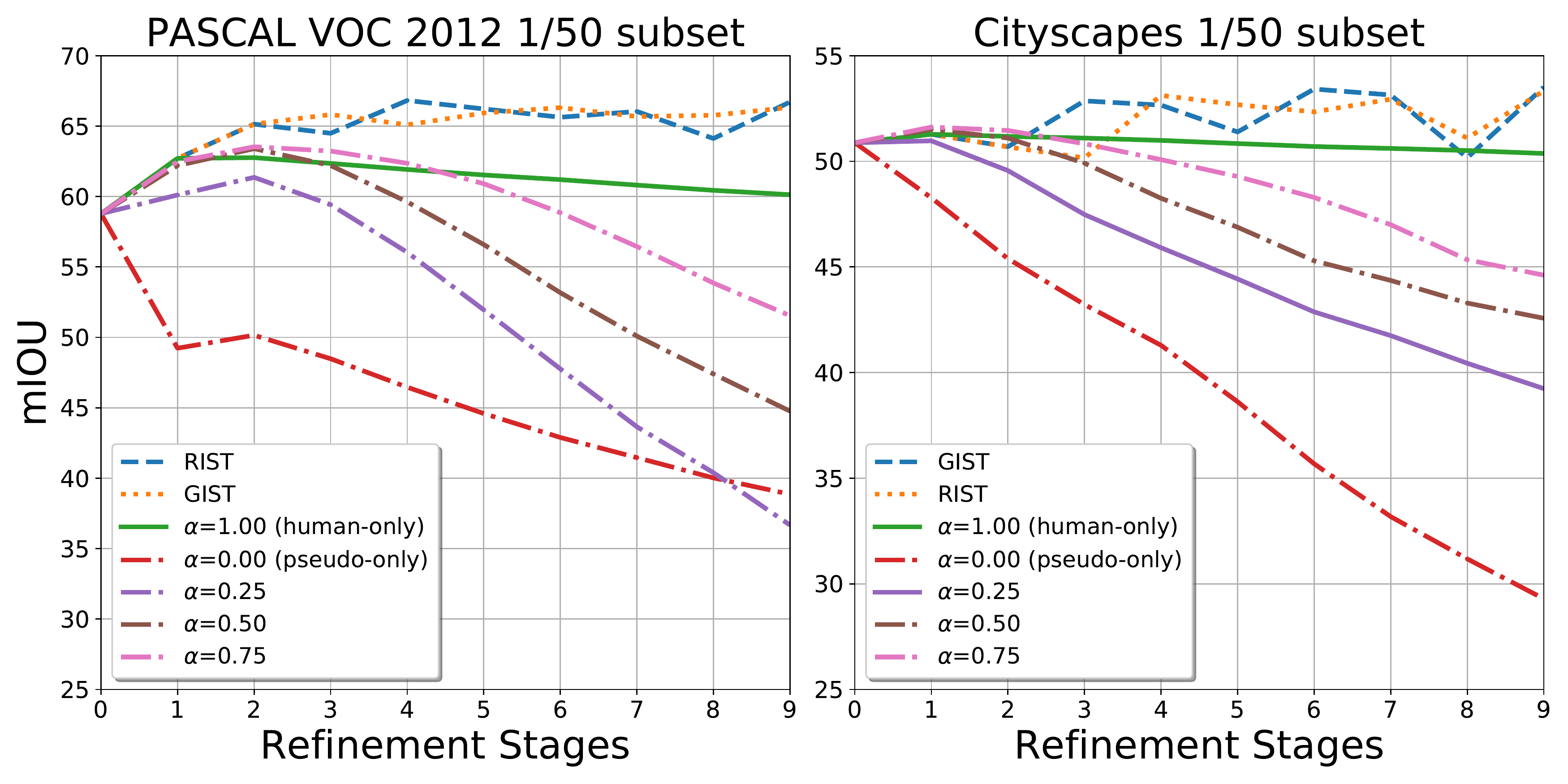}
  \caption{The performance of iterative self-training with various $\alpha$ ratios\protect\footnote[3]{} of human-labels to pseudo-labels on the PASCAL VOC 2012 and Cityscapes validation datasets. Models are refined iteratively by bootstrapping on weights trained on a previous refinement stage, with only $2\%$ of human-labels. A development set is used to select the best refinement stage.\protect\footnote[4]{}
  }
  \label{fig:voc}
\end{figure}

\begin{figure}[H]
  \centering
  \includegraphics[width=6.40in, ]{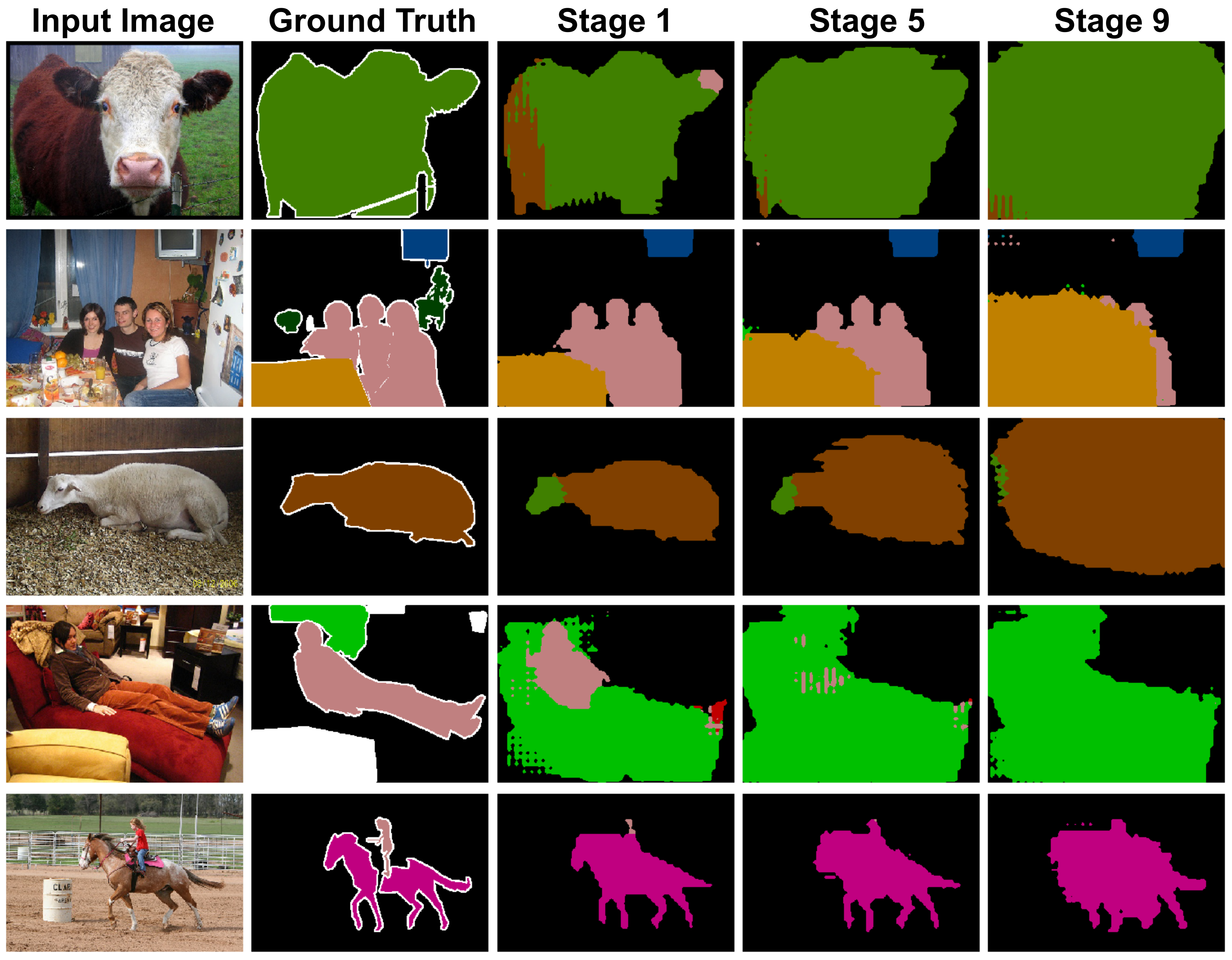}
  \caption{Pseudo-label degradation when a model is trained iteratively with a fixed human-labels to pseudo-labels ratio ($\alpha=0.75$). The first column consists of input images. The second column consists of ground truth labels. The third to fifth columns show pseudo-labels generated after at refinement stage 1, 5 and 9.
  }
  \label{fig:noisy}
\end{figure}

Iterative refinement on a small number of human-labels\footnote{For convenience, we refer to human-labeled training examples as ``human-labels'' and pseudo-labeled training examples as ``pseudo-labels''.}
may cause over-fitting on the training set, as no new information is introduced. Iterative refinement on pseudo-labels 
does introduce new information which can improve performance. However, it also results in a feedback loop that repeatedly reinforces and compounds incorrect predictions from previous iterations, ultimately resulting in ``pseudo-label bloat'', where a single dominant class prediction spreads to cover an entire image eventually (see Figure~\ref{fig:noisy}).

The na\"ive solution of combining both human-labels and pseudo-labels in each batch slows the rate at which pseudo-label bloat occurs but does not combat it entirely. Instead, we find that alternating training on only human-labels or only pseudo-labels results in a more controlled training dynamic where pseudo-labels help expand predictions to regions that may have been missed, while human-labels prevent pseudo-labels from drifting too far away from the expected annotations. By switching between these two extremes in a greedy fashion (GIST) or random fashion (RIST), our model enjoys the benefits of both label types, ultimately yielding better performance (see Figure~\ref{fig:voc}). Our contributions are the following:
\begin{itemize}
    \itemsep-0em
    \item We show that na\"ive application of iterative self-training to the problem of semi-super- vised segmentation via a fixed human-labels to pseudo-labels ratio results in significant performance degradation when $\alpha < 1$.
    \item We introduce Greedy Iterative Self-Training (GIST) and Random Iterative Self-Train- ing (RIST) to overcome performance degradation through iteratively training on either human- or pseudo-labels.
    \item  We demonstrate that both RIST and GIST can improve existing semi-supervised learning methods, yielding performance boost in both the PASCAL VOC 2012 and City- scapes datasets across all eight subsets.
\end{itemize}

\addtocounter{footnote}{1}
\footnotetext{$\alpha$ is defined as the human-labels to pseudo-labels ratio (Section 3.1.).}\stepcounter{footnote}
\footnotetext{A development set is the set of additional images used for meta-parameter selection (Sec 4 paragraph 1).}\stepcounter{footnote}

\section{Related work}\label{sec:related_work}
 
Semi-supervised semantic segmentation has gained ground in recent years. 
Souly et al.~\cite{souly2017semi} extend a typical Generative Adversarial Network (GAN) network by designing a discriminator that accepts images as input and produces pixel-wise predictions, consisting of confidence maps for each class and a fake label. Hung et al.~\cite{hung2019adversarial} improve GAN-based semantic segmentation by using a segmentation network as a conditional generator.  They also redesign the discriminator to accept the segmentation mask as input and restrict it to produce pixel-wise binary predictions. Mittal et al.~\cite{mittal2019semi} extend Hung et al.'s network by making the discriminator produce an image-level binary prediction. They also add a feature matching loss and self-training loss in their training pipeline. 
They further propose a separate semi-supervised classification network~\cite{tarvainen2017mean} to clean up segmentation masks' prediction.

Recently, a few works propose non-GAN based solutions for semi-supervised semantic segmentation. 
Mendel et al.~\cite{mendelsemi} propose an error-correcting network that aims to fix the predictions of the main segmentation network. This error-correcting network is also applied to unlabeled images to correct the generated mask.
French et al.~\cite{french2019semi} adapt CutMix~\cite{yun2019cutmix} to augment and regularize a segmentation network by creating composite images via mixing two different images and their corresponding segmentation masks.
Similarly, Olsson et al.~\cite{olsson2020classmix} also aim to regularize the segmentation network by mixing segmentation masks from two images by selecting half of the classes from one image and the other half from another image.
Ouali et al.~\cite{Ouali_2020_CVPR} minimize consistency loss between features of multiple auxiliary decoders, in which the perturbed encoder outputs are fed.

We explore the iterative self-training method to tackle semi-supervised semantic segmentation. 
Self-training or pseudo-labeling is a classic semi-supervised learning recipe that can be traced back to 1996, where it was used in an NLP application~\cite{yarowsky1995unsupervised}. A major benefit of self-training is that it allows easy extension from an existing supervised model without discarding any information.
In the deep learning literature, Lee et al.~\cite{lee2013pseudo} popularized self-training in semi-supervised classification. After its reappearance, it has gained traction in recent years. Zhai et al.~\cite{zhai2019s4l} show the effectiveness of mixing self-supervised and semi-supervised learning along with pseudo-labeling in their training regime. Zoph et al.~\cite{zoph2020rethinking} show that a randomly initialized model with self-training via a joint-loss can yield better performance than a model initialized with a pre-trained model without self-training. Xie et al.~\cite{xie2020self} show that iterative self-training with noisy labels improves a classification model's accuracy and robustness.
Radosavovic et al.~\cite{radosavovic2018data} use self-training in Omni-supervised learning, where they generate pseudo-labels by taking the average prediction of multiple perturbations of a single unlabeled image.
Most of the recent self-training works~\cite{lee2013pseudo,radosavovic2018data,zhai2019s4l,zoph2020rethinking} are confined to only one refinement stage, 
with the exception of
Xie et al.'s work~\cite{xie2020self}, where they benefit from iterative self-training by repeating self-training for three stages of refinement.
In this work, we aim to extend self-training for semi-supervised semantic segmentation and investigate self-training behavior under many (i.e.~$>3$) refinement stages.\looseness=-1

\section{Methods}\label{sec:methods}

Self-training is a semi-supervised learning method that uses pseudo-labels to guide its learning process. As we improve the model, we also improve the quality of pseudo-labels. Self-training typically consists of the following steps: (1)~training a model using human-labeled data; (2)~generating pseudo-labels using the trained model; and (3)~finetuning the trained model with the combination of human-labeled data and pseudo-labels.

\subsection{Combining both human-labels and pseudo-labels in training}\label{sec1}

Given a set of images $\{(x_i, y_i)\}$, where $x_i$ represents the image and $y_i$ represents the corresponding human-label, we begin training a segmentation model, $\text{SEG}$ for $K$ iterations using a 2D Cross-Entropy Loss, \text{ENT} on available human-labeled data. The loss for iteration $j$ is computed as:
\begin{align}
 L_j & = \frac{1}{B}\sum_{i}^{B}\text{ENT}(o_i, y_i), \label{eq:1} \\
 o_i & = \text{SEG}(x_i).\ \nonumber
\end{align}
\noindent where $B$ represents the batch size, $i$ indexes examples in a batch, and $o_i$ represents a model's output.

After a model is trained on available human-labeled data, we can now use this pre-trained model to generate pseudo-labels on unlabeled data.
Given another set of images $\{(x_i^p, y_i^p)\}$, where $x_i^p$ represents the unlabeled image and $y_i^p$ represents the corresponding pseudo-label, we can now combine human-labels and pseudo-labels by a linear combination of respective losses computed by Eq.~\ref{eq:1}.
\begin{equation}\label{eq:3}
 L_j^{\alpha} = \frac{1}{B}\sum_{i}^{B}(\text{ENT}(o_i, y_i) * \alpha + \text{ENT}(o_i^p, y^p_i) * (1 - \alpha)) .\
\end{equation}

We define $\alpha$ as the ratio of human-labels to pseudo-labels.
Eq.~\ref{eq:1} is a small modification to the classic self-training loss by Lee et al.~\cite{lee2013pseudo},
who apply a coefficient only to the unlabeled loss term, where that coefficient is iteration-dependent.

\subsection{Iterative Self-training}\label{sec2}

Self-training can be repeated through multiple refinement stages, where each refinement stage consists of a pseudo-label generation step and a finetuning step. A na\"ive solution is to fix $\alpha$ throughout all refinement stages,
which we call
Fixed Iterative Self-Training (FIST).
Given that $\alpha$ can take a floating-point number between zero and one, there are infinitely many possible $\alpha$ values at each stage of refinement. By making $\alpha$ binary, we turn the problem into discrete path selection. In this setting, our goal is to find the sequence of stages which yields the best solution. We explore two different selection strategies:  Greedy Iterative Self-Training (GIST) and Random Iterative Self-Training (RIST).


We define $S$ as the maximum number of refinement stages and $K$ as the maximum number of training iterations at a given stage. The number of possible paths in the search space ($2^S$) is exponential in the number of refinement stages.


\begin{minipage}[t]{7cm}
\vspace{-5pt}
 \begin{algorithm}[H]
   \caption{GIST}
         \label{alg:GIST}
    \begin{algorithmic}[1]
    \scriptsize
    \STATE $\theta_{list} \xleftarrow{} [\theta_{0}]; \hspace{5pt} \alpha_{list} \xleftarrow{} [0,1]$
    \FOR{$ s \xleftarrow{} 1 \ldots S$ }
        \STATE $\theta_{list}^{*} \xleftarrow{} [\ ]; \hspace{5pt} \text{R}_{list}^{} \xleftarrow{} [\ ]$
        \FOR{$\theta_c \ \ \text{in} \ \ \theta_{list}$}
            \FOR{$\alpha \ \ \text{in} \ \ \alpha_{list}$}
                \FOR{$m \xleftarrow{} 1 \ldots $ M}
                    \STATE $y^p \xleftarrow{} \argmax(\text{SEG}(x_m^p))$
                \ENDFOR

                \FOR{$j \xleftarrow{} 1 \ldots $ K}
                    \STATE $ L_j^{\alpha} \xleftarrow{} \frac{1}{B}\sum_i^B(\text{ENT}\left(o_i, y_i\right) * \alpha  +
                     \text{ENT}\left( o_i^p, y^p_i\right) * (1 - \alpha))$
                    \STATE $\theta_{c} \xleftarrow{} \theta_{c} - \lambda \frac{\partial L_j^{\alpha}}{\partial \theta_{c}}$
                \ENDFOR
                \STATE append $\theta_c$ to $\theta_{list}^{*}$
                \STATE append $\text{EVAL}(\theta_c)$ to $R_{list}$
            \ENDFOR
        \ENDFOR
        \STATE sort $\theta_{list}^{*}$ based on $R_{list}$ in descending order
        \STATE keep top $G$ items in $\theta_{list}^{*}$ and assign it to $\theta_{list}$
    \ENDFOR
\end{algorithmic}
\end{algorithm}
\end{minipage}%
\hfil 
\begin{minipage}[t]{7cm}
\vspace{-5pt}
\begin{algorithm}[H]
   \caption{RIST}
         \label{alg:RIST}
    \begin{algorithmic}[1]
    \scriptsize
    \FOR{$ s \xleftarrow{} 1 \ldots S$ }
        \STATE $\alpha \xleftarrow{} \left(\text{Uniform}(0,1) > 0.5\right)$
        \STATE $\theta_{s} \xleftarrow{} \theta_{s-1}$
        \FOR{$m \xleftarrow{} 1 \ldots $ M}
            \STATE $y^p \xleftarrow{} \argmax(\text{SEG}(x_m^p))$
        \ENDFOR

        \FOR{$j \xleftarrow{} 1 \ldots $ K}
            \STATE $ L_j^{\alpha} \xleftarrow{} \frac{1}{B}\sum_i^B(\text{ENT}\left(o_i, y_i\right) * \alpha  +   \text{ENT}\left( o_i^p, y^p_i\right) * (1 - \alpha))$
            \STATE $\theta_{s} \xleftarrow{} \theta_{s} - \lambda \frac{\partial L_j^{\alpha}}{\partial \theta_{s}}$
        \ENDFOR
    \ENDFOR
\end{algorithmic}
\end{algorithm}
\end{minipage} \newline \newline

GIST works by finding the best stage of refinement by relying on the development set.
Algorithm~\ref{alg:GIST} describes the GIST algorithm, which is a beam search strategy. In line 3, we keep a list of candidates $\theta$, and the corresponding evaluation results, $R_{list}$, using the $\text{EVAL}$ function on the development set. These lists are updated in lines 13 and 14. In line 17, we sort $\theta^*_{list}$ in descending order based on $R_{list}$. In line 18, we keep the top G $\theta$ in the $\theta^*_{list}$. and assign it to $\theta_{list}$, where it will be used as the initial model for the next stage of refinement.
In Line 6 to 8, we generate our pseudo-labels for $M$ unlabeled images. In Line 9 to 12, we finetune a segmentation model for $K$ iterations with learning rate $\lambda$.

One weakness of GIST is its smaller search space when compared to a random search (RIST). Bergstra et al.~\cite{bergstra2012random} show that in low dimensions, random search is an effective search strategy compared to a grid search. One can see that a greedy search is a subset of a grid search where we only explore the top performing branches. With a beam size of 1, the search space for a greedy search is $\log(S)$, and the search space for a random search is $2^S$. At first glance, RIST may seem counter-intuitive, but RIST's performance spread is relatively small ($\pm$0.93), and this spread can be reduced by eliminating obvious degenerate solutions and increasing search paths (see Sec.~\ref{sec:discussion} for details).
Algorithm~\ref{alg:RIST} describes the RIST algorithm.  In line 2, we randomly set $\alpha$ to either zero or one.  Line 7 to 10 is similar to the GIST algorithm (line 9 to line 12).

\begin{figure}[H]
  \centering
  \includegraphics[width=6.40in, ]{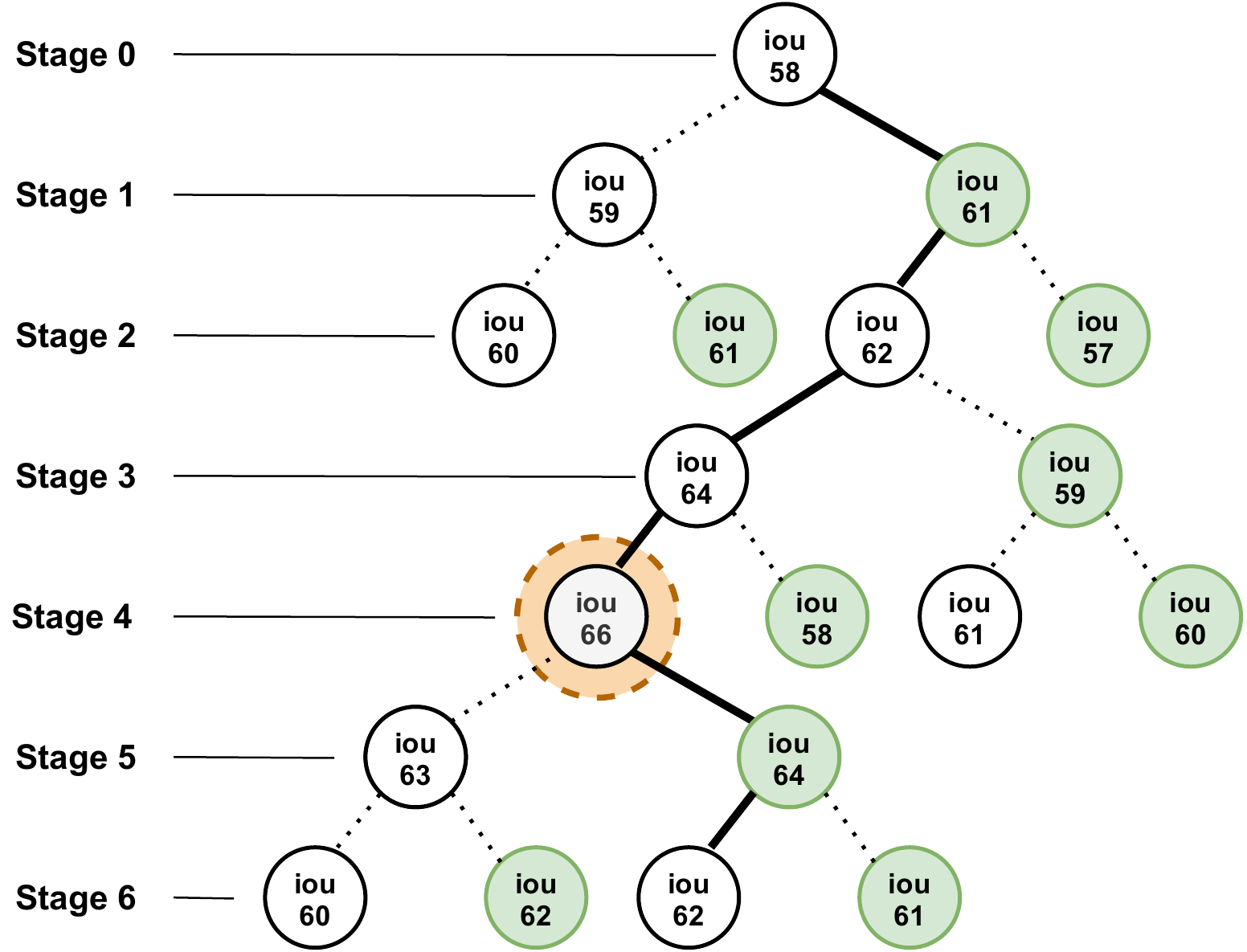}
  \caption{A hypothetical $\alpha$ path selection scenario with six refinement stages in self-training using the greedy approach (GIST). $\alpha$ indicates the ratio of human-labels to pseudo-labels. The open nodes indicate that only human-labels ($\alpha = 1$) are being used for training and the shaded nodes indicate only pseudo-labels ($\alpha = 0$) are being used. The number of possible paths is exponential in the number of refinement stages ($2^6$).
  At each stage of refinement, we evaluate the mean intersection over union (mIOU) of a model using a development set. Here, the optimum value is found at stage four of the refinement process.
  }
  \label{fig:sel}
\end{figure}

In GIST, the time complexity of a beam search is $O(S \cdot G)$. A beam search is only as efficient as a random search if we can parallelize the training for each $\alpha$ at each stage, and this condition requires a beam search to be trained on consecutive numbers of GPU resources. Unlike a beam search, each random search can be trained independently on a single GPU resource. It is easier and cheaper to obtain $N$ independent GPUs rather than $N$ consecutive GPU resources, making RIST computationally cheaper and faster in practical settings. Figure~\ref{fig:sel} shows a hypothetical $\alpha$ path selection scenario using GIST.

\subsection{Additional Add-ons}

In addition to cross entropy loss in Eq~\ref{eq:3}, we include three additional add-ons to boost the performance of FIST, GIST and RIST: Consistency Loss (CL), Label Erase (LE), and Temperature Scaling (TS). CL is a common loss in semi-supervised learning~\cite{mittal2019semi,olsson2020classmix}. Mittal et al.~\cite{mittal2019semi} use CL via a feature matching loss to minimize the discrepancy between predicted features and ground truth features. Olsson et al.~\cite{mittal2019semi} use CL via mean-teacher~\cite{tarvainen2017mean} method. We also use the mean-teacher method as our CL loss.
Additionally, LE is also used in ~\cite{mittal2019semi,olsson2020classmix} during pseudo-label generation to filter out noisy pseudo-labels by thresholding the output activation. If a given pixel has activation below a certain threshold, we ignore the loss contributed by this pixel.
Finally, we employ TS~\cite{hinton2015distilling} to overcome over-confident prediction in our model, where the activation values after softmax are highly skewed towards $100\%$ (see Figure~\ref{fig:ts}).  We include the implementation details and ablation studies for CL, LE, and TS in the supplementary material. All of our experiments with FIST, GIST, and RIST include all add-ons.

\begin{figure}[H]
  \centering
  \includegraphics[width=6.4in, ]{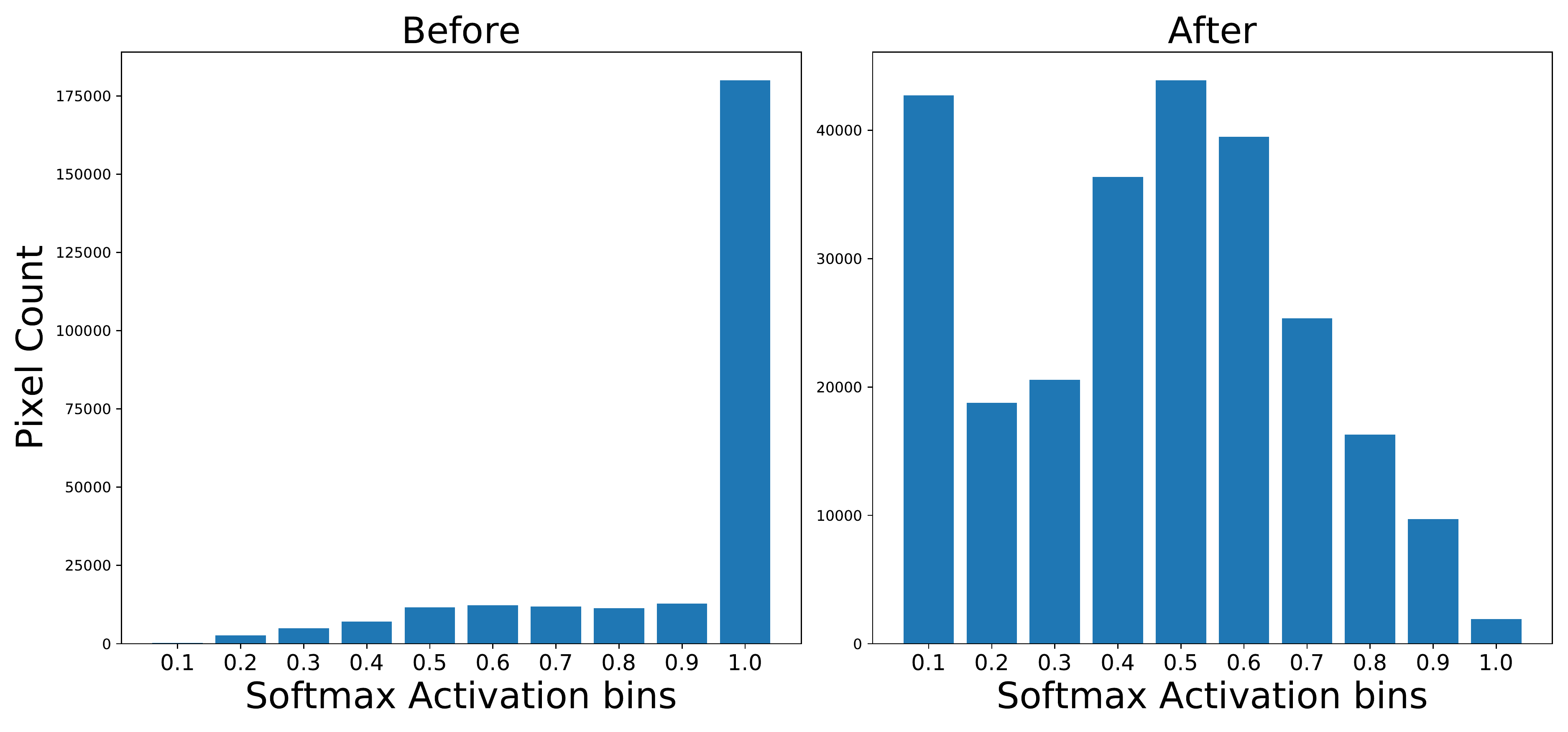}
  \caption{The before and after effects of applying temperature scaling to the output activation of a single image. A temperature scale of 0.2 is applied to the output activation.
  }
  \label{fig:ts}
\end{figure}

\section{Consistency Loss (CL)}

Consistency Loss (CL) is a common loss in semi-supervised learning where the goal is to minimize features between two perturbations of the same input. Our CL is  similar to the previously proposed Mean Teacher semi-supervised learning methods for classification~\cite{tarvainen2017mean}.
We use the following equations (Equations~\ref{eq:mt} to~\ref{eq:f2}) to calculate CL for both $x_i$ and $x_i^p$. 

\setcounter{equation}{2}
\begin{align}
 & \theta^* = \theta^ * \beta + \theta * (1 - \beta) )\label{eq:mt}\\
 & o_i,f_i = \text{SEG}(x_i)\label{eq:x1}\\
 & o_i^*,f_i^* = \text{SEG}^*(x_i)\label{eq:x2}\\
 & f_i = \text{drop}(\text{pool}(f_i))\label{eq:f1}\\
 & f_i^* = \text{drop}(\text{pool}(f_i^*))\label{eq:f2}\\
 &  L_{\text{feature}} = | f_i - f_i^*|
\end{align}

\begin{figure}[htb]
  \centering
  \includegraphics[width=6.40in, ]{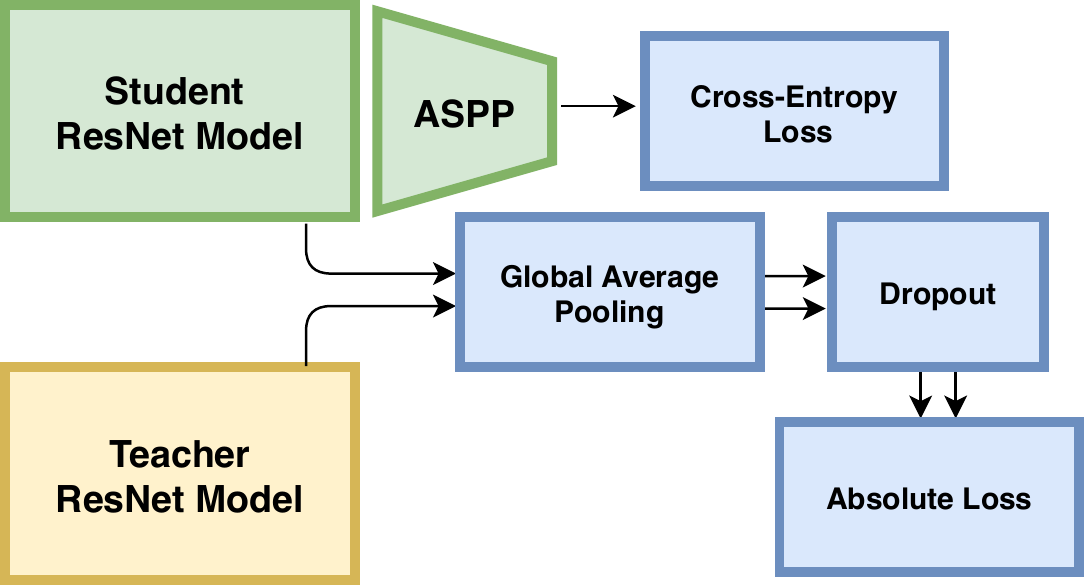}
  \caption{Depiction of the Consistency Loss (CL). At each batch of training, there are two copies of Resnet Models. The student model is the main model that is used to predict the segmentation mask, and the teacher model is a copy of the student model with an exponential moving weight update.  The goal of CL is to minimize the differences between the features of the student and teacher models.}
  \label{fig:cl}
\end{figure}

Following ~\cite{mittal2019semi,hung2019adversarial}, we use DeepLabV2~\cite{chen2017deeplab} as our segmentation model; therefore, $f_i$ represents features before an Atrous Spatial Pyramid Pooling (ASPP) layer. Equation~\ref{eq:mt} describes the weight update rule for the teacher model, $\text{SEG}^*$, where an exponential moving average rule is applied to its weights, $\theta^*$, which is controlled by $\beta$, $(0 \leq \beta \leq 1)$.
Equations~\ref{eq:x1} and \ref{eq:x2} illustrate the extraction of features before the ASPP layer in both the student model, $\text{SEG}$, and the teacher model, $\text{SEG*}$. In Equations~\ref{eq:f1} and \ref{eq:f2}, we apply global average pooling followed by a dropout perturbation to these features. After that, we compute the absolute differences between these two features.
Figure~\ref{fig:cl} summarizes the feature consistent loss in our model.

\section{Label Erase (LE)}

Pseudo-labels are often very noisy, especially when only a small amount of human-labeled data is present. 
To remove noise from pseudo-labels, we only keep predictions that pass a certain confidence threshold, $\phi$. This technique was also used in other self-training works~\cite{radosavovic2018data,zoph2020rethinking}.

Equation~\ref{eq:conf} illustrates the process of flagging low confidence prediction regions so that they are ignored by the loss function in Equation 1. We define pixel-wise confidence, $c_{i,j}^p$ as the pixel softmax output of $o_{i,j}^p$, where $i$ represents the image index and $j$ represents the pixel index.

\begin{align}\label{eq:conf}
        y_{i,j}^p = \begin{cases}
            \argmax(c_{i,j}^p), & \text{if}\; \max(c_{i,j}^p) >= \phi,\\
            \text{ignore label}, & \text{if}\; \max(c_{i,j}^p) < \phi,
        \end{cases}
\end{align}

\section{Temperature Scaling (TS)}

Temperature scaling (TS) is introduced by Hinton et al.~\cite{hinton2015distilling}, where it is used to create a softer probability distribution for knowledge distillation. The formula for TS is defined as 
$q_{i} = \frac{\exp(y_i*T)}{\sum_j \exp(y_j*T)}$. As $T$ becomes smaller, the output of the softmax function will tend towards uniform distribution. We use TS to regularize the model's confidence during training. 

\section{CL, LE, and TS hyperparameters}

For CL, we set the $\beta$ value to 0.9 and dropout probability to 0.5 for both the PASCAL VOC 2012 and Cityscapes dataset. For LE, we set the threshold value, $\phi$ to 0.999 for the Pascal VOC 2012 dataset, and we set the threshold value, $\phi$ to 0.9 for the Cityscapes dataset. For TS, we set the temperature value, $T$ to 0.1 for the Pascal VOC 2012 dataset. For the Cityscapes dataset, we set the $T=0.5$ for $1/50$ subset, $T=0.9$ for $1/20$ subset, and $T=1$ for $1/8$ subset.

\section{Experiments}\label{sec:experiment}

We perform experiments on two semantic segmentation datasets: PASCAL VOC 2012~\cite{pascal-voc-2012} and Cityscapes~\cite{Cordts2016Cityscapes}. In each dataset, there are three subsets, where pre-defined ratios ($1/50$, $1/20$, and $1/8$) of training images are selected as images with human-labels. We also experiment on two additional subsets ($1/30$ and $1/4$) for the Cityscapes dataset. These subsets of labeled images are selected using the same split as ~\cite{hung2019adversarial,mittal2019semi,olsson2020classmix}. We treat the remaining images as unlabeled examples.  We use the mean intersection-over-union (mIoU) as a performance metric. The validation images for both datasets are set aside and used for evaluation, which is consistent with~\cite{hung2019adversarial, mittal2019semi,olsson2020classmix}. We select 50 additional images from the training set as a modest ``development set'' for meta-parameter tuning.

For all of our experiments, we follow the same experimental setup as \cite{hung2019adversarial,mittal2019semi,olsson2020classmix}. We use a DeepLabV2~\cite{chen2017deeplab} segmentation model that is initialized with MS-COCO pre-trained weights~\cite{lin2014microsoft}~\footnote[5]{DeepLabV2 backbone is used in our experiments so that our method is comparable to S4GAN and ClassMix. }. We also freeze all the BatchNorm Layers in DeepLabV2. We optimize our model using the Stochastic Gradient Descent (SGD) optimizer with a base learning rate of 2.5e-4, a momentum of 0.9, and a weight decay of 5e-4. We use a polynomial learning rate decay policy, where we adjust the learning rate with the following equation:
$\lambda_{\text{iter}} = \lambda_0 (1- \frac{\text{iter}}{\text{max\_iter}})^{0.9}$ where $\lambda_0$ is a base learning rate. To augment the dataset, we use random-cropping ($321\times321$ for PASCAL VOC 2012 and $256\times512$ for Cityscapes), horizontal-flipping (with a probability of 0.5), and random-resizing (with a range of 0.5 to 1.5) in all of our experiments.

For all subsets, we train our supervised models and stage-0 models for 25,000 iterations. 
Additionally, for the Pascal VOC 2012 dataset, we use a batch-size of 8, and we refine our model for 3,000 iterations at each refinement stage (Stage 1 to 9). For the Cityscapes dataset, we use a batch-size of 6, and we refine our models for 4,000 iterations at each refinement stage (Stage 1 to 9). We use a search cost of two for both GIST and RIST and use the development set to select the best path for all experiments.

Table~\ref{table:det} shows the results of our experiments as well as relevant
results that others have reported~\cite{hung2019adversarial,mittal2019semi,french2019semi,olsson2020classmix}. We use the code provided by the respective authors for our experiments with S4GAN~\cite{mittal2019semi} and ClassMix \cite{olsson2020classmix}. For a fair comparison, we set the batch size for S4GAN and ClassMix to match with our experiments, and we also select the best iterations
using our development set. 
We notice performance differences in S4GAN and ClassMix when compared to performances reported in the original papers. We speculate that the differences are caused by best iteration selection and batch sizes (see supplementary material for details).
For experiment with S4GAN+GIST/RIST and ClassMix+GIST/RIST, we 
first train the segmentation model with S4GAN and ClassMix algorithm (stage-0 models). After that, we further refine the segmentation model using the GIST/RIST algorithm by bootstrapping on the DeepLabV2 model trained with S4GAN/ClassMix.

\begin{table*}[h]
\centering
\scalebox{1.0}{
\setlength{\tabcolsep}{6pt}
\begin{tabular}{|c||c|c|c||c|c|c|c|c|}
\hline
&\multicolumn{3}{c||}{VOC 2012 ($\approx$10k images)}&\multicolumn{5}{c|}{Cityscapes ($\approx$3k images)} \\ \hline \hline
$\#$ of labeled images & 211 & 529 & 1,322 & 59 & 100 & 148 & 371 & 743  \\ \hline
Subset & 1/50 & 1/20 & 1/8 & 1/50 & 1/30 & 1/20 & 1/8 & 1/4   \\ \hline
Supervised & 54.15 & 62.94 & 67.44 & 49.68 & 53.96 & 54.71 & 59.90 & 62.21 \\
Supervised + GIST & 66.33 & 66.95 & 70.27 & \textbf{53.51} & 56.38 & 58.11 & 60.94 & 63.04 \\
Supervised + RIST & 66.71 & 68.28 & 69.90 &  53.33 & 56.28 & 57.81 & 61.38 & 63.92 \\

\hline
S4GAN~\cite{mittal2019semi} & 62.87 & 62.35 & 68.56 & 50.48 & 54.58 & 55.61 & 60.95 & 61.30 \\
S4GAN~\cite{mittal2019semi} + GIST & \textbf{67.21} & 68.50 & 70.61 & 52.36 & 57.18 & 57.40 & 61.27 & 64.24 \\
S4GAN~\cite{mittal2019semi} + RIST & 66.51 & 68.50 & 70.31 & 53.47 & 57.12 & 57.48 & 62.50 & 64.64 \\
\hline
ClassMix~\cite{olsson2020classmix} & 63.63 & 66.74 & 66.14 & 52.14 & 57.02 & 58.77 & 61.56 & 63.90 \\
ClassMix~\cite{olsson2020classmix} + GIST & 65.60 & 69.05 & 70.65 & 52.43 & \textbf{58.70} &  \textbf{59.98} & 62.44 & 64.53 \\
ClassMix~\cite{olsson2020classmix} + RIST & 66.30 & \textbf{69.40} & \textbf{70.76} & 53.05 & 58.55 & 59.54 & \textbf{62.57} & \textbf{65.14} \\
\hline 
\end{tabular}
}

\caption{Semantic segmentation results (mIoU) on the PASCAL VOC 2012 and Cityscapes validation datasets.}
\label{table:det}
\end{table*}
\subsection{Discussion}\label{sec:discussion}

Figure~\ref{fig:voc} shows that a na\"ive application of iterative self-training leads to significant performance degradation in both datasets. Figure~\ref{fig:noisy} shows the qualitative evidence of performance degradation in FIST at $\alpha=0.75$. FIST suffers from over-confident pseudo-label predictions which spread to the surrounding pixel. Over multiple stages of refinement, the pseudo-labels expand and eventually engulf most of the image. We speculate that this may be why most of the recent self-training works are confined to one refinement stage.

\begin{figure}[H]
  \centering
  \includegraphics[width=6.40in, ]{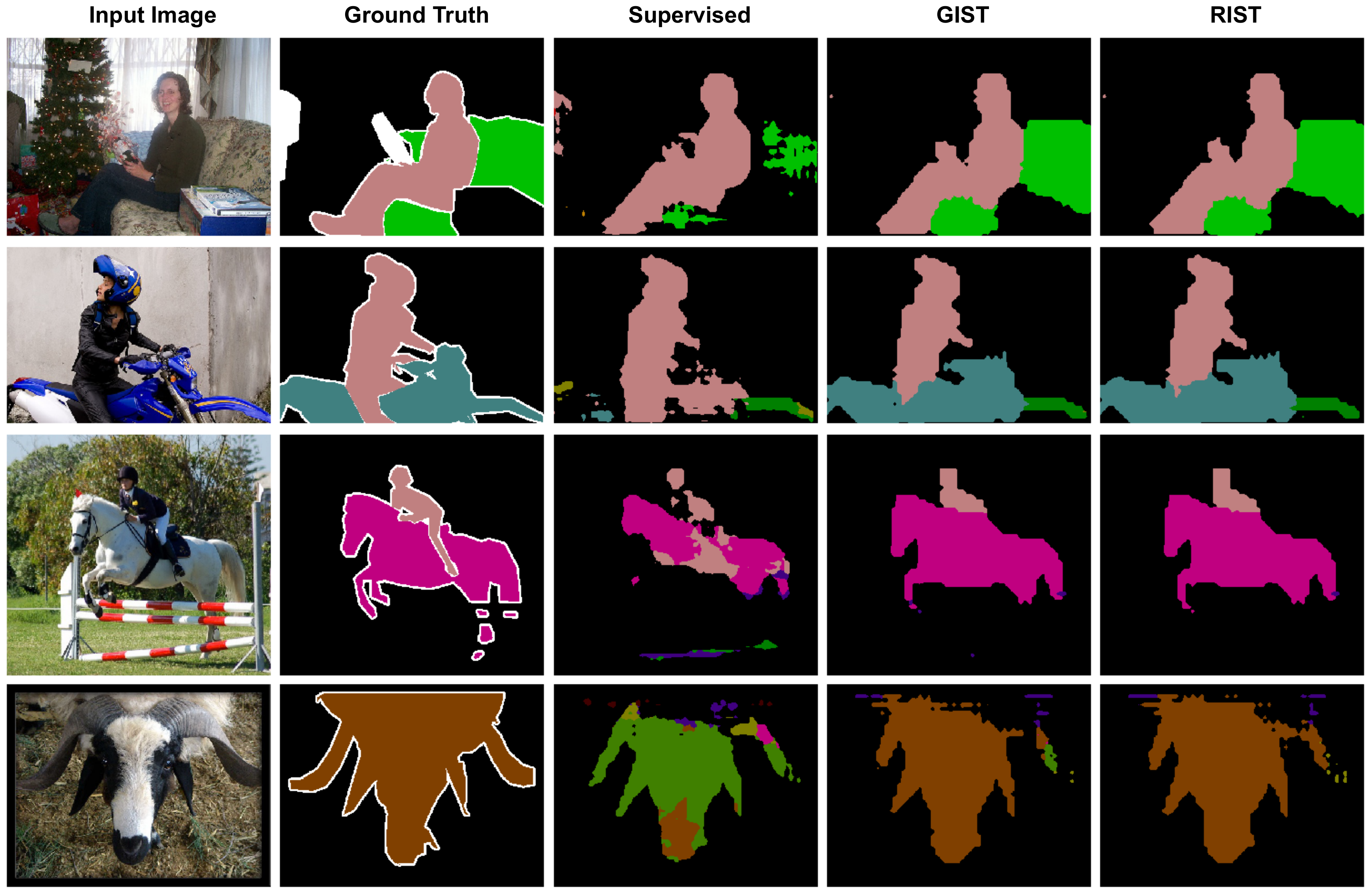}
  \caption{The qualitative results of our model train on $2\%$ human-labels from the PASCAL VOC 2012 dataset.}
  \label{fig:voc_vis}
\end{figure}

\begin{figure}[H]
  \centering
  \includegraphics[width=6.40in, ]{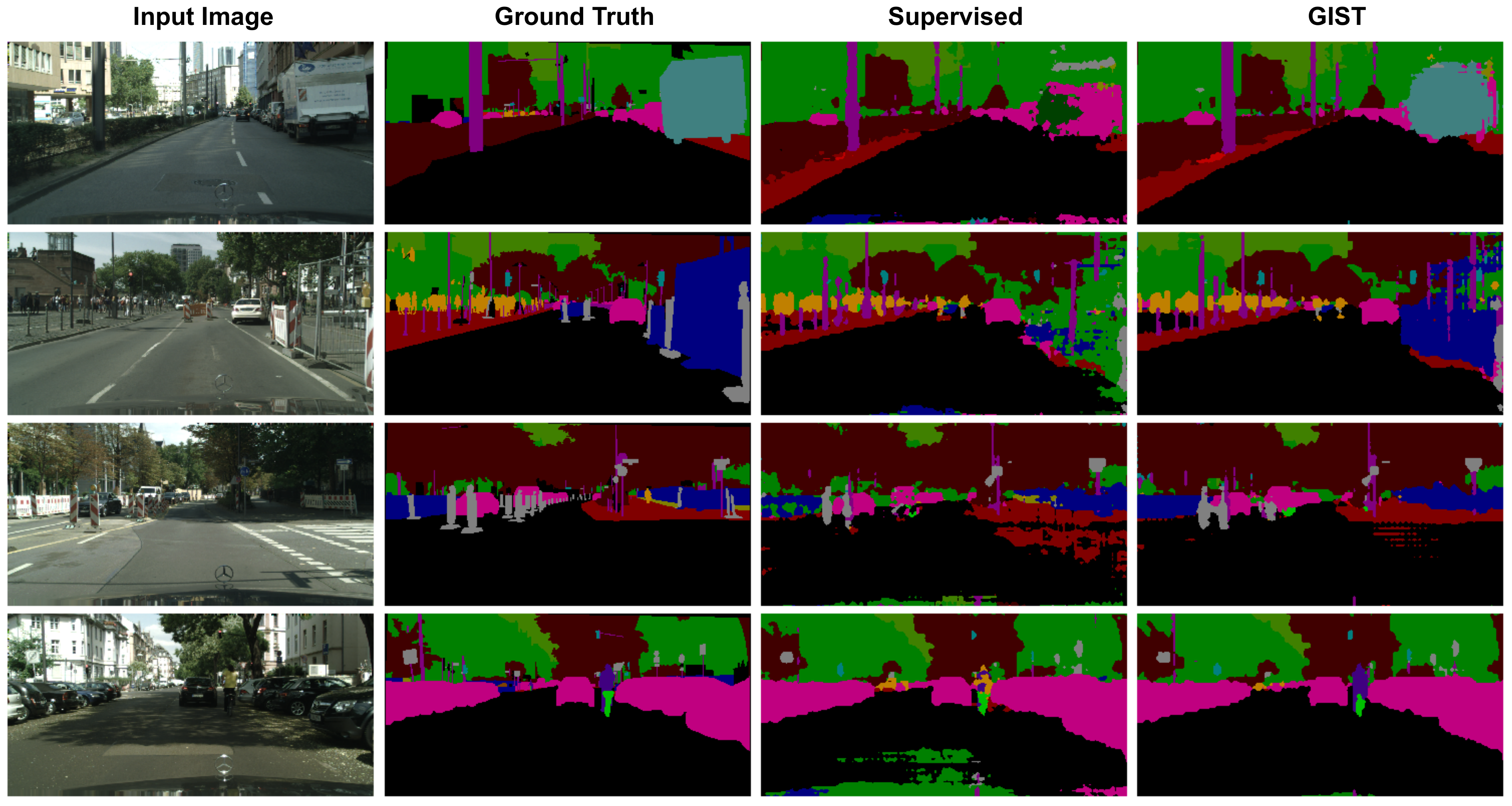}
  \caption{The qualitative results of our model trained on a $2\%$ human-labels from the Cityscapes dataset.}
  \label{fig:city_vis}
\end{figure}

Figure~\ref{fig:voc} also shows that both RIST and GIST overcome performance degradation. Additionally, both RIST and GIST generalize better than FIST in each successive refinement.
Table~\ref{table:det} shows that both RIST and GIST can improve other semi-supervised segmentation techniques such as S4GAN and ClassMix. 
We show that both RIST and GIST can further refine S4GAN and ClassMix yielding performance boosts across all subsets in the Pascal VOC 2012 and Cityscapes datasets.
Figures~\ref{fig:voc_vis} and~\ref{fig:city_vis} show RIST and GIST's qualitative results that are trained with $2\%$ of human-labels in both datasets. 

\begin{figure}[!htb]
  \centering
  \includegraphics[width=5.40in, ]{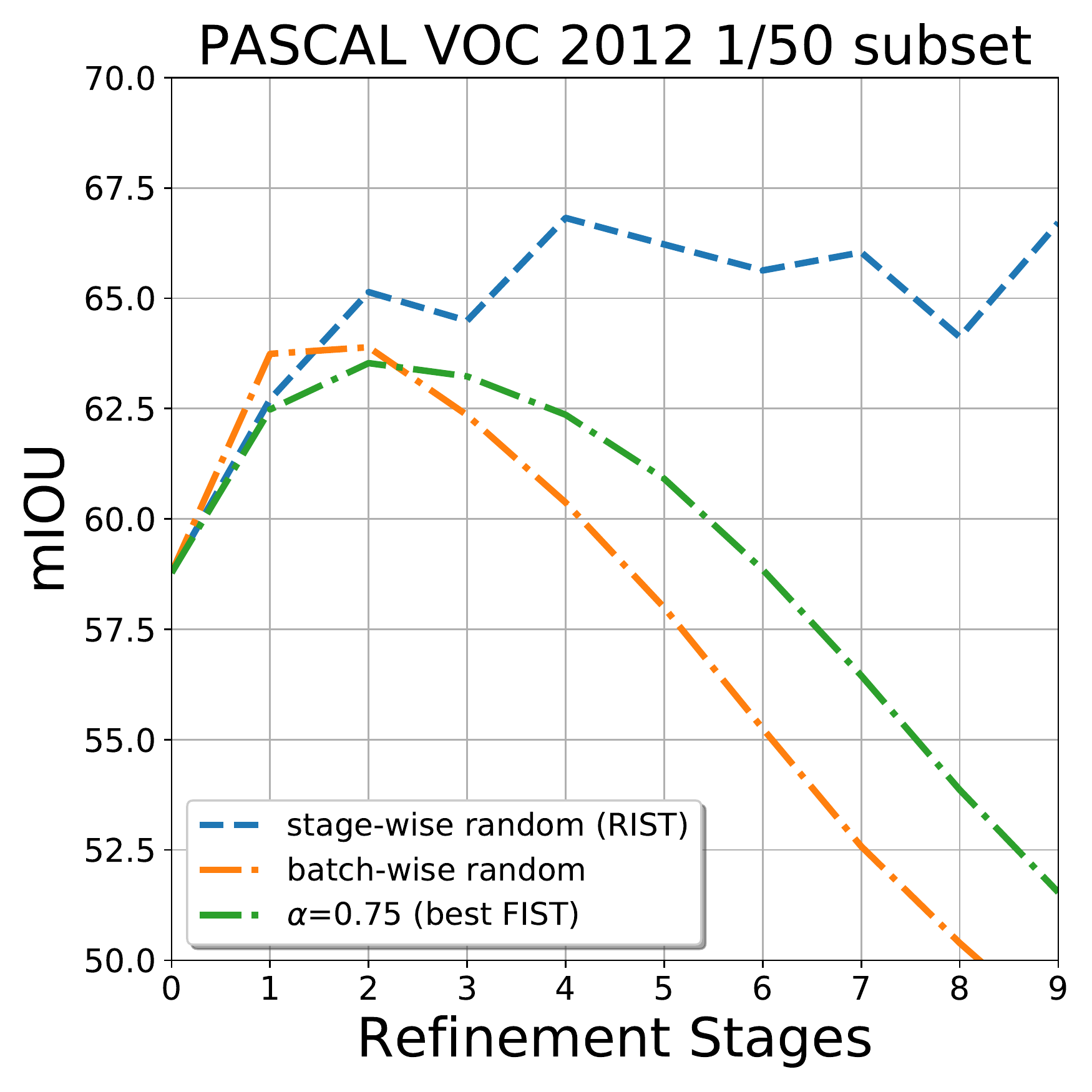}
  \caption{Self-training performance at various stages between batch-wise and stage-wise random selection strategies on the PASCAL VOC 2012 validation set.}
  \label{fig:batch}
\end{figure}

Figure~\ref{fig:batch} demonstrates the importance of training on a single label type (i.e., human-label or pseudo-label) for an extended number of iterations. We find that randomly selecting the label type for each batch (batch-wise random) performs just as poorly as training with both label types in each batch (FIST). We speculate that the clean human-labels and the noisy pseudo-labels represent competing objectives, which are difficult for the model to satisfy simultaneously, resulting in it getting stuck at poor solutions. By applying stage-wise training, we allow the model to focus on a single objective at a time, potentially escaping sub-optimal solutions from a previous stage.

Table~\ref{table:stable} examines the stability of performance for RIST on the PASCAL VOC 2012 1/50 subset. We use the same subset for training and generate fifteen different permutations of binary $\alpha$ values uniformly at random.
The degenerate solutions occur in row 5 and 6, where we have more than four consecutive numbers of the same $\alpha$ choice during training. If we were to perform a random search once, the standard deviation of mIoU is 0.93. If we were to remove the obvious degenerate solutions (row 5 and 6) using some heuristics, the standard deviation of mIoU is reduced to 0.52. Nevertheless, we can reduce this standard deviation further by selecting the best solution out of five different random solutions yielding a standard deviation of 0.23.

\begin{table}[H]
\centering
\scalebox{1.0}{
\setlength{\tabcolsep}{20pt}
\begin{tabular}{|l|c|c|c|}
\hline
Selection & mIoU (devel) & mIoU (val) & Group mIoU (val) \\ \hline
PPLLPLPLL & 54.35 & 66.14 & 67.03 \\
LPPLLLPPL & 55.05 & 66.71 &   \\
PPLLPPPLP & 54.21 & 66.05 &   \\
LPLPPLLPL & 55.12 & 67.03 &   \\
LLLLLLPPP & 54.43 & 63.75 &  \\
\hline
PPPPPPLLP & 50.43 & 64.37 & 66.63  \\
PPLLLPLPL & 54.00 & 66.02 &   \\
LLPLPLLPL & 55.19 & 66.24 &   \\
LPLPLLPLP & 55.29 & 66.63 &   \\
LLLPLPPLP & 54.88 & 65.17 &   \\
\hline
PLPLPPPLP & 54.64 & 66.40 & 66.62 \\
LPPLPPLPP & 54.46 & 66.83 &   \\
LPPLLLLPP & 54.67 & 65.61 &   \\
LPPPLLLLP & 54.82 & 66.62 &  \\
LPPPLLPPL & 54.64 & 66.61 &   \\
\hline
Mean$\pm$1 std.~dev.&  & 66.01$\pm$0.93 & 66.76$\pm$0.23 \\
\hline
\end{tabular}
}
\caption{RIST semantic segmentation results on the PASCAL VOC 2012 validation set. Group experiment results are selected based on the best development mIoU. Random selection choices are P (pseudo-label only) or L (human-label only). There are a total of nine refinement stages ordered sequentially from left to right. }
\label{table:stable}
\end{table}

\begin{table}[H]
\centering
\setlength{\tabcolsep}{43pt}
\scalebox{1.0}{
\begin{tabular}{|l|c|c|}
\hline
Sample Size & mIoU (RIST) & mIoU (GIST) \\ \hline
10 & 66.04 & 66.23\\
25 & 65.63 & 62.76\\
50 & 66.71 & 66.33\\
100 & 66.71 & 66.67 \\
200 & 66.71 & 66.69\\
500 & 66.71 & 67.00 \\
\hline
\end{tabular}
}
\caption{Semantic segmentation results on the PASCAL VOC 2012 validation set for RIST and GIST based on best epoch selected while varying the number of examples in the development set.}
\label{table:devel}
\end{table}

\begin{table}[H]
\centering
\setlength{\tabcolsep}{28pt}
\scalebox{1.0}{
\begin{tabular}{|p{1cm}|p{1cm}|p{1cm}|p{1cm}|c|}
\hline
Beam Size & Search Cost & mIoU (devel) & mIoU (val) & Solution \\ \hline
1 & 2 & 55.17 & 66.33 & LPLLPLLPL\\
2 & 4 & 55.46 & 64.95 & LLPLLPLLP\\
3 & 6 & 55.46 & 64.95 & LLPLLPLLP\\
\hline
\end{tabular}
}
\caption{Semantic segmentation results on the PASCAL VOC 2012 validation set for GIST on different beam size. selection choices at each stage are P (pseudo-label only) or L (human-label only).}
\label{table:gist_beam}
\end{table}

Table~\ref{table:devel} explores the sensitivity of RIST and GIST's performance on the PASCAL VOC 2012 $1/50$ subset with respect to the size of the development set. In general, RIST and GIST are relatively stable. RIST performance remains the same for 50 to 500 sample sizes. GIST performance improves slightly as we increase the sample size from 50 to 500.
On the Pascal VOC 2012 dataset, we show that our supervised+GIST and supervised+RIST trained with 211 human-labels ($1/50$ subset) outperform the supervised model that is trained with 529 human-labels ($1/20$ subset). This result shows that GIST and RIST improve the model, not just because they got more supervised signals from the development set (see Table~\ref{table:det}).

Table~\ref{table:gist_beam} explores GIST at various beam sizes. GIST can find a better solution for the development set; however, since there is a mismatch between the distribution of the development set and the original validation set due to the small sample size, the best solution of the development set is not the best solution for the original validation set. This study shows that GIST has a higher chance to overfit the development set when compared to RIST.

\section{Additional experiments}

\begin{figure}[H]
  \centering
  \includegraphics[width=5.40in, ]{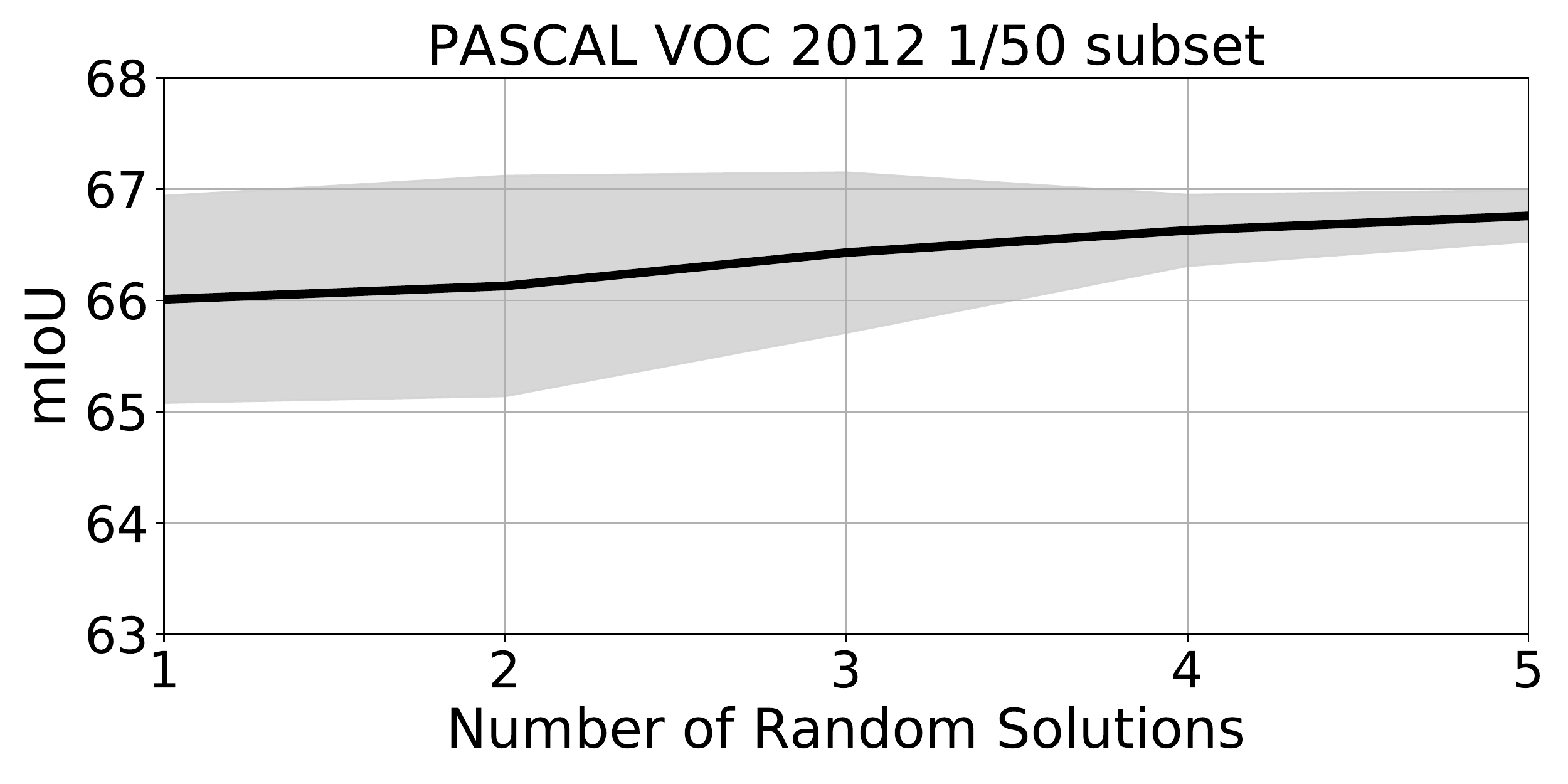}
  \caption{The effect of increasing stability of RIST when we increase the number of random solution. The best results of the random solutions are selected based on best development mIoU. The shaded area represent one standard deviation of uncertainty.
  }
  \label{fig:stable}
\end{figure}

Table~\ref{table:det2} shows the comparison of FIST, GIST, and RIST. We show that both GIST and RIST outperformed FIST. 
Figure~\ref{fig:stable} shows that as we increase the number of random solutions, the spread decreases. 
The ablation study of the add-ons' effect is shown in Table~\ref{table:addons}.

\begin{table*}[htb]
\centering
\setlength{\tabcolsep}{14pt}
\begin{tabular}{|c||c|c|c||c|c|c|}
\hline
&\multicolumn{3}{c||}{VOC 2012 ($\approx$10k images)}&\multicolumn{3}{c|}{Cityscapes ($\approx$3k images)} \\ \hline \hline
$\#$ of labeled images & 211 & 529 & 1322 & 59 & 148 & 371 \\ \hline
Subset & 1/50 & 1/20 & 1/8 & 1/50 & 1/20 & 1/8   \\ \hline
Supervised & 54.15 & 62.94 & 67.44 & 49.68 & 54.71 & 59.90 \\
Supervised + FIST & 63.54 & 63.83 & 69.41 & 51.62 & 55.58 & 60.48 \\
Supervised + GIST & 66.33 & 66.95 & \textbf{70.27} & \textbf{53.51} & \textbf{58.11} & 60.94 \\
Supervised + RIST & \textbf{66.71} & \textbf{68.28} & 69.90 & 53.33 & 57.81 & \textbf{61.38} \\

\hline
\end{tabular}
\caption{Our semantic segmentation results (mIOU) on the PASCAL VOC 2012 and Cityscapes validation dataset.All add-ons are added to FIST, RIST and GIST.}
\label{table:det2}
\end{table*}

\begin{table}[H]
\centering
\setlength{\tabcolsep}{55pt}
\begin{tabular}{|l||l|l|}
\hline
Method & RIST & GIST \\
\hline
no add-on & 60.80 & 58.23  \\
+CL  & 61.72 & 62.37 \\
+CL +LE & 64.69 & 64.56\\
+CL +LE +TS & 66.71 & 66.33 \\
\hline
\end{tabular}
\caption{Ablation study of the add-ons for both RIST and GIST for models trained on the Pascal VOC 1/50 subset. Results are reported in mIOU on the validation set.}
\label{table:addons}
\end{table}

\section{Meta-parameters differences in S4GAN and ClassMix}

For Pascal VOC 2012, ClassMix uses a total batch-size of 20 in a single iteration where they load 10 labeled images and 10 unlabeled images~\cite{olsson2020classmix}. For Cityscapes, ClassMix uses a total batch-size of 4 in a single iterations. 
ClassMix also selects the best iterations using the validation set~\footnote{https://github.com/WilhelmT/ClassMix/blob/master/trainSSL.py}. 

For Pascal VOC 2012, S4GAN uses a total batch-size of 24 in a single iteration, where they load 16 labeled images and 8 unlabeled images~\footnote{https://github.com/sud0301/semisup-semseg/blob/master/train\_s4GAN.py}.  
For Cityscapes, S4GAN uses a total batch-size of 15 in a single iteration. 
S4GAN also selects the best iterations using the \href{https://web.archive.org/web/20210316045243/https://github.com/sud0301/semisup-semseg/issues/12}{validation set}~\footnote{It is not clear how the best iterations were selected in the \href{https://github.com/sud0301/semisup-semseg/blob/master/train_s4GAN.py}{source code}, but in the GitHub issues, the author recommended selecting the best iterations using the validation set} \footnote{https://github.com/sud0301/semisup-semseg/issues/12}.

For Pascal VOC 2012, we use a total batch-size of 8 in a single iteration for our ClassMix experiments and we use a total batch-size of 12 in a single iteration for our S4GAN experiments. For Cityscapes, we use a total batch-size of 6 in a single iteration for our ClassMix experiments and we use a total batch-size of 9 in a single iteration for our S4GAN experiments. For both methods, we use a development set to select the best iterations.

\section{Conclusion}

We show that iterative self-training with a fixed human-labels to pseudo-labels ratio (FIST) leads to performance degradation. This degradation can be overcome by alternating training on only human-labels or only pseudo-labels in a greedy (GIST) or random (RIST) fashion. A clear benefit of self-training is that it can easily extend existing architectures. 
We show that both GIST and RIST can further refine models trained with other semi-supervised techniques resulting in a performance boost. 

\clearpage

\chapter{Concluding Remarks}\label{chap:con}

Annotation Efficient Learning (AEL) is the study of algorithms to train machine learning models efficiently with fewer annotations. AEL is important because new data are being uploaded at an unprecedented rate, and it takes a massive workforce to annotate a small fraction of the data. Additionally, studies have shown that classification, the most common machine learning task, does not transfer well to the unseen dataset~\cite{van2017devil,liu2019large}. This finding means that we must retrain our model for every classification task whenever we encounter new data.
We explore different methods for handling AEL in Digital Pathology (DP) and Natural Images (NI), including transfer learning (Chapters~\ref{isbi2020} and~\ref{isbi2022}), data augmentation (Chapter~\ref{midl2022}), metric learning (Chapter~\ref{eccv2020}), and semi-supervised learning (Chapter~\ref{crv2022}). These methods have practical implications for society as they allow machine learning models to be trained with fewer computational and human resources.

The key insight to tackling AEL is to leverage weakly annotated datasets. The strength of annotation depends on the annotation effort and the dataset's domain. For instance, it is reasonable to use classification annotations to tackle semantic segmentation problems, as classification annotations are easier to collect when compared to the collection of segmentation annotations. Another instance is to use a natural image domain dataset to tackle the medical image domain problem, as it is cheaper to collect annotations in the natural image domain compared to the medical image domain.  

There are a few limitations to our work. We did not explore self-supervised learning. Many recent advancements in deep learning show that knowledge learned from self-supervised learning is more generalizable to a new task~\cite{caron2018deep, yan2020clusterfit, caron2020unsupervised, jaiswal2020survey}. Our work leverages pre-trained models, which are trained with human annotations. A simple replacement of pre-trained models with self-supervised learning should boost the performance of our tasks. 

We also did not consider all the possible dataset combinations that could yield better transfer learning performance. The sheer number of datasets makes this task practically challenging. A better way to approach this is to develop a good metric to gauge the usefulness of transferring knowledge from one dataset to another. Ideally, we would be able to select samples from all datasets to create the perfect transfer learning dataset for a given task. 

Additionally, we did not consider synthetic data generation, which is a challenging topic in machine learning. Generative adversarial networks (GAN) have shown promising results in data augmentation by conditioning on unseen poses~\cite{zhi2021pose}. GAN can also generate images with various lighting conditions, such as turning a day-time image into a night-time image for data augmentation purposes in self-driving car~\cite{zhu2017unpaired}. 
For future work, synthetic data generators that can extrapolate beyond the original dataset into the tails of the data generating distribution are a promising direction.

Furthermore, our work in semi-supervised segmentation did not explore efficient training methods in semi-supervised environments. One major bottleneck of training a segmentation model is the batch-size during training, as a segmentation model usually takes a lot of GPU resources. We will need at least twice as many GPU resources in a semi-supervised environment. One way to alleviate this problem is to explore the use of more advanced normalization techniques such as InstaceNorm~\cite{ulyanov2016instance}, LayerNorm~\cite{ba2016layer}, and GroupNorm~\cite{wu2018group}. These newer normalization techniques are better at handling machine learning training with smaller batch sizes than BatchNorm. We can open doors to better semi-supervised techniques that leverage ensembling with a more memory-efficient training method. 

We hope researchers will benefit from our remarks and discover new opportunities for future research in AEL. We hope to see machine learning models that can adapt to new environments with fewer human and computing resources, allowing us to close the gap between artificial intelligence and human intelligence.

\singlespace
\newpage
\addcontentsline{toc}{chapter}{Bibliography}
\bibliography{thesis}
\bibliographystyle{acm}

\onehalfspace








\end{document}